\pdfoutput=1

\documentclass{article}
\usepackage{preprint,times}

\usepackage{graphicx}
\usepackage{amsmath,amssymb}
\usepackage{booktabs}
\usepackage{multirow}
\usepackage{float}   %
\usepackage{microtype}
\usepackage{xcolor}
\usepackage{url}
\usepackage[colorlinks,allcolors=blue,breaklinks]{hyperref}
\title{Should This Case Be Adapted? Prediction Fragmentation Controls
Test-Time Adaptation}

\author{\textbf{Lili Wang, Jing Li, Xiaowen Sun, Xiangyu Hu, Zhuangzhuang Gu, Jian Liu,}\\
\textbf{Srihari Nelakuditi, Yan Tong}}

\ppfinalcopy          %
\renewcommand{\headrulewidth}{0pt}

\begin{document}
\maketitle
\lhead{}\chead{}\rhead{}   %

\begin{abstract}
Episodic test-time adaptation (TTA) resets a frozen segmenter to source weights $M_0$ on
each case and adapts for a fixed number of steps. A fixed horizon conflates a cohort-level
question, \emph{how far} to adapt, with an irreducibly per-case one, \emph{whether this
case should be adapted at all}. Cohort means average over that per-case decision: on cross-vendor cardiac MRI the mean $\Delta$Dice from adaptation is statistically indistinguishable from zero while $58.7\%$ of cases are individually made worse. We quantify this harm as
\emph{harmful accepted area} (HA), the harmful fraction of the edited area a controller
actually deploys. Held-out tuning gives a stronger baseline than a fixed horizon, but the budget
it selects does not transfer on either of the two main medical benchmarks, and no global budget can condition on
the case. We show that prediction fragmentation---the disagreement geometry between $M_0$ and
the adapted mask $M_k$---predicts HA with no labels and no additional backward
passes at decision time, at comparable strength on three benchmarks (Spearman $\rho$ $0.50$--$0.60$), at a quarter of gradient-norm's latency. A case-level
router built on it cuts HA from $0.228$ to $0.139$ on a benchmark that took no part in its design, with that design frozen and only its cut-points recalibrated there. On the cardiac
benchmark the design was selected on, the same router cuts HA from $0.129$ to $0.013$ at
matched Dice and $1.10$ deployed updates, against the retrospective-best budget identified
post hoc from evaluation labels, and reduces that $58.7\%$ to $20.0\%$---an upper bound
we quantify. Where the retained cases are not net-helped (as on prostate), the router still cuts HA but concedes accuracy, a boundary we
report explicitly. Thresholds are calibrated once on a labeled split disjoint from
evaluation; deployment decisions use no labels and no gradients. The template ports across
architecture and domain (nnU-Net$\rightarrow$SegFormer, Cityscapes$\rightarrow$ACDC) with coordinate, thresholds and per-bucket actions instantiated per domain.
\end{abstract}

\section{Introduction}

Test-time adaptation (TTA) updates a trained model on the test data itself, with no source data and no labels, to recover accuracy lost to distribution shift.
Episodic TTA is its per-case form and is increasingly deployed on frozen segmentation models: reset to source
weights $M_0$, adapt for a short horizon on each new case, and return the adapted state
$M_k$~\citep{tent2021,eata2022,cotta2022}. In practice that horizon is fixed in advance
and applied uniformly to every case, which conflates two decisions. \emph{How far to
adapt} is a cohort-level question, answerable by tuning the budget on held-out data.
\emph{Whether this case should be adapted at all} is irreducibly per-case, and no global budget can make that decision. That per-case axis is where the reliability of episodic TTA is settled, and cohort means do not show it.

\begin{figure}[t]
\centering
\includegraphics[width=\linewidth]{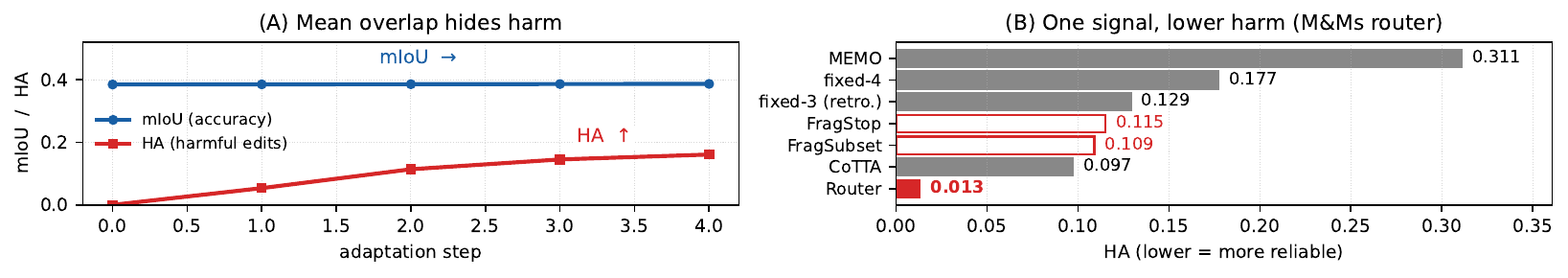}
\caption{\textbf{One signal, from problem to payoff.} (A)~mIoU flat while HA rises
(ACDC rain, lr $10^{-4}$, $K{=}4$; the aggressive budget of Sec.~\ref{sec:acdc} is
steeper still). (B)~M\&Ms router attains the lowest HA at matched Dice (ours in red, solid), a per-deployment result, not universal dominance (Sec.~\ref{sec:prostate}).
Against the retrospective fixed-3 only the router separates cleanly; the two outlined
budget-axis controllers are marginal or not separated (Sec.~\ref{sec:supp-pairwise}).
Cohorts, budgets and panel~(A)'s pooling convention:
Secs.~\ref{sec:supp-eval},~\ref{app:calib}.}
\label{fig:teaser}
\end{figure}

\paragraph{Mean overlap hides harm.}
Adaptation can raise average Dice\,/\,mIoU while, in localized regions, editing the
prediction from correct to incorrect. We call the fraction of edited area that becomes
locally worse the \emph{harmful accepted area} (HA). For instance, on cross-vendor
cardiac MRI cohort-mean $\Delta$Dice is statistically indistinguishable from zero while
$58.7\%$ of cases are individually made worse (Fig.~\ref{fig:teaser}); HA tracks a clinically used boundary metric
we never optimize, and is non-monotone in the budget (Sec.~\ref{sec:prelim}). A
held-out-tuned budget is a stronger baseline than the fixed horizon, and we adopt it. But the reliability-optimal budget does not transfer from calibration to evaluation on either of the two main medical benchmarks, while thresholds fit there are applied unchanged (Sec.~\ref{sec:budget}).

What a global budget cannot do at all is condition on the case. We therefore read the
disagreement geometry between $M_0$ and $M_k$ as a per-case signal, \emph{prediction
fragmentation}. It uses predictions only, with no labels and no gradients at decision time. Among the read-only signals we evaluate, it is the strongest predictor of HA by pooled correlation, at a quarter the latency of gradient-norm and negligible extra memory. It also drops into an episodic TTA routine without changing the adaptation objective. Our main instance is a
case-level router: after one step it sends each case to rollback, a curtailed
adaptation, or the least restricted one. The same signal also instantiates
a stop rule and a modulation of \emph{which} parameters adapt, reported as ablations.

\paragraph{Contributions.}
We define HA and show that mean overlap hides the reliability failure of fixed-budget episodic TTA and that the HA-optimal budget does not survive a change of split on M\&Ms or prostate OOD-all (Sec.~\ref{sec:prelim}). Prediction fragmentation predicts that harm without
labels or gradients at decision time, at comparable strength on three benchmarks
(Sec.~\ref{sec:signal}). A case-level router built on it cuts the individually harmed fraction at matched Dice on cardiac and ports across architecture and domain,
with coordinate, thresholds and per-bucket actions instantiated per domain. Frozen but
for its cut-points, the design transfers to a benchmark it was not selected on
(Secs.~\ref{sec:controllers},~\ref{sec:router}--\ref{sec:acdc}).

\section{Related Work}

\paragraph{Test-time adaptation and fail-safe mechanisms.} Episodic TTA optimizes, without source
data or labels, proxy objectives such as entropy minimization~\citep{tent2021},
batch-norm statistics alignment~\citep{schneider2020bn}, augmentation-consistent
teacher--student updates~\citep{cotta2022}, marginal-entropy minimization over
augmentations~\citep{memo2022}, or a self-supervised auxiliary task~\citep{sun2020ttt}; EATA~\citep{eata2022} adds
sample selection and anti-forgetting, and SAR~\citep{sar2023} filters updates by
entropy and sharpness. These methods change \emph{what} objective adapts $M_0$,
and their sample selection decides \emph{which samples} to adapt on. We take their trajectories as given and ask a downstream question they do not address: given a
materialized trajectory, should \emph{this} case keep its adaptation? A parallel
line monitors adaptation rather than modifying it. AETTA~\citep{lee2024aetta}
estimates accuracy from dropout-inference disagreement to trigger model resets,
building on agreement-based OOD accuracy
estimation~\citep{baek2022agreement}; TEGDA~\citep{tegda2025} proposes a
confidence-calibrated dropout-agreement metric for the same lack of ground-truth
monitoring; and \citet{schirmer2025monitoring} raise an alarm when an adapting
model's risk drifts over a continuous test stream. The last is closest in framing
and differs in unit and output: under per-case reset there is no drift to
accumulate, and a monitor emits an alarm where we emit an action on this case's
prediction. We differ from all three in the signal, which needs no stochastic
forward passes. FOA~\citep{foa2024} removes backpropagation from the update
itself, an axis orthogonal to our backprop-free \emph{control} signal and one that
matters on its own, since rankings shift under a compute
budget~\citep{alfarra2024evaluation}.

\paragraph{Label-free quality assessment and disagreement signals.} Estimating
segmentation quality without annotations is a long-standing problem in medical
imaging~\citep{audelan2021unsupervised}. Under domain shift, existing signals
derive from per-pixel confidence, stochastic-pass
disagreement~\citep{lee2024aetta,tegda2025}, or model-space statistics:
GraTa~\citep{grata2025} corrects gradient directions via alignment between pseudo-
and auxiliary gradients, and DeYO~\citep{deyo2024} augments entropy with a
shape-sensitivity score. Model-space controllers change
\emph{how} an update is formed, so the two axes compose rather than compete. A fragmentation router wraps a GraTa-optimized trajectory as it wraps ours, though there a matched-quota random control reproduces most of what it buys (Sec.~\ref{app:newobj}). We include GraTa's alignment as a signal-level baseline (Sec.~\ref{sec:signal}). Our signal reads the connected-component
geometry of the disagreement between $M_0$ and $M_k$, a per-case, backprop-free
statistic. Early-exit
methods~\citep{branchynet2016,graves2016act} halt inference; our stop rule halts
\emph{adaptation} of a frozen predictor.

\paragraph{Evaluation granularity and harm.} Region-sensitive metrics such as Surface Dice~\citep{nikolov2021surfacedice}, Boundary IoU~\citep{cheng2021boundaryiou} and the Metrics Reloaded recommendations~\citep{maierhein2024metricsreloaded} address localized errors that
aggregate overlap metrics can underemphasize, but all compare against
ground truth offline. Harmful accepted area (HA, Sec.~\ref{sec:prelim}) shares
that concern with a different reference: it scores the edits a controller actually
deploys against the source prediction $M_0$.

\section{Preliminaries: The Reliability Failure of Fixed-Budget TTA}
\label{sec:prelim}

Episodic TTA on a frozen $M_0$ produces predictions $M_1, \dots, M_T$. Let
$a$, $b$, $y$ be the per-voxel argmax labels of $M_0$, $M_k$ and GT. On the medical benchmarks all three are first binarized to foreground, so only foreground/background edits enter $D$. On driving, the $19$-class labels are used as is, so any label
change does (Sec.~\ref{app:stoprule}).
The disagreement mask is $D = \{x : a(x) \neq b(x)\}$; we extract connected components
$\{R_j \subseteq D\}$ with $|R_j| \geq \tau_{\text{area}}$. A region is \emph{harmful} if
$\mathrm{err}(b, R_j) > \mathrm{err}(a, R_j)$, where
$\mathrm{err}(m, R) = \frac{1}{|R|}\sum_{x \in R}\mathbb{I}[m(x) \neq y(x)]$. Deployment
HA is the harmful fraction of edited area:
\begin{equation}
\mathrm{HA}(M_k) = \sum_{j:\,\mathrm{harmful}(R_j)} \frac{|R_j|}{|D|},
\qquad \mathrm{HA}(M_k)=0 \text{ when } D=\emptyset.
\label{eq:ha}
\end{equation}
Two conventions make HA a deployment metric. It is \emph{$M_0$-referenced}: every
controller's output is scored against the same source $a{=}M_0$, so ``harmful edit''
means the same thing for every method. The edited set $D$, by contrast, is the one that controller actually deploys ($M_0$ vs.\ $M_4$ for fixed-4, vs.\ $M_\tau$ for a stop, the
post-policy edits for a router). HA is therefore comparable across methods, but its
denominator $|D|$ moves with the gate, a property with consequences in both the budget
(Sec.~\ref{sec:budget}) and the gate (Sec.~\ref{sec:controllers}). And cohort HA is the
unweighted mean of per-case HA, so a small edit set counts as much as a large one.

\paragraph{$\mathrm{HA}{=}0$ means ``not judged harmful'', not ``judged harmless''.}
Two things score zero. Reverting to $M_0$ deploys no edits ($D{=}\emptyset$), so full
rollback attains $\mathrm{HA}{=}0$ trivially and HA is not meaningful alone. It is therefore always reported beside Dice\,/\,mIoU and deployed steps, with edit coverage,
matched-rollback controls and the composition of the declined half in
Secs.~\ref{app:accounting}, \ref{app:matched}, \ref{app:defenses}. A case whose
disagreement set is non-empty but has no component of at least $\tau_{\text{area}}$
($16$ voxels on medical, $200$\,px on ACDC) also scores zero; on cardiac this covers $376$ of $920$
(case, step) rows. The absolute value depends on that threshold; on cardiac, the router's rank and the sign of its margin do not, across an eightfold change in it (Sec.~\ref{app:tau}).

HA is a reliability lens, not a loss we optimize or a replacement for Dice\,/\,mIoU
(Sec.~\ref{sec:defenses}). It tracks a boundary
metric we never target: Spearman $\rho{=}0.62$ with HD95 of $M_k$ on prostate OOD-hard
($n{=}34$) while cohort-mean $\Delta$HD95 is ${-}13$\,mm. The failure is visible case by
case on a clinically used metric even as that metric's own mean improves
(Sec.~\ref{app:hd95}).

\subsection{HA Is Not Monotone in the Adaptation Budget}
\label{sec:budget}

Because HA is a \emph{fraction} of a controller-dependent edit set, it need not increase
with the budget: additional steps enlarge the denominator $|D|$ as well as the harmful
numerator. Table~\ref{tab:budget-sweep} sweeps the fixed budget on the same trajectories
used throughout. On all three medical splits HA is minimized at $K{=}3$ and is higher
at $K{=}4$, so the default fixed horizon is not the strongest fixed baseline on these trajectories. Dice, meanwhile, moves little across the sweep ($0.002$, $0.007$ and $0.015$ on the
three rows). We therefore adopt the held-out-tuned budget, not the fixed horizon, as our
primary point of comparison. (On ACDC rain under TENT the ladder rises monotonically over $K{\leq}8$; the shape belongs to EATA on this horizon, not to HA.)

\begin{table}[t]
\centering
\small
\begin{tabular}{lcccc}
\toprule
Split & $K{=}1$ & $K{=}2$ & $K{=}3$ & $K{=}4$ \\
\midrule
M\&Ms ($n{=}230$)            & 0.186 & 0.164 & \textbf{0.129} & 0.177 \\
Prostate OOD-all ($n{=}93$)  & 0.242 & 0.211 & \textbf{0.171} & 0.262 \\
Prostate OOD-hard ($n{=}34$) & 0.290 & 0.309 & \textbf{0.257} & 0.318 \\
\bottomrule
\end{tabular}
\caption{\textbf{HA vs.\ fixed budget} (EATA trajectories, anchor variant in Sec.~\ref{sec:supp-eata}; Dice omitted, it varies by
$0.002$, $0.007$ and $0.015$ across the three rows). Per-split
holdout$\to$eval selection is in Sec.~\ref{sec:supp-ladders}.}
\label{tab:budget-sweep}
\end{table}

\paragraph{But the optimal budget cannot be selected in advance.} Holdout selection gives
$K{=}1$ on prostate and $K{=}2$ on M\&Ms against an evaluation optimum of $K{=}3$ on both,
and under joint bootstrap ($B{=}5000$) the two optima agree on only $7.1\%$ and $0.3\%$ of
resamples. Every selection rule we tried makes the same choice, while the thresholds and
routing cut-points fit on the same splits are applied unchanged
(Secs.~\ref{sec:router},~\ref{sec:prostate},~\ref{sec:supp-ladders}).

\paragraph{A global budget cannot decide case by case.} On M\&Ms the cohort-mean
$\Delta$Dice from fixed-budget adaptation is $+0.00006$ ($95\%$ CI $[-0.0012,+0.0014]$,
i.e.\ $+0.000$ at the precision of every table here), yet $58.7\%$ of cases are
individually made \emph{worse}. Tuning $K$ moves the mean but cannot separate these two populations, because it does not condition on the case. That axis is what the rest of the paper addresses.

\section{Method}

\begin{table}[t]
\centering
\small
\setlength{\tabcolsep}{4pt}
\begin{tabular}{llccc}
\toprule
Benchmark (objective, budget) & Signal & $\rho$ vs HA & $\rho(\Delta\text{signal},\Delta\mathrm{HA})$ & backprop \\
\midrule
Prostate MRI, $n{=}124$                & $n_{\mathrm{reg}}$ & $\mathbf{0.594}$ & $0.464$ & No \\
(EATA, lr $5{\times}10^{-4}$, $K{=}4$) & $\delta$           & $0.557$ & $\mathbf{0.515}$ & No \\
                                       & entropy            & $0.119$ & $-0.290$ & No \\
                                       & MSP                & $-0.161$ & $0.278$ & No \\
                                       & grad-norm          & $0.270$ & $-0.035$ & Yes \\
                                       & GraTa-cos          & $-0.093$ & $-0.009$ & Yes ($\times 2$) \\
\midrule
Cardiac MRI, $n{=}230$                 & $n_{\mathrm{reg}}$ & $\mathbf{0.597}$ & $\mathbf{0.455}$ & No \\
(EATA, lr $5{\times}10^{-4}$, $K{=}4$) & $\delta$           & $0.516$ & $0.347$ & No \\
                                       & entropy$^{\dagger}$ & $0.032$ & $-0.006$ & No \\
                                       & MSP$^{\dagger}$     & $-0.012$ & $0.017$ & No \\
                                       & $1-$agreement$^{\dagger}$ & $0.264$ & $0.051$ & No ($8\times$ fwd) \\
                                       & grad-norm          & $0.146$ & $0.035$ & Yes \\
\midrule
Driving, $n{=}406$                     & $n_{\mathrm{reg}}$ & $\mathbf{0.503}$ & $\mathbf{0.290}$ & No \\
(TENT, lr $10^{-4}$, $K{=}4$)          & $\delta$           & $0.478$ & $0.238$ & No \\
\midrule
Driving, $n{=}306$                     & $n_{\mathrm{reg}}$ & $-0.147$ & $0.046$ & No \\
(TENT, lr $2{\times}10^{-3}$, $K{=}8$) & $\delta$           & $\mathbf{0.139}$ & $\mathbf{0.187}$ & No \\
\bottomrule
\end{tabular}
\caption{Harm-predictive comparison across three benchmarks (EATA rows use the anchor
variant of Sec.~\ref{sec:supp-eata}). Correlations pool $(\text{case},\text{step})$ pairs
with step ${\geq}1$; $\Delta$ differences consecutive steps within a case. Driving is ACDC
pooled over conditions, reported at both budgets because the association inverts at the
aggressive one. $^{\dagger}$From a verification rerun of the same cardiac trajectories,
whose shared-signal correlations match the archived run to $\pm0.002$
(Sec.~\ref{app:rerun}); the archived run stays authoritative for every main-table number.
The GraTa-cos comparison is available on prostate only. MSP is maximum softmax probability. Clustered intervals, the per-step decomposition and the case-level ranking AUROCs are in Sec.~\ref{app:ci}.}
\label{tab:signal-ha}
\end{table}

\subsection{Prediction Fragmentation as a Harm-Predictive Signal}
\label{sec:signal}

At step $k$, define $s_k = \phi(M_0, M_k) = (n_k, \Delta n_k, \delta_k, \Delta\delta_k,
\bar H_k)$, where $n_k$ is the number of disagreement regions, $\delta_k$ the disagreement
ratio, $\Delta n_k = n_k - n_{k-1}$, and $\bar H_k$ the mean predictive entropy on
disagreement voxels. We use ``fragmentation'' as shorthand for this family of
disagreement-geometry statistics; $n_{\mathrm{reg}}$ and $\delta$ are read from the same
two masks (Table~\ref{tab:signal-ha}), and which one is deployed is a per-domain choice
(Sec.~\ref{sec:controllers}, Sec.~\ref{app:algo}). On a per-site holdout we fit z-score
statistics $(\mu_n, \sigma_n), (\mu_\delta, \sigma_\delta)$ and define a scalar
fragmentation risk
\begin{equation}
r_k = \tfrac{1}{2} z(n_k;\mu_n,\sigma_n) + \tfrac{1}{2} z(\delta_k;\mu_\delta,\sigma_\delta).
\label{eq:risk}
\end{equation}

\paragraph{What the signal is, and what it is not.} Fitting $(\mu,\sigma,\tau)$ is done
\emph{once} on a held-out labeled split disjoint
from the online test stream, exactly as TENT/EATA/CoTTA hyperparameters are. At deployment no labels and no gradients are used, so the label-free\,/\,backprop-free claim refers strictly
to per-case decision time. The signal operates at the \emph{case} level and not below it:
a higher $n_{\mathrm{reg}}$ reflects many mutually unrelated changes rather than one
coherent revision, and it is that aggregate instability that correlates with harm. We do
\emph{not} claim an individual harmful edit is geometrically identifiable; a region-level
analysis reverses the ``harmful edits are small debris'' picture on prostate
(Sec.~\ref{sec:supp-edit-geometry}).

\paragraph{The association, and how it compares.} Step $0$ is excluded because HA${=}0$ there by construction, and every value comes from stored per-step statistics (exceptions in Sec.~\ref{app:rerun}).
Fragmentation is largely complementary to confidence: on prostate its level correlation
with entropy and MSP is near zero ($0.03$ and $-0.10$; Sec.~\ref{sec:supp-signal-crossdomain}), as expected if harm is carried by
the distribution of disagreement rather than by per-pixel certainty, since a wrong edit can be
made confidently. Against HA it ranks highest of the read-only signals on both medical
benchmarks by pooled correlation. The association also replicates at comparable budgets: $n_{\mathrm{reg}}$ reaches $0.594$ / $0.597$ / $0.503$ on prostate, cardiac and driving, and the step-differenced column $0.464$ / $0.455$ / $0.290$, across nnU-Net and SegFormer,
MRI and natural images, EATA and TENT. The comparison that motivates the signal survives the change of benchmark in our favor. Grad-norm falls from $0.270$ to $0.146$ between the
two medical benchmarks while fragmentation holds near $0.59$, so the read-only signal
separates \emph{further} from the backprop one on the second benchmark (cardiac).
GraTa's alignment~\citep{grata2025} is near-orthogonal to harm despite two backward
passes per step; we compare at the signal level since alignment governs \emph{how} an
update is formed, not how many. Ranking cases is only possible for the harm HA measures. No signal we test predicts case-level Dice degradation (AUROC $0.37$--$0.56$; grad-norm on cardiac is inverted, $[0.30,0.45]$), while every signal ranks HA-level harm ($0.62$--$0.72$; MSP inverted; Sec.~\ref{app:ci}). On Prostate158 neither $\delta$ nor $n_{\mathrm{reg}}$ does ($0.475$, $0.506$; Sec.~\ref{app:matched}).

\subsection{Compiling the Signal into Controllers}
\label{sec:controllers}

The signal is read through a different reduction in each controller: FragStop halts on a
rise in $n_k$ with nothing calibrated, the ACDC $\tau$-stop thresholds the single-term
$z(n_k)$, FragSubset thresholds the two-term risk $r_k$ of Eq.~\ref{eq:risk} at zero, and
the router
splits one per-domain coordinate $u_k = g_d(s_k)$ at holdout tertiles
(Sec.~\ref{sec:supp-calib}). Eq.~\ref{eq:risk} therefore enters at one point only, the
parameter-subset modulation. Sec.~\ref{sec:budget} showed the budget axis cannot be settled in advance, so
we take the \emph{per-case} action as primary and report the budget-axis actions as
ablations that establish the signal compiles.

\paragraph{COQR Router.} Case-level online quantile routing (COQR) acts on the axis no
global budget reaches: whether \emph{this} case should keep its adaptation. The skeleton is one rule: score the case after step~1, split the score at holdout-calibrated
tertiles (which need not divide the evaluation stream evenly; Sec.~\ref{sec:supp-calib}),
and attach an action of decreasing severity to each bucket, from rollback through a curtailed adaptation to the least restricted one (Table~\ref{tab:supp-routing-coord}). Rollback
returns $M_0$ and deploys no edits. What the other two buckets do is instantiated per domain. On the two medical benchmarks each bucket deploys a fixed depth: $3$ and $2$-then-shrink ($\alpha{=}0.5$) on M\&Ms, $1$ and $1$ on prostate. The partition is three-way; prostate's action set is two-way. On driving, both buckets run the stop rule below within a per-bucket cap, and their depth varies by case. The fragmentation state therefore enters the bucket assignment in every domain, and
the depth only on driving. The scoring coordinate is likewise per domain ($\delta$ on
M\&Ms and ACDC, $n_{\mathrm{reg}}$ on prostate). We do not claim one coordinate serves every setting: Sec.~\ref{app:swin} gives a backbone on which $\delta$ fails and $n_{\mathrm{reg}}$ does not (on few scorable cases). Sec.~\ref{app:algo} documents how each was chosen, including four choices made with an evaluation split visible. COQR is \emph{not} a learned meta-controller, and
no evaluation label enters the per-case decision: the cut-points are fixed once on the
held-out split, then applied unchanged online. The decision needs no gradient, since it reads $s_k$ from the two masks. But it is taken \emph{after} one adaptation step, so a rolled-back case still pays that step. What rollback saves is the remaining budget, and the
\emph{Steps} column counts deployed updates (Sec.~\ref{app:calib}).

\paragraph{Two budget-axis instantiations.} \textbf{FragStop} halts at the
first rise $\Delta n_k > 0$, evaluated from $k{\geq}2$, so at least one step is always kept and
the rule is the most conservative feasible instance thereafter. \textbf{FragSubset} adds
modulation on top of that stop: it updates decoder normalization only when $r_k>0$ and
all normalization otherwise, so high fragmentation narrows \emph{what may move} rather
than reweighting the objective. Whether that condition fires at all is a property of
the benchmark's risk distribution rather than of the rule: it fires on $14$--$24\%$ of
prostate steps and on no M\&Ms step (Sec.~\ref{app:ablations}). The two other members of the modulation family, and the
configuration names, are in Sec.~\ref{app:ablations}. Both improve on the
\emph{deployable} fixed budget; against the retrospective one only FragSubset on M\&Ms
separates, and marginally
(Sec.~\ref{sec:ablations}); we report them to establish that one statistic instantiates
stopping, modulation and routing, not as gains over a tuned budget.

\paragraph{HA cannot be certified directly.}
HA's denominator moves with the gate, which defeats monotone conformal certification: the risk must be monotone in the threshold and deployment HA is not. Fixing the denominator
gives a surrogate that admits the standard guarantee unchanged. Sec.~\ref{sec:supp-crc-op} reports the construction and the gap between that surrogate and deployed HA, $1.7\times$ to $4.0\times$ across all seven feasible operating points.

\section{Experiments}

All methods share one episodic loop and one calibration rule. Every threshold is fit once on a labeled holdout disjoint from the evaluation stream: $(\mu,\sigma,\tau)$, the routing cut-points and $\hat\lambda$. Ground truth enters only offline (Sec.~\ref{sec:supp-eval} lists where). Our primary comparison is against the
\emph{deployable} budget, the fixed $k$ that holdout selects.\footnote{Tiers:
\textbf{A} vs.\ the deployable budget ($k{=}2$ on M\&Ms, $k{=}1$ on prostate),
\textbf{A$'$} vs.\ fixed-4, \textbf{B} vs.\ the retrospective eval-best $k$ (not
deployable), \textbf{C} descriptive. Intervals are $B{=}5000$ paired resamples; $P$
denotes $P(\Delta\mathrm{HA}{<}0)$ over them; full protocol in Sec.~\ref{sec:supp-eval}.}

\subsection{Cardiac MRI: Matched Dice, Lower Harm}
\label{sec:headline}
\label{sec:router}

\begin{table}[t]
\centering
\small
\setlength{\tabcolsep}{3pt}
\begin{minipage}[t]{0.49\linewidth}\centering
\begin{tabular}{lccc}
\toprule
Method & Dice & HA & Steps \\
\midrule
$M_0$ (no adapt, ref.) & 0.849 & 0.000 & --- \\
\midrule
fixed-2 \emph{(deployable)} & 0.850 & 0.164 & 2.00 \\
fixed-3 \emph{(retrospective)} & 0.850 & 0.129 & 3.00 \\
fixed-4 \emph{(default)} & 0.849 & 0.177 & 4.00 \\
\midrule
FragStop (ours) & $\mathbf{0.850}$ & 0.115 & 3.43 \\
FragSubset (ours) & $\mathbf{0.850}$ & 0.109 & 3.43 \\
COQR Router (ours) & 0.849 & $\mathbf{0.013}$ & $\mathbf{1.10}$ \\
\midrule
MEMO & 0.841 & 0.311 & 4.00 \\
CoTTA & 0.849 & 0.097 & 4.00 \\
\bottomrule
\end{tabular}
\caption{M\&Ms cardiac Vendor-B OOD ($n{=}230$, online); $k{=}1$ is in
Table~\ref{tab:budget-sweep}. Fixed-$k$ rows are successive checkpoints of one
episodic EATA trajectory. The router rolls back $130$ of $230$ cases; \emph{Steps}
is the deployed depth (executed updates: Sec.~\ref{app:defenses}). The FragSubset
modulation never fires on this benchmark; its difference from FragStop traces to
four cases (Secs.~\ref{sec:supp-eata}, \ref{app:ablations}).}
\label{tab:mnm}
\end{minipage}\hfill
\begin{minipage}[t]{0.49\linewidth}\centering
\begin{tabular}{lccc}
\toprule
Method & Dice & HA & Steps \\
\midrule
$M_0$ (no adapt, ref.) & 0.738 & 0.000 & --- \\
fixed-1 \emph{(deployable)} & 0.752 & 0.242 & 1.00 \\
fixed-2 & 0.753 & 0.211 & 2.00 \\
fixed-3 \emph{(retrospective)} & 0.751 & 0.171 & 3.00 \\
fixed-4 \emph{(default)} & 0.746 & 0.262 & 4.00 \\
FragStop (ours) & 0.749 & 0.183 & 2.77 \\
FragSubset (ours) & $\mathbf{0.755}$ & 0.166 & 2.68 \\
COQR Router (ours) & 0.739 & $\mathbf{0.099}$ & $\mathbf{0.55}$ \\
MEMO & 0.740 & 0.278 & 4.00 \\
CoTTA & 0.732 & 0.122 & 4.00 \\
\bottomrule
\end{tabular}
\caption{Prostate MRI OOD-all ($n{=}93$), online; conventions as in
Table~\ref{tab:mnm}. Every method is run on the identical case set. The router
rolls back $42$ of $93$ cases. The harder $n{=}34$ subset and the ablation
configuration names are in Secs.~\ref{sec:supp-oodhard} and~\ref{app:ablations}.}
\label{tab:prostate}
\end{minipage}
\end{table}

\begin{figure}[t]
\centering
\includegraphics[width=0.72\linewidth]{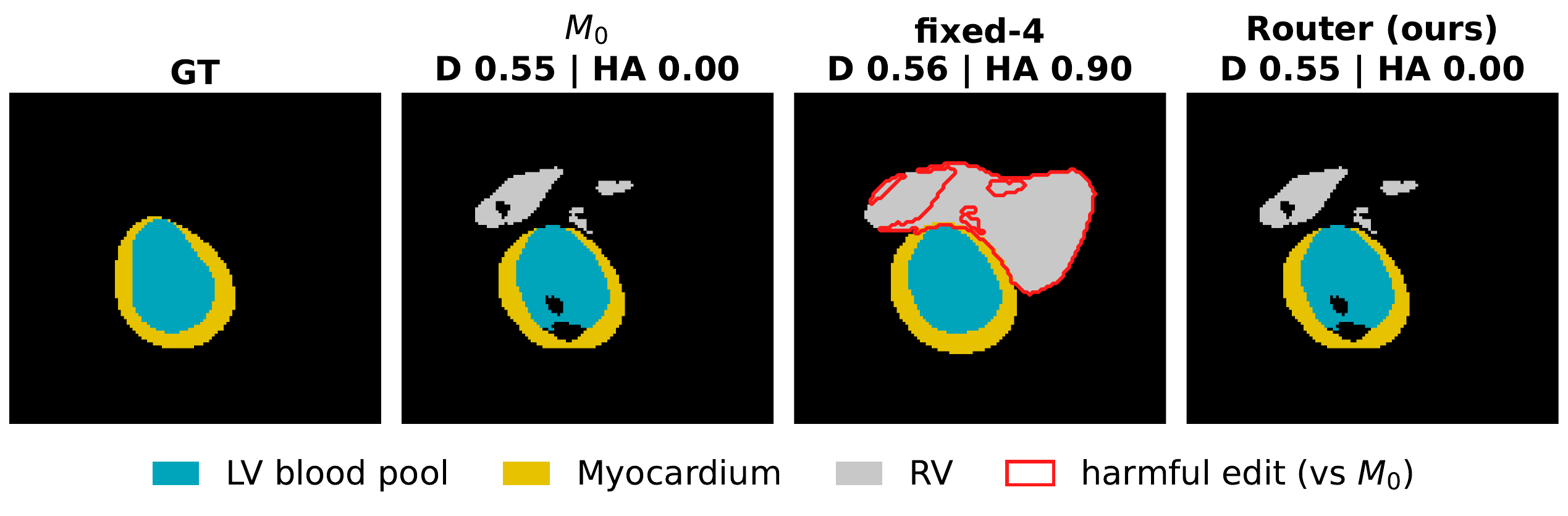}
\caption{Harm invisible to Dice (M\&Ms, router, \texttt{A8E1F4\_frame09}). L$\rightarrow$R:
GT (evaluation reference only); $M_0$; fixed-4, with every harmful accepted edit outlined in
red; ours, which routes this case to rollback and so returns $M_0$ unchanged. This slice has
no RV in GT, so the foreground Dice we report is blind to the false-RV expansion HA flags. Panel metrics are slice-level; selection protocol in
Sec.~\ref{sec:supp-qual-select}.}
\label{fig:med-qual}
\end{figure}

On the cross-vendor cardiac split of M\&Ms~\citep{mms2021} with a frozen
nnU-Net~\citep{nnunet2021} (deployable budget $k{=}2$, retrospective $k{=}3$; Sec.~\ref{sec:budget}), the router attains the lowest HA of any adaptive method
($0.013$) at $1.10$ deployed updates per case. It clears both references: Tier-A against fixed-2, and Tier-B against
the retrospective fixed-3 ($\Delta$HA $-0.116$ $[-0.152,-0.081]$), a budget that could
only have been identified post hoc, with evaluation labels. The
router rolls back $130$ of $230$ cases, and of the $100$ it retains only $17$ deploy an edit with a component above the size threshold. Retained-only HA is $0.031$, and the
deployed edit set covers $\delta{=}3{\times}10^{-5}$ over all $230$ cases, an order of
magnitude below the
fixed budgets ($3.2{\times}10^{-4}$ at fixed-3). The headline is the abstention component Sec.~\ref{sec:prelim} says to look for: a small,
well-chosen edit set, not a large harmless one (Sec.~\ref{app:accounting}). It is not bought with overlap: $\Delta$Dice vs.\ fixed-3 is
$-0.0008$ $[-0.0022,+0.0006]$ and vs.\ $M_0$ at full precision $+0.000016$
$[-0.00007,+0.00010]$, the retained buckets nearly cancelling. On a stream with no net
accuracy to gain, that is the claim: the router recovers the source model's accuracy
\emph{and} its safety without being told in advance which cases adaptation would have
hurt. The objection that $M_0$ already attains that pair here is the per-case form of
Sec.~\ref{sec:budget}'s budget objection. \emph{Not adapting} is the right action on this
stream, but which streams it is right on is known only after the fact. The sign is not even constant within a dataset: at one learning rate and one backbone the router
sits above $M_0$ on ACDC fog, rain and snow ($+0.007$ to $+0.010$ mIoU) and below it on night ($-0.0015$; Sec.~\ref{app:defenses}). No cohort-level rule reads that off a stream
in advance.
Fig.~\ref{fig:med-qual}
shows fixed-4 buying $+0.01$ slice Dice at HA $0.90$ while the router returns $M_0$.
Remaining pairwise intervals are in Sec.~\ref{sec:supp-pairwise}.

\paragraph{The same design, frozen, on a benchmark it never saw.}
Because the M\&Ms design was selected with evaluation-split visibility
(Sec.~\ref{app:algo}), we test it where that exposure cannot act: coordinate, tertile structure
and per-bucket actions frozen, cut-points refit on a $32$-case holdout, on Prostate158~\citep{prostate158}
($n{=}126$), which took no part in the sweep. The frozen design cuts HA against that
benchmark's own retrospective-best budget from $0.228$ to $0.139$ ($-0.089$
$[-0.124,-0.055]$) at a $\Delta$Dice interval containing zero, and sits above $M_0$. Its
relative HA reduction there ($39\%$) matches prostate OOD-all ($42\%$) rather than
M\&Ms ($90\%$); Sec.~\ref{app:algo} gives the panel and three accounts of that gap; at matched
quota and depths the ranking there is not better than random (Sec.~\ref{app:matched}).

\subsection{Prostate MRI: Where the Per-Case Trade Is Unfavorable}
\label{sec:prostate}

\begin{table}[t]
\centering
\setlength{\tabcolsep}{4pt}
\small
\begin{tabular}{llcccc}
\toprule
Split & Method & helped ($\mu_{\Delta^+}$) & hurt ($\mu_{\Delta^-}$) & rolled back & overall $\mu_\Delta$ \\
\midrule
M\&Ms & fixed-4 & 41.3 ($+0.007$) & 58.7 ($-0.005$) & --- & $+0.000$ \\
M\&Ms & Router & \textbf{23.5} ($+0.001$) & 20.0 ($-0.001$) & 56.5 & $+0.000$ \\
\midrule
Prostate & fixed-4 & 51.6 ($+0.046$) & 48.4 ($-0.031$) & --- & $+0.009$ \\
Prostate & Router & 21.5 ($+0.019$) & \textbf{29.0} ($-0.011$) & 45.2 & $+0.001$ \\
\midrule
ACDC night & TENT$_8$ & 42.5 ($+0.028$) & 57.5 ($-0.042$) & --- & $-0.012$ \\
ACDC night & Router & \textbf{34.0} ($+0.018$) & 25.5 ($-0.030$) & 40.6 & $-0.002$ \\
\bottomrule
\end{tabular}
\caption{Case-level decomposition of $\Delta$ against $M_0$ (\% of all cases;
$\mu_\Delta$ at three decimals, which is what hides the $58.7\%$). On M\&Ms and
night the router retains more helped than hurt; on prostate it does not (bold marks the
larger of the two on each router row).
Remaining ACDC conditions and the prostate zero-$\Delta$ convention are in
Sec.~\ref{sec:supp-help-hurt}.}
\label{tab:helphurt}
\end{table}

On the multi-site prostate benchmark~\citep{prostate2020}, again with a frozen nnU-Net
(Table~\ref{tab:prostate}), the router still cuts HA but fails our own Dice-matched
standard, and we report that as the boundary. It clears both references (Tier-A vs.\
fixed-1, $\Delta$HA $-0.142$ $[-0.194,-0.094]$; Tier-B vs.\ fixed-3, $-0.072$
$[-0.139,-0.007]$) at $0.55$ deployed updates ($1.00$ SGD steps counting the discarded
probe, a quarter of the default budget). The cost is on the other axis: $\Delta$Dice vs.\
fixed-1 is $-0.0130$ $[-0.0252,-0.0022]$, an interval that excludes zero, unlike every
Dice comparison on M\&Ms; at $0.739$ the router sits at the un-adapted source, having
rolled back $42$ of $93$ cases.

Table~\ref{tab:helphurt} says why. The router separates two opposed populations rather
than shifting an average: on M\&Ms the individually harmed fraction falls $58.7\% \rightarrow 20.0\%$ at the same cohort mean. The router also makes the boundary a condition rather than a split we happen to lose on. On both benchmarks the signal orders source-model difficulty: rolled-back cases have a weaker $M_0$ than kept
ones ($0.836$ vs.\ $0.866$ on M\&Ms, $0.578$ vs.\ $0.869$ on prostate). What differs is
the kept half: on M\&Ms helped outnumber hurt and $\Delta$Dice
against the deployable budget contains zero ($-0.0015$ $[-0.0032,+0.0001]$); on prostate
hurt outnumber helped and it does not. Ordering difficulty is
not separating benefit from harm, and only the second buys reliability for free. On prostate fixed-4 is close to even over the whole split ($48.4\%$ hurt), so ranking by risk selects against benefit. The failure is testable in advance: the helped/hurt balance of the
retained bucket can be read off a labeled holdout before deployment. Matched-bucket controls agree (Sec.~\ref{app:matched}):
fragmentation beats random at the same rollback rate ($P{=}0.010$) but is
indistinguishable from confidence-scored routing here, whereas on M\&Ms only
fragmentation separates from every alternative (Sec.~\ref{sec:defenses}). On the hardest
$n{=}34$ subset MEMO attains the best HA and, among the non-fixed methods, the best Dice, at
full budget with extra backprop, and the router is not reportable (the low bucket is empty; Sec.~\ref{sec:supp-oodhard}).

\subsection{Transfer and the Signal's Boundary: Cityscapes $\rightarrow$ ACDC}
\label{sec:acdc}

We transfer the controller template to driving segmentation, changing backbone, domain and objective: a frozen Cityscapes-pretrained SegFormer-b0~\citep{segformer2021,cityscapes2016} (19 classes), the ACDC adverse conditions~\citep{acdc2021} as targets, episodic TENT-style adaptation, the same class-agnostic HA and signal, thresholds and tertiles calibrated on ACDC \emph{train} and evaluated on \emph{val}. Under a budget large enough to over-adapt (lr $2{\times}10^{-3}$, $K{=}8$) the ladder on rain is monotone rather than U-shaped: over
$k\in\{1,2,4,8\}$ mIoU moves $0.378\rightarrow0.385\rightarrow0.389\rightarrow0.387$
while HA rises $0.264\rightarrow0.327\rightarrow0.385\rightarrow0.448$ (all steps in
Sec.~\ref{sec:supp-ladders}). The HA-optimal budget is simply $K{=}1$, and
train$\rightarrow$val selection recovers it in every resample. That poses the objection in its cleanest form: if the budget is both optimal and selectable, why route? Against fixed-1 the router still cuts HA by $-0.048$ $[-0.088,-0.012]$ at
Tier-A, while the stop rule is significantly \emph{worse} ($+0.047$ $[+0.025,+0.072]$),
since it is defined against an over-adaptation that $k{=}1$ never produces.

Read against TENT$_8$, adding the stop rule cuts HA on every condition and routing cuts
it again ($0.501\rightarrow0.375\rightarrow0.232$ on night; the other three in
Sec.~\ref{sec:supp-acdc-fogsnow}), at about one deployed update per case except on night ($1.67$), against $K{=}8$. The hard tertile again lands where $M_0$ is weakest. Night is the
boundary: $M_0$ is near-collapsed, TENT$_8$ degrades \emph{below} it ($0.182$ vs.\
$0.194$), and the router's $0.193$ exceeds TENT$_8$ only because TENT falls under the source (damage control, not gain). The same regime is why CoTTA's low HA is not a
stronger safety result: its mIoU sits at $M_0$ on all four conditions, as its Dice does
on M\&Ms and below $M_0$ on prostate. Per-condition thresholds assume the condition is
known at deployment; we did not run the pooled alternative (Sec.~\ref{sec:supp-calib}).

\paragraph{Where the signal stops working.}
\label{sec:signal-transfer}
At this budget the $n_{\mathrm{reg}}$ association with HA inverts to $-0.147$
$[-0.238,-0.053]$ while $\delta$ stays positive ($0.139$; $\Delta$ $0.187$). Restricting the same trajectories to $k{\leq}4$ does not restore $n_{\mathrm{reg}}$ ($-0.054$ rain, $-0.222$ snow, $-0.142$ night), so the cause is update size, not trajectory length.
A near-collapsed $M_0$ makes disagreement large-scale re-guessing rather than sparse
local edits, and counting components stops being informative once almost everything is
recounted. This rules $n_{\mathrm{reg}}$ out as the driving coordinate. The statistic itself is computed without labels, but its correlation with HA, and hence this boundary, is not. The boundary is consistent with a matched control on $\delta$: at matched rollback quota, per-case selection here is not better than random overall, fog excepted (Sec.~\ref{app:matched}), so the routing increment at this
budget comes chiefly from the quota itself.

\subsection{Ablations, Controls and Cost}
\label{sec:ablations}
\label{sec:defenses}

\paragraph{Stopping and modulation.} Both budget-axis controllers lower HA against the
deployable budget (Tables~\ref{tab:mnm}, \ref{tab:prostate}). Paired bootstrap confirms
it for FragSubset ($-0.076$ $[-0.113,-0.041]$ vs.\ fixed-1 on prostate, $-0.055$
$[-0.084,-0.028]$ vs.\ fixed-2 on M\&Ms, where the modulation does not fire and the
reduction is the stop rule's). Against the retrospective optimum the separation
largely disappears (FragSubset vs.\ fixed-3 on OOD-all: $-0.006$ $[-0.046,+0.035]$; the rest in
Sec.~\ref{sec:supp-pairwise}). Modulation beyond stopping is not established by these experiments (OOD-hard: Wilcoxon
$p{=}0.016$, Holm-adjusted $0.065$, CI crossing zero). On OOD-hard, against fixed-4 the variants are ordered in point estimate (Sec.~\ref{app:ablations}): parameter subset $-0.042$, stop alone $-0.033$, entropy weight $-0.005$, objective switch $+0.024$. SAR~\citep{sar2023} bounds the stop rule instead: it falls below the source from step one, leaving no early segment worth keeping (Sec.~\ref{sec:supp-sar}).
The modulation rests on a narrow base: it fires on $14.1\%$ of OOD-hard steps, never on M\&Ms.

\paragraph{Controls and cost.} At its own rollback rate on M\&Ms, the router beats every equally conservative rule on HA (random, confidence- or agreement-scored routing) while staying within $0.001$ on Dice (Sec.~\ref{app:matched}). The controllers never read HA; they
act on fragmentation, which correlates with it at $\rho{=}0.516$ for the coordinate
deployed here, not $1.0$.
Decision-time costs per signal are in Table~\ref{tab:app-compute}.

\section{Discussion and Limitations}

\paragraph{What the controller presumes, and how to check it beforehand.} Three conditions matter, two for the router and one for the budget-axis controllers, and each fails somewhere in this paper; only the first can be checked
without labels. \emph{(i) A usable $M_0$ whose edits are contiguous.} A near-collapsed
$M_0$ turns the controllers into damage control (ACDC night). A transformer backbone edits in scattered voxels rather than regions, so $\delta$ stops carrying the signal while $n_{\mathrm{reg}}$ still does (on few scorable cases, Sec.~\ref{app:swin}). Under large modality shift the
disagreement scale itself moves, by roughly $40\times$ on fundus
(Sec.~\ref{sec:supp-fundus-delta}). These show in the geometry of the edits (Sec.~\ref{app:swin}). \emph{(ii) A trajectory whose HA dips before it rises.} For the budget-axis controllers, per-case control is worth having because HA is non-monotone in $k$, and that shape depends on the objective and the horizon. It holds for EATA on the medical benchmarks and not on ACDC rain, where objective, backbone and budget all differ (Sec.~\ref{sec:budget}); under two other objectives it reproduces in one of six cells, and there only formally; in the other five HA rises monotonically from $k{=}1$ (Sec.~\ref{app:newobj}). A fixed-$k$ ladder on the calibration split indicates which case one is in, though on prostate the two splits disagree (Sec.~\ref{sec:supp-ladders}) and its HA needs labels. \emph{(iii) A retained bucket that is
net-helped.} Where it is not, the router still cuts HA but concedes accuracy (prostate,
Sec.~\ref{sec:prostate}). This too can be read off the labeled calibration split.

\paragraph{What the headline number estimates.} The $90\%$ HA reduction on M\&Ms comes from
the benchmark whose evaluation split was visible during design; how much of it is selection effect is not settled here. Frozen on Prostate158 the same design cuts HA by $39\%$, and the same template with its own coordinate and depths by $42\%$ on prostate OOD-all (Sec.~\ref{app:algo}). Under two other objectives the router lowers HA in all six cells of Sec.~\ref{app:newobj}, but most of each reduction is reproduced without the router's ranking: a matched-quota random control recovers $69\%$--$86\%$ of it, and a different coordinate ($n_{\mathrm{reg}}$) $79\%$--$100\%$. A reader who
treats $39\%$--$42\%$ as the estimate of what the design and its template buy, and $90\%$ as specific to the
benchmark it was selected on, reads the evidence as we do.

\paragraph{Baselines, calibration, and scope.} The baseline suite is not the published one. Our EATA and CoTTA differ in ways Sec.~\ref{sec:supp-eata} lists, our SAR disables EMA recovery (Sec.~\ref{sec:supp-sar}), and MEMO is omitted on ACDC for cost. In Sec.~\ref{app:newobj}, GraTa is ported with recorded deviations and DeYO translated to slices. The routing cut-points are the least stable
part of the pipeline: bootstrapping the $n{=}20$ M\&Ms calibration split moves the bucket
assignment for a median of $15.2\%$ of evaluation cases, and propagating that perturbation
recovers the regime split (Sec.~\ref{app:calibsens}). Every threshold presumes a labeled holdout fit once
offline, so the label-free claim is about per-case decision time rather than the pipeline,
and a site with no labeled data is outside our scope. Estimating the helped/hurt balance
without labels, scale-free formulations of the statistic, learning a policy rather than hand-specifying one, and the same keep-or-revert decision on
test-time-adapted foundation segmenters are open.

\section{Conclusion}
Episodic TTA conflates a cohort-level budget choice with a per-case one, and cohort means
average over the second. Prediction fragmentation reads that decision from two
masks, and risky adaptation is easier to identify than beneficial adaptation. Reporting HA beside mean overlap, and a
budget \emph{ladder} over a single horizon, is actionable today.

\newpage

\subsection*{Ethics statement}
Every dataset used here is public and released by its authors for research use, and the
four medical datasets are de-identified: the M\&Ms cardiac challenge~\citep{mms2021}, the multi-site prostate benchmark~\citep{prostate2020}, whose sites come from the
{NCI-ISBI} 2013 challenge~\citep{bloch2015nciisbi,clark2013tcia}, {I2CVB}~\citep{lemaitre2015i2cvb}
and {PROMISE12}~\citep{litjens2014promise12}, Prostate158~\citep{prostate158},
RIM-ONE-DL~\citep{rimonedl}, Cityscapes~\citep{cityscapes2016} and ACDC~\citep{acdc2021}.
We collected no new human-subject data, accessed no protected health information, and made
no attempt to re-identify any subject; each dataset was used within its published licence
and intended use.

Two limits on how the results should be read. First, HD95 enters this paper only as a
construct-validity check---evidence that HA tracks a boundary metric used in clinical segmentation evaluation (Sec.~\ref{app:hd95})---and not as evidence of clinical performance. Second,
the controllers we study are a reliability layer around test-time adaptation, evaluated
retrospectively on research benchmarks; nothing here constitutes validation for clinical
deployment, which would require prospective evaluation, site-specific calibration and
regulatory clearance. We would also flag the failure mode itself as the safety-relevant
finding: on the cardiac benchmark a cohort-mean $\Delta$Dice indistinguishable from zero
conceals harm to more than half the cases, so a deployment monitored only by mean overlap
can degrade individual predictions without any aggregate signal that it is doing so.

\subsection*{AI use statement}
We used generative AI tools---a code-assistant LLM operated interactively by the
authors---for the tasks listed below. The tools helped
\emph{implement methods} and analysis code: they drafted and, under the authors' direction, ran the scripts that re-ran one
adaptation pipeline on GPU to regenerate the per-voxel probabilities the certification
section needs (Sec.~\ref{sec:supp-crc-op}, Sec.~\ref{app:rerun}); the authors specified the
acceptance criteria and verified every output against the archived run. They also drafted
the scripts that produce several figures. They \emph{provided feedback on methodology and on
experiments}: they proposed the layered acceptance criteria used to validate that rerun
(step-1 gradient-norm agreement to $10^{-6}$; cohort-level Dice and HA tolerances; a pre-specified failure disposition fixed before any result was computed), and they argued
for and against options that the authors then decided, including whether to re-run that
pipeline at all rather than only disclose its provenance. They assisted in
\emph{interpreting results}: cross-checking reported numbers against archived outputs and
source code, and identifying inconsistencies between text, tables and figures. We did not
use generative AI tools to generate synthetic datasets, to develop theoretical models or
conceptual frameworks, to formulate or prove mathematical claims, to propose or refine the
hypotheses of this work, to clean or reformat datasets, or for translation; qualitative and
thematic data analysis is not applicable to this work.

We also used them for creating and editing software code, creating and modifying the figures of this
paper, drafting parts of the paper and editing it for readability, editing the \LaTeX{}
source directly, estimating typeset length, checking terminology and notation for
consistency, identifying and formatting references, searching for information, and
proposing keywords and a one-sentence summary.

The research questions, the experimental design, the benchmarks and the claims of this
paper were determined by the authors. Every change entered the paper only after the authors
reviewed and accepted it, and a substantial number of AI-proposed changes were rejected or corrected on inspection,
among them an incorrect table number asserted from memory rather than from the compiled
cross-references, a table convention that four of the paper's own tables contradicted. We also solicited
several rounds of review from general-purpose AI assistants before submission: one batch
contained fabricated references and was discarded; corrections adopted from the others,
to scope statements, disclosures, figure encoding and wording, were each verified against
the source before entry. We have
reviewed all AI-assisted work, and we take responsibility for the final content of this
work, including text, claims or artifacts produced with the aid of generative AI.

\subsection*{Reproducibility statement}
Every number in this paper comes from a frozen source model, and the appendix is organized
so that each can be traced to its source. Most come from archived per-step trajectories; the
exceptions are identified where they appear; these include the certification operating points
(Sec.~\ref{sec:supp-crc-op}), the verification-only cardiac signal columns
(Sec.~\ref{app:rerun}), the GraTa-cos row (Sec.~\ref{app:rerun}) and the prostate OOD-hard
ablation rows (Table~\ref{tab:supp-oodhard}).
Sec.~\ref{sec:supp-repro} gives the compute environment, software versions, the source models
and the episodic protocol, including the learning rates and budgets behind every table.
Sec.~\ref{app:calib} states what is calibrated, on which split, and with which quantile
convention---which differs between the M\&Ms and driving routers and must be copied per
domain rather than assumed. Sec.~\ref{app:algo} gives the controller in pseudocode together
with the decision-time cost of every signal we compare, and documents the provenance of the
design choices that are not derived, including four made with an evaluation split
visible. Sec.~\ref{sec:supp-eval} fixes the statistical protocol and
Sec.~\ref{sec:supp-pairwise} is the single authoritative source for every
controller-versus-budget interval quoted anywhere in the paper, apart from the SAR probe
and the frozen-design test, which are reported in place; no number has two sources.

Three appendix sections exist specifically so that a reader can check rather than trust.
Sec.~\ref{app:rerun} reports an independent rerun of the cardiac trajectories: it reproduces
the archived run's step-1 gradient norms to float precision, its per-step cohort Dice to
$3.6\times10^{-5}$, and it establishes that a cohort HA mean is reproducible across hardware
only to about $\pm0.002$---a figure we then apply to our own claims, one of which we mark as
too small to read as an ordering. Sec.~\ref{app:matched} reports pre-specified analysis plans
alongside their outcomes, including one expectation that failed because the specification, not
the execution, was wrong. Sec.~\ref{app:calibsens} propagates calibration-set perturbation all
the way to deployed metrics rather than stopping at routing labels.

All implementation code, dataset configurations, aggregated per-region statistical
features, and the archived per-step trajectories will be released on publication; no raw or
reconstructed image files from any benchmark are redistributed. The trajectories are the
input to every offline analysis reported here; analyses that need per-voxel probabilities or
intermediate masks are computed on the verification rerun of Sec.~\ref{app:rerun} and are
identified where reported. None of the control experiments in
Secs.~\ref{app:matched}--\ref{app:calibsens} requires a GPU or a re-run of adaptation.

\bibliographystyle{preprint}
\bibliography{main}

@inproceedings{tent2021,
  author    = {Wang, Dequan and Shelhamer, Evan and Liu, Shaoteng and Olshausen, Bruno and Darrell, Trevor},
  title     = {Tent: Fully Test-Time Adaptation by Entropy Minimization},
  booktitle = {International Conference on Learning Representations (ICLR)},
  year      = {2021},
}

@inproceedings{eata2022,
  author    = {Niu, Shuaicheng and Wu, Jiaxiang and Zhang, Yifan and Chen, Yaofo and Zheng, Shijian and Zhao, Peilin and Tan, Mingkui},
  title     = {Efficient Test-Time Model Adaptation without Forgetting},
  booktitle = {International Conference on Machine Learning (ICML)},
  year      = {2022},
}

@inproceedings{cotta2022,
  author    = {Wang, Qin and Fink, Olga and Van Gool, Luc and Dai, Dengxin},
  title     = {Continual Test-Time Domain Adaptation},
  booktitle = {IEEE/CVF Conference on Computer Vision and Pattern Recognition (CVPR)},
  year      = {2022},
}

@inproceedings{memo2022,
  author    = {Zhang, Marvin and Levine, Sergey and Finn, Chelsea},
  title     = {{MEMO}: Test Time Robustness via Adaptation and Augmentation},
  booktitle = {Advances in Neural Information Processing Systems (NeurIPS)},
  year      = {2022},
}

@inproceedings{sar2023,
  author    = {Niu, Shuaicheng and Wu, Jiaxiang and Zhang, Yifan and Wen, Zhiquan and Chen, Yaofo and Zhao, Peilin and Tan, Mingkui},
  title     = {Towards Stable Test-Time Adaptation in Dynamic Wild World},
  booktitle = {International Conference on Learning Representations (ICLR)},
  year      = {2023},
}

@inproceedings{crc2024,
  author    = {Angelopoulos, Anastasios N. and Bates, Stephen and Fisch, Adam and Lei, Lihua and Schuster, Tal},
  title     = {Conformal Risk Control},
  booktitle = {International Conference on Learning Representations (ICLR)},
  year      = {2024},
}

@inproceedings{segformer2021,
  author    = {Xie, Enze and Wang, Wenhai and Yu, Zhiding and Anandkumar, Anima and Alvarez, Jose M. and Luo, Ping},
  title     = {{SegFormer}: Simple and Efficient Design for Semantic Segmentation with Transformers},
  booktitle = {Advances in Neural Information Processing Systems (NeurIPS)},
  year      = {2021},
}

@inproceedings{cityscapes2016,
  author    = {Cordts, Marius and Omran, Mohamed and Ramos, Sebastian and Rehfeld, Timo and Enzweiler, Markus and Benenson, Rodrigo and Franke, Uwe and Roth, Stefan and Schiele, Bernt},
  title     = {The Cityscapes Dataset for Semantic Urban Scene Understanding},
  booktitle = {IEEE Conference on Computer Vision and Pattern Recognition (CVPR)},
  year      = {2016},
}

@inproceedings{acdc2021,
  author    = {Sakaridis, Christos and Dai, Dengxin and Van Gool, Luc},
  title     = {{ACDC}: The Adverse Conditions Dataset with Correspondences for Semantic Driving Scene Understanding},
  booktitle = {IEEE/CVF International Conference on Computer Vision (ICCV)},
  year      = {2021},
}

@article{mms2021,
  author  = {Campello, V{\'\i}ctor M. and Gkontra, Polyxeni and Izquierdo, Cristian and Mart{\'\i}n-Isla, Carlos and others},
  title   = {Multi-Centre, Multi-Vendor and Multi-Disease Cardiac Segmentation: The {M\&Ms} Challenge},
  journal = {IEEE Transactions on Medical Imaging},
  volume  = {40},
  number  = {12},
  pages   = {3543--3554},
  year    = {2021},
}

@article{nnunet2021,
  author  = {Isensee, Fabian and Jaeger, Paul F. and Kohl, Simon A. A. and Petersen, Jens and Maier-Hein, Klaus H.},
  title   = {{nnU-Net}: A Self-Configuring Method for Deep Learning-Based Biomedical Image Segmentation},
  journal = {Nature Methods},
  volume  = {18},
  number  = {2},
  pages   = {203--211},
  year    = {2021},
}

@inproceedings{prostate2020,
  author    = {Liu, Quande and Dou, Qi and Yu, Lequan and Heng, Pheng-Ann},
  title     = {Shape-Aware Meta-Learning for Generalizing Prostate {MRI} Segmentation to Unseen Domains},
  booktitle = {Medical Image Computing and Computer Assisted Intervention (MICCAI)},
  year      = {2020},
}

@inproceedings{branchynet2016,
  author    = {Teerapittayanon, Surat and McDanel, Bradley and Kung, H. T.},
  title     = {{BranchyNet}: Fast Inference via Early Exiting from Deep Neural Networks},
  booktitle = {International Conference on Pattern Recognition (ICPR)},
  year      = {2016},
}

@misc{graves2016act,
  author        = {Graves, Alex},
  title         = {Adaptive Computation Time for Recurrent Neural Networks},
  year          = {2016},
  eprint        = {1603.08983},
  archivePrefix = {arXiv},
  primaryClass  = {cs.NE},
  note          = {arXiv:1603.08983},
}

@inproceedings{grata2025,
  title     = {Gradient Alignment Improves Test-Time Adaptation for Medical Image Segmentation},
  author    = {Chen, Ziyang and Ye, Yiwen and Pan, Yongsheng and Xia, Yong},
  booktitle = {Proceedings of the AAAI Conference on Artificial Intelligence},
  volume    = {39},
  number    = {3},
  pages     = {2429--2437},
  year      = {2025},
  doi       = {10.1609/aaai.v39i3.32244},
}

@inproceedings{sun2020ttt,
  author    = {Sun, Yu and Wang, Xiaolong and Liu, Zhuang and Miller, John and Efros, Alexei A. and Hardt, Moritz},
  title     = {Test-Time Training with Self-Supervision for Generalization under Distribution Shifts},
  booktitle = {International Conference on Machine Learning (ICML)},
  year      = {2020},
}

@inproceedings{schneider2020bn,
  author    = {Schneider, Steffen and Rusak, Evgenia and Eck, Luisa and Bringmann, Oliver and Brendel, Wieland and Bethge, Matthias},
  title     = {Improving Robustness against Common Corruptions by Covariate Shift Adaptation},
  booktitle = {Advances in Neural Information Processing Systems (NeurIPS)},
  year      = {2020},
}

@book{vovk2005algorithmic,
  author    = {Vovk, Vladimir and Gammerman, Alexander and Shafer, Glenn},
  title     = {Algorithmic Learning in a Random World},
  publisher = {Springer},
  year      = {2005},
}

@article{nikolov2021surfacedice,
  author  = {Nikolov, Stanislav and Blackwell, Sam and Zverovitch, Alexei and Mendes, Ruheena and Livne, Michelle and De Fauw, Jeffrey and Meyer, Clemens and Askham, Harry and Romera-Paredes, Bernardino and Chu, Christopher and others},
  title   = {Clinically Applicable Segmentation of Head and Neck Anatomy for Radiotherapy: Deep Learning Algorithm Development and Validation Study},
  journal = {Journal of Medical Internet Research},
  volume  = {23},
  number  = {7},
  pages   = {e26151},
  year    = {2021},
  doi     = {10.2196/26151},
}

@inproceedings{cheng2021boundaryiou,
  author    = {Cheng, Bowen and Girshick, Ross and Doll{\'a}r, Piotr and Berg, Alexander C. and Kirillov, Alexander},
  title     = {Boundary {IoU}: Improving Object-Centric Image Segmentation Evaluation},
  booktitle = {IEEE/CVF Conference on Computer Vision and Pattern Recognition (CVPR)},
  year      = {2021},
}

@article{maierhein2024metricsreloaded,
  author  = {Maier-Hein, Lena and Reinke, Annika and Godau, Patrick and Tizabi, Minu D. and Buettner, Florian and Christodoulou, Evangelia and Glocker, Ben and Isensee, Fabian and Kleesiek, Jens and Kozubek, Michal and others},
  title   = {Metrics Reloaded: Recommendations for Image Analysis Validation},
  journal = {Nature Methods},
  volume  = {21},
  number  = {2},
  pages   = {195--212},
  year    = {2024},
  doi     = {10.1038/s41592-023-02151-z},
}

@inproceedings{deyo2024,
  author    = {Lee, Jonghyun and Jung, Dahuin and Lee, Saehyung and Park, Junsung and Shin, Juhyeon and Hwang, Uiwon and Yoon, Sungroh},
  title     = {Entropy is not Enough for Test-Time Adaptation: From the Perspective of Disentangled Factors},
  booktitle = {International Conference on Learning Representations (ICLR)},
  year      = {2024},
  url       = {https://openreview.net/forum?id=9w3iw8wDuE},
}

@inproceedings{foa2024,
  author    = {Niu, Shuaicheng and Miao, Chunyan and Chen, Guohao and Wu, Pengcheng and Zhao, Peilin},
  title     = {Test-Time Model Adaptation with Only Forward Passes},
  booktitle = {International Conference on Machine Learning (ICML)},
  series    = {Proceedings of Machine Learning Research},
  volume    = {235},
  pages     = {38298--38315},
  year      = {2024},
  publisher = {PMLR},
  url       = {https://proceedings.mlr.press/v235/niu24a.html},
}

@inproceedings{tegda2025,
  author    = {Zhou, Yubo and Wu, Jianghao and Liao, Wenjun and Zhang, Shichuan and Zhang, Shaoting and Wang, Guotai},
  title     = {{TEGDA}: Test-Time Evaluation-Guided Dynamic Adaptation for Medical Image Segmentation},
  booktitle = {Medical Image Computing and Computer Assisted Intervention ({MICCAI})},
  series    = {Lecture Notes in Computer Science},
  volume    = {15965},
  pages     = {628--637},
  year      = {2025},
  publisher = {Springer},
}

@inproceedings{lee2024aetta,
  title={{AETTA}: Label-Free Accuracy Estimation for Test-Time Adaptation},
  author={Lee, Taeckyung and Chottananurak, Sorn and Gong, Taesik and Lee, Sung-Ju},
  booktitle={IEEE/CVF Conference on Computer Vision and Pattern Recognition (CVPR)},
  pages={28643--28652},
  year={2024}
}

@inproceedings{baek2022agreement,
  title={Agreement-on-the-line: Predicting the Performance of Neural Networks under Distribution Shift},
  author={Baek, Christina and Jiang, Yiding and Raghunathan, Aditi and Kolter, J Zico},
  booktitle={Advances in Neural Information Processing Systems (NeurIPS)},
  year={2022}
}

@article{audelan2021unsupervised,
  title={Unsupervised quality control of segmentations based on a smoothness and intensity probabilistic model},
  author={Audelan, Beno{\^\i}t and Delingette, Herv{\'e}},
  journal={Medical Image Analysis},
  volume={68},
  pages={101895},
  year={2021}
}

@inproceedings{alfarra2024evaluation,
  title={Evaluation of Test-Time Adaptation Under Computational Time Constraints},
  author={Alfarra, Motasem and Itani, Hani and Pardo, Alejandro and Alhuwaider, Shyma Yaser and Ramazanova, Merey and Perez, Juan Camilo and Cai, Zhipeng and M{\"u}ller, Matthias and Ghanem, Bernard},
  booktitle={International Conference on Machine Learning (ICML)},
  year={2024}
}

@inproceedings{schirmer2025monitoring,
  title={Monitoring Risks in Test-Time Adaptation},
  author={Schirmer, Mona and Jazbec, Metod and Naesseth, Christian A. and Nalisnick, Eric},
  booktitle={Advances in Neural Information Processing Systems (NeurIPS)},
  year={2025}
}

@article{prostate158,
  title   = {Prostate158 -- An expert-annotated 3{T} {MRI} dataset and algorithm for prostate cancer detection},
  author  = {Adams, Lisa C. and Makowski, Marcus R. and Engel, G{\"u}nther and Rattunde, Maximilian and Busch, Felix and Asbach, Patrick and Niehues, Stefan M. and Vinayahalingam, Shankeeth and van Ginneken, Bram and Litjens, Geert and Bressem, Keno K.},
  journal = {Computers in Biology and Medicine},
  volume  = {148},
  pages   = {105817},
  year    = {2022}
}

@article{rimonedl,
  title   = {{RIM-ONE DL}: A unified retinal image database for assessing glaucoma using deep learning},
  author  = {Batista, Francisco Jos\'e Fumero and Diaz-Aleman, Tinguaro and Sigut, Jose and Alayon, Silvia and Arnay, Rafael and Angel-Pereira, Denisse},
  journal = {Image Analysis \& Stereology},
  volume  = {39},
  number  = {3},
  pages   = {161--167},
  year    = {2020}
}

@misc{bloch2015nciisbi,
  author       = {Bloch, N. and Madabhushi, A. and Huisman, H. and Freymann, J. and Kirby, J. and Grauer, M. and Enquobahrie, A. and Jaffe, C. and Clarke, L. and Farahani, K.},
  title        = {{NCI-ISBI} 2013 Challenge: Automated Segmentation of Prostate Structures},
  howpublished = {The Cancer Imaging Archive},
  year         = {2015},
  doi          = {10.7937/K9/TCIA.2015.zF0vlOPv}
}

@article{clark2013tcia,
  author  = {Clark, K. and Vendt, B. and Smith, K. and Freymann, J. and Kirby, J. and Koppel, P. and Moore, S. and Phillips, S. and Maffitt, D. and Pringle, M. and Tarbox, L. and Prior, F.},
  title   = {The {Cancer Imaging Archive} ({TCIA}): Maintaining and Operating a Public Information Repository},
  journal = {Journal of Digital Imaging},
  volume  = {26},
  number  = {6},
  pages   = {1045--1057},
  year    = {2013},
  doi     = {10.1007/s10278-013-9622-7}
}

@article{lemaitre2015i2cvb,
  author  = {Lema{\^i}tre, G. and Mart{\'i}, R. and Freixenet, J. and Vilanova, J. C. and Walker, P. M. and Meriaudeau, F.},
  title   = {Computer-Aided Detection and Diagnosis for Prostate Cancer Based on Mono and Multi-Parametric {MRI}: A Review},
  journal = {Computers in Biology and Medicine},
  volume  = {60},
  pages   = {8--31},
  year    = {2015}
}

@article{litjens2014promise12,
  author  = {Litjens, G. and Toth, R. and van de Ven, W. and Hoeks, C. and Kerkstra, S. and van Ginneken, B. and others},
  title   = {Evaluation of Prostate Segmentation Algorithms for {MRI}: The {PROMISE12} Challenge},
  journal = {Medical Image Analysis},
  volume  = {18},
  number  = {2},
  pages   = {359--373},
  year    = {2014},
  doi     = {10.1016/j.media.2013.12.002}
}

\newpage
\appendix
\setcounter{section}{0}
\setcounter{figure}{0}
\setcounter{table}{0}
\renewcommand{\thesection}{S\arabic{section}}
\renewcommand{\thefigure}{S\arabic{figure}}
\renewcommand{\thetable}{S\arabic{table}}
\section{Reproducibility Details}
\label{sec:supp-repro}

\subsection{Compute and Software}
All experiments run on one NVIDIA RTX 6000 48GB;
no cluster or machine identifiers are given here.
Software: Python~3.11.15, PyTorch~2.6.0, CUDA~12.4,
MONAI~1.3.2, nnU-Net~v2.2.1.
Source models are frozen: no offline or persistent target-domain training is performed,
and no adapted weights survive a case (Sec.~\ref{sec:supp-eata}); ``frozen'' refers to the
source checkpoint, not to an absence of gradient steps.
Code and configuration files will be released on publication.

\subsection{Source Models ($M_0$)}
\label{sec:supp-sourcemodels}
\begin{itemize}
\item \emph{Cardiac / prostate MRI:} frozen nnU-Net source model
      (InstanceNorm backbone); no retraining.
\item \emph{Driving (ACDC):} frozen Cityscapes-pretrained SegFormer-b0
      (19 classes); no retraining. Adaptation updates the affine parameters of the encoder \texttt{LayerNorm} modules only ($30$ modules, $7{,}168$ parameters, $0.19\%$ of the network); the single \texttt{BatchNorm} in the decode head is not updated.
\end{itemize}

\subsection{Episodic TTA Protocol}
\label{sec:supp-eata}
Per case: reset to source weights $M_0$, adapt for a fixed budget of
$K$ steps, then reset. Adaptation objective: entropy minimization on
normalization affine parameters (TENT-style; EATA adds entropy filtering with $e_{\mathrm{margin}}{=}0.45$ and an anti-forgetting anchor with weight $1.0$). The margin never binds on the medical benchmarks: at $M_0$ all $2{,}434$ cardiac evaluation slices and all $1{,}420$ slices of the $124$-case prostate pool, a superset of the evaluation split, pass it, the largest slice entropy observed being $22\times$ below it, so those rows are entropy-filtered in configuration only and are effectively unfiltered in effect. The anchor is an \emph{unweighted} $\ell_2$ penalty toward the source weights, $\sum_p \lVert p - p_0\rVert^2$, not the Fisher-weighted form of the original EATA; we keep the symbol $\lambda_{\mathrm{Fisher}}$ for the configuration key but the term it multiplies carries no Fisher information matrix. This differs from the published EATA regularizer, and uniform anchoring was used throughout, so our EATA rows should be read as an entropy-filtered, source-anchored baseline rather than as a reproduction of that method. Baseline configurations: MEMO uses $4$ augmentations with consistency weight $1.0$; our episodic CoTTA variant uses $2$ light augmentations, teacher EMA decay $0.999$, stochastic restore probability $0.01$ and consistency weight $10.0$. Optimizer: Adam (default $\beta$,
no weight decay), learning rate as below, batch size $1$
(episodic, one case at a time).
Main tables use a deterministic episodic reset; CRC / multi-seed
stability sweeps use seeds $\{0,\ldots,4\}$ where reported.

\paragraph{Learning rates / budgets used.}
\begin{itemize}
\item Medical (prostate/cardiac): lr $5{\times}10^{-4}$, $K{=}4$
      (fixed-4 baseline); controllers adapt within the same $K{=}4$
      budget (mean steps ${<}4$). Main-table fixed-$1/2/3$ are the
      step-$k$ checkpoints of the same episodic EATA fixed-4
      trajectory (no separate re-run).
\item ACDC mild budget: lr $10^{-4}$, $K{=}4$.
\item ACDC aggressive budget: lr $2{\times}10^{-3}$, $K{=}8$.
\item Budget-axis sweep (Table~\ref{tab:supp-acdc-lr}, Sec.~\ref{sec:supp-lrsweep}):
      lr $\in\{10^{-4}, 5{\times}10^{-4},
      10^{-3}, 2{\times}10^{-3}\}$.
\end{itemize}

\subsection{Budget Selection and Fixed-$k$ Ladders}
\label{sec:supp-ladders}
This section holds the full budget-axis evidence behind the main-text ladders: the
holdout$\to$eval transfer experiments on both medical splits, and the complete eight-step
ACDC rain ladder that the main text abridges to $k{\in}\{1,2,4,8\}$.

\paragraph{Fixed-$k$ holdout$\to$eval transfer (prostate OOD-all).}
Using the same per-site $20\%$ holdout as the main tables
($n_{\mathrm{holdout}}{=}23$, $n_{\mathrm{eval}}{=}93$) and the
EATA fixed-4 trajectory checkpoints:
\begin{itemize}
\item Holdout mean HA by $k$: $0.148$ / $0.169$ / $0.206$ / $0.287$
      for $k{=}1{..}4$ (argmin $k{=}1$).
\item Eval mean HA by $k$: $0.242$ / $0.211$ / $0.171$ / $0.262$
      (argmin $k{=}3$).
\item Every standard selection rule on the holdout
      (min HA, max Dice, HA${-}0.5$Dice, HA${-}$Dice) picks $k{=}1$;
      none recovers the eval optimum.
\item Deploying holdout-selected $k{=}1$ on eval incurs HA regret
      $+0.070$ vs.\ the eval oracle $k{=}3$.
\item Case bootstrap ($B{=}5000$): $P($holdout selects $k{=}1){=}0.68$,
      $P($selects $k{=}3){=}0.07$; joint resampling gives
      $P(k_{\mathrm{holdout}}{=}k_{\mathrm{eval}}){=}0.071$, with mass on
      the pair $(1,3)$.
\item Per-site argmin HA agrees on only $3/6$ sites between holdout
      and eval.
\end{itemize}
Paired bootstrap of the controllers against these budgets ($B{=}5000$, seed $1000$)
is collected in Sec.~\ref{sec:supp-pairwise}, which states its own scope.

\paragraph{Fixed-$k$ holdout$\to$eval transfer (M\&Ms vendor~B).}
Holdout is the official vendor-B calib split ($n{=}20$;
\texttt{vendorB\_calib.txt}); eval is B-test ($n{=}230$).
Overlap is empty. Holdout uses a fresh EATA fixed-4 run on B-calib;
eval reuses the archived B-test fixed-4 trajectory checkpoints:
\begin{itemize}
\item Holdout mean HA by $k$: $0.084$ / $0.046$ / $0.070$ / $0.221$
      for $k{=}1{..}4$ (argmin $k{=}2$).
\item Eval mean HA by $k$: $0.186$ / $0.164$ / $0.129$ / $0.177$
      (argmin $k{=}3$).
\item Eval mean Dice by $k$: $0.8506$ / $0.8504$ / $0.8497$ / $0.8490$
      (spread $0.002$, the Dice range quoted in Table~\ref{tab:budget-sweep}'s caption).
\item Min-HA, max-Dice, and HA${-}0.5$Dice on the holdout all pick
      $k{=}2$; none recovers the eval optimum $k{=}3$.
\item Deploying holdout-selected $k{=}2$ on eval incurs HA regret
      $+0.035$ vs.\ the eval oracle $k{=}3$.
\item Case bootstrap ($B{=}5000$, seed $1000$):
      $P($holdout selects $k{=}2){=}0.64$,
      $P($selects $k{=}1){=}0.36$,
      $P($selects $k{=}3){=}0$;
      joint resampling gives
      $P(k_{\mathrm{holdout}}{=}k_{\mathrm{eval}}){=}0.003$,
      with mass on the pair $(2,3)$.
\end{itemize}
Paired bootstrap on B-test is in Sec.~\ref{sec:supp-pairwise}.

\paragraph{Fixed-$k$ ladder on ACDC rain (TENT, lr $2{\times}10^{-3}$).}
Extracted from the archived val trajectory checkpoints
($n{=}100$), identical to the TENT$_8$ run the main text reports:
\begin{itemize}
\item Mean HA by $k{=}1{\ldots}8$: $0.264$ / $0.327$ / $0.358$ /
      $0.385$ / $0.408$ / $0.422$ / $0.439$ / $0.448$ (strictly
      monotone; argmin $k{=}1$).
\item Mean mIoU: $0.378$ / $0.385$ / $0.387$ / $0.389$ / $0.389$ /
      $0.389$ / $0.390$ / $0.387$ (peaks at $k{=}7$, dips at $k{=}8$).
\item Train$\to$val min-HA selection of $k$: both pick $k{=}1$;
      bootstrap $P(k_{\mathrm{train}}{=}k_{\mathrm{val}}){=}1.0$
      ($B{=}5000$; $n_{\mathrm{train}}{=}400$).
\end{itemize}
Controller-vs-budget intervals on this ladder are in Sec.~\ref{sec:supp-pairwise}.

\subsection{Fragmentation Signal}
\label{app:stoprule}
Computed from predicted masks only (no labels, no gradients at
decision time): number of disagreement connected components $n_{\mathrm{reg}}$,
disagreement ratio $\delta$, their step deltas, and mean predictive
entropy on disagreement voxels. Connected components use
\texttt{scipy.ndimage.label} default face-connectivity
(4-connected in 2D; 6-connected in 3D);
minimum region size $16$ voxels (medical) / $200$\,px (ACDC). Mask convention: on the
medical benchmarks $M_0$ and $M_k$ are binarized to foreground (any positive label)
before the disagreement mask is formed, so only foreground/background edits count;
on ACDC the $19$-class label maps are compared directly, so any label change counts.
HA (Sec.~\ref{sec:prelim}) uses the same masks, with GT binarized the same way on the
medical benchmarks.

\subsection{Signal--Harm Association Across Domains}
\label{sec:supp-signal-crossdomain}
The cross-benchmark rows of main-text Table~\ref{tab:signal-ha} use the protocol stated
there: Spearman correlations pooled over
$(\text{case},\text{step})$ pairs with step ${\geq}1$ (step $0$ is excluded, since
$\mathrm{HA}{=}0$ there by construction), and a $\Delta$ column that differences consecutive
steps within a case. Apart from the verification-rerun rows marked in its caption and the GraTa-cos row (Sec.~\ref{app:rerun}), no new runs were performed for it; every other value is computed from the per-step statistics already stored for the main-table trajectories. The same statistics give
the signal--signal correlations the main text quotes for complementarity: on prostate, over
the same step-${\geq}1$ pairs, $n_{\mathrm{reg}}$ correlates $+0.034$ with mean entropy and
$-0.101$ with mean MSP ($\delta$: $+0.035$ and $-0.113$). Clustered intervals, the
per-condition and per-step breakdowns, and the attribution of the aggressive-budget collapse
to update size rather than trajectory depth are all in Sec.~\ref{app:ci}, which is the single
source for those numbers.

\subsection{Calibration Protocol}
\label{app:calib}
\label{sec:supp-calib}

\begin{table}[t]
\centering
\small
\setlength{\tabcolsep}{4pt}
\begin{tabular}{lllll}
\toprule
Domain & Coordinate & Low & Mid & Hard \\
\midrule
Prostate & $n_{\mathrm{reg}}$ & depth 1 & depth 1 & rollback \\
M\&Ms & $\delta$ & depth 3 & depth 2, then shrink ($\alpha{=}0.5$) & rollback \\
ACDC & $\delta$ & stop rule, cap 8 & stop rule, cap 4 & rollback \\
\bottomrule
\end{tabular}
\caption{Routing coordinate and per-bucket action by domain. Cut-points are tertile percentiles of the
step-1 coordinate on the calibration split, with the exact convention differing between
the M\&Ms and driving pipelines (Sec.~\ref{app:calibsens}); prostate: per-site $20\%$ OOD holdout,
$n{=}23$; M\&Ms: vendor-B calib, $n{=}20$; ACDC: per-condition train. The two medical routers deploy a fixed depth
per bucket, so low and mid coincide on prostate and differ by one step and the shrink action on M\&Ms; the driving router runs the parameter-free $\Delta n_k$ stop rule of
Sec.~\ref{sec:controllers} within a per-bucket cap (the calibrated $\tau$-stop is a separate controller), and its depth therefore varies by case (Table~\ref{tab:supp-depth-var}).}
\label{tab:supp-routing-coord}
\end{table}

\begin{table}[t]
\centering\small
\begin{tabular}{llccl}
\toprule
Domain & Bucket & $n$ & std & range \\
\midrule
M\&Ms & low / mid & 54 / 46 & 0.00 / 0.00 & 3 / 2 \\
Prostate & low / mid & 27 / 24 & 0.00 / 0.00 & 1 / 1 \\
\midrule
ACDC fog & low / mid & 52 / 30 & 0.64 / 0.66 & 1--5 / 1--4 \\
ACDC rain & low / mid & 38 / 41 & 0.37 / 0.64 & 1--2 / 1--4 \\
ACDC snow & low / mid & 31 / 40 & 0.77 / 0.99 & 1--4 / 1--4 \\
ACDC night & low / mid & 22 / 41 & 2.56 / 1.30 & 1--8 / 1--4 \\
\bottomrule
\end{tabular}
\caption{Within-bucket dispersion of deployed depth. The four medical buckets are
single-valued; all eight driving buckets are not. This is the measurement behind the
distinction drawn in Table~\ref{tab:supp-routing-coord}: on the medical benchmarks the
fragmentation state enters the bucket assignment and not the depth. The dispersion is
largest on night, where $M_0$ is near-collapsed and $\Delta n_k$ has no stable rise
point (Sec.~\ref{sec:acdc}).}
\label{tab:supp-depth-var}
\end{table}
z-score statistics and the
stop threshold $\tau$ are fit \emph{once} on a held-out labeled split
(per-site for medical, ACDC train split for driving; sizes in
Table~\ref{tab:supp-routing-coord}), disjoint from the
online test stream. Which statistics are needed depends on the controller.
FragSubset scores with the two-term risk of Eq.~\ref{eq:risk} and therefore fits both
pairs on the medical benchmarks; the ACDC $\tau$-stop scores with
$r_k{=}z(n_k)$ alone, so $(\mu_\delta,\sigma_\delta)$ do not enter there. The routers
use no z-scores at all: they split a raw per-domain coordinate at tertiles of the
calibration split (Table~\ref{tab:supp-routing-coord}). FragStop is parameter-free.
Fitted values: ACDC full-train $\mu_n{=}20.1$, $\sigma_n{=}6.8$,
$\tau{=}{-}0.16$; prostate holdout
$\mu_n{=}14.87$, $\sigma_n{=}21.20$, $\mu_\delta{=}0.0068$,
$\sigma_\delta{=}0.0113$; M\&Ms calibration split
$\mu_n{=}0.70$, $\sigma_n{=}0.78$, $\mu_\delta{=}0.00028$,
$\sigma_\delta{=}0.00020$. No labels or
gradients are used at decision time.

\paragraph{COQR router.} Case-level online quantile routing makes only the routing decision
at test time, from tertile thresholds of the fragmentation score. On
prostate OOD-all ($n{=}93$) it routes $27$ cases low,
$24$ mid and $42$ hard: low and mid cases each deploy at depth 1 by bucket policy, not by a stop-rule trigger (only $3$ of the $51$ satisfy $\Delta n_2 > 0$), and hard cases are rolled back to $M_0$. The hard bucket is
also where the source model is weakest (mean $M_0$ Dice $0.578$,
against $0.869$ in the two adapting buckets). Because the cut-points are
percentiles \emph{of the calibration split}, the bucket shares on the
evaluation set need not be even, and on both medical benchmarks they are
not: $45.2\%$ of prostate cases and $56.5\%$ of M\&Ms cases land in the
hard bucket. A near-even split would in fact be the suspicious outcome,
since it is what fitting the percentiles on the evaluation pool itself
would produce. What decides whether the per-case decision is worthwhile is
not how evenly the buckets fill but the helped/hurt composition of the
retained bucket (main text). We report the router on OOD-all
rather than on the $n{=}34$ OOD-hard subset because the deployed cut-points put $28$ of the $34$ in the hard bucket and leave the low bucket empty, so there are no per-bucket means to report. On ACDC the router uses per-condition cut-points, since
shift severity differs sharply across conditions; low and mid bucket sizes are in
Table~\ref{tab:supp-depth-var}, the hard buckets hold $18$ (fog), $21$ (rain), $29$ (snow) and
$43$ (night) cases, and the actions are in Table~\ref{tab:supp-routing-coord}.

\paragraph{Conformal gate.} $\hat\lambda$ is selected on the calibration split via
mean-risk conformal risk control, with a fixed-denominator surrogate.

\paragraph{Routing coordinate by domain.} Table~\ref{tab:supp-routing-coord} gives the
instantiation. This qualifies what ``transfers'' means: the M\&Ms$\rightarrow$ACDC port keeps the routing coordinate, refits the tertiles and replaces the fixed depths with the stop rule under per-bucket caps, whereas the prostate router uses a different coordinate of the same fragmentation state $s_k$.

\paragraph{What ``seed-only'' means.} On M\&Ms and ACDC the cut-points are
percentiles of the step-1 scores of the calibration cases only, $33.33$/$66.67$ on M\&Ms and $33$/$66$ on ACDC (\emph{seed-only}: the seed pool is never extended with test
cases and no online sliding update is applied). The implementation also offers a \emph{hybrid} mode that
appends each observed test case to the pool and refits the percentiles over a sliding
window; we do not use it, because it would make the thresholds a function of the test
stream. Seed-only is the setting reported everywhere in this paper. On ACDC the cut-points are
fitted per condition, which assumes the condition is known at deployment; a single pooled
threshold would be the honest alternative for a system that must infer it, and we did not
run it. The pools are the calibration splits of Table~\ref{tab:supp-routing-coord}; on prostate the
step-1 $n_{\mathrm{reg}}$ tertiles are $4.0$ and $10.7$.

\paragraph{Stability of the cut-points.} Because the buckets are fixed offline,
their sensitivity to the calibration sample matters, and that sample is small.
Bootstrapping the step-1 $\delta$ pool of the $n_{\mathrm{calib}}{=}20$ M\&Ms
calibration split ($5000$ resamples) gives $\tau_{\text{lo}}=0.000169$ $[0.000107, 0.000216]$ and
$\tau_{\text{hi}}=0.000263$ $[0.000191, 0.000559]$, and the induced bucket
assignment on the evaluation set changes for a median of $15.2\%$ of cases
(upper tail ${\approx}38\%$). The cut-points are therefore the least stable part of the controller. That is
the price of calibrating on twenty cases, and it is a price we pay deliberately:
the only larger pool available is the evaluation stream itself, which seed-only
calibration rules out. We
would rather report a wide resampling interval than a narrow one obtained that
way. The same constraint bounds the certified gate at small $n_{\mathrm{cal}}$
(Sec.~\ref{sec:supp-crc-op}), so calibration-set size is a real limit on this
pipeline rather than an artifact of one experiment. The consequences of that perturbation for deployed Dice and HA are propagated in Sec.~\ref{app:calibsens}; neither analysis re-runs adaptation under new seeds.

\paragraph{How steps are counted.} Across all domains the reported
\emph{Steps} column is the deployed depth of the state actually
returned: a case routed hard is probed for one step, reset to $M_0$ and
recorded as $0$ steps (\texttt{rolled\_back}$=$1), while low and mid cases
are recorded at the step where they stop. One qualification: on M\&Ms the mid
bucket shrinks rather than merely halting, so its $46$ cases deploy the weights
interpolated half-way back to $M_0$ ($\alpha{=}0.5$) after their second step.
That state is not any $M_k$ checkpoint, and we record it at the step it was
taken from ($2$). The \emph{Steps} column is therefore the depth along the
trajectory at which the deployed state was formed, not a count of updates
surviving in the returned weights; for every method except the M\&Ms mid bucket
the two coincide. The $|D|$ used for its HA is computed against the prediction
actually deployed from the interpolated weights, consistent with the
$M_0$-referenced convention of the main text. The definition is identical in
the M\&Ms and driving pipelines. Differences between domains therefore
reflect the policies attached to each bucket, not different accounting:
on M\&Ms the mean of $1.10$ comes from $130$ hard cases at $0$ steps, $46$
mid at $2$ and $54$ low at $3$; on prostate OOD-all the mean of $0.55$ comes from
$42$ hard at $0$ steps and $51$ low or mid cases that each stop at $1$; on ACDC the per-condition means are
$0.96$ (fog), $0.97$ (rain), $1.01$ (snow) and $1.67$ (night), the last
being higher because low and mid cases there run longer ($3.23$ and
$2.59$ steps) before the rule halts them. The $\tau$-stop controller uses
a different convention and is not comparable on this column: it never
rolls back and is recorded with a floor of one step, so its rain mean is
$2.49$ rather than ${\sim}1$.

\subsection{Baselines}
\label{sec:supp-baselines}
The four stop rules compared in Sec.~\ref{app:defenses} share one parameter-free form, halting at the first step at which the statistic worsens relative to the previous step, and differ only in the statistic ($n_{\mathrm{reg}}$ rising, mean entropy rising, MSP falling,
gradient norm rising). None of them uses a calibrated threshold, so that comparison is
controlled by construction; the calibrated $\tau$-stop of the ACDC section is a separate
controller. As adaptation objectives we wrap
TENT, EATA (fixed-$K$), MEMO, and an episodic light-augmentation CoTTA
variant ($n_{\mathrm{aug}}{=}2$), all under the same episodic reset and
the same $M_0$-referenced HA. GraTa's alignment enters
Table~\ref{tab:signal-ha} only as a read-only signal; the full objective is in
Sec.~\ref{app:newobj}.

\subsection{Evaluation and Statistics}
\label{sec:supp-eval}
Split sizes: prostate OOD-hard $n{=}34$, OOD-all $n{=}93$, pooled
trajectories $n{=}124$ (the $93$ evaluation cases plus the $23$ per-site holdout cases and
$8$ source-domain G-val cases, i.e.\ every case with a stored EATA trajectory; used only for
the signal--HA correlations, never for a deployment number), HA--HD95 cohort $n{=}34$ (the
OOD-hard split). The prostate
OOD-hard \emph{pool} is exactly sites B and E ($30{+}12{=}42$ cases); the $34$ of them that
fall in the evaluation split are the OOD-hard split reported in the tables, and the remaining
$8$ B/E cases fall in the holdout (a different set from the $8$ G-val cases above). We call the full $42$ the \emph{B/E signal-analysis pool}: it is used only for
descriptive signal analysis (the region-geometry contrast of Sec.~\ref{sec:supp-edit-geometry} is drawn from these sites), never to fit a deployed
threshold. All deployed thresholds come from
the per-site $20\%$ holdout (Table~\ref{tab:supp-routing-coord}). M\&Ms vendor-B
$n{=}230$; ACDC val per condition $n{=}100$ for fog/rain/snow and $n{=}106$ for night (the dataset's own split). Every method in the prostate
OOD-all panel is run on the identical case set, so the router,
the fixed-budget baseline and the external baselines are directly
comparable there.

\paragraph{Pooling conventions.} Two distinct senses of ``pooled'' appear in this paper
and are unrelated. Figure~\ref{fig:teaser}(A) reports mIoU micro-averaged over all pixels of
the condition, which puts $M_0$ at $0.385$ on ACDC rain, whereas every table reports the
per-image mean, which puts it at $0.364$ (Table~\ref{tab:supp-acdc-fogsnow}); the panel is
there for the shape of the two curves, and its $y$ values are not comparable with the tables.
Elsewhere ``pooled'' refers to case sets (the $n{=}124$ prostate signal-analysis pool above) and to correlations pooled over (case,\,step) pairs, neither of which concerns
pixel averaging.

\paragraph{Table conventions.} Bold marks the best value among those compared within a
block (highest for Dice, mIoU, $\rho$ and AUROC, lowest for HA and deployed steps) and is
omitted where the text reports no separation. Interval tables carry no bold: the
comparison is made by the intervals themselves. Structural bold (row names, algorithm keywords) is not a comparison. One table departs
from the rule and says so in its caption: Table~\ref{tab:helphurt} bolds the larger of
helped and hurt on each router row, which on prostate is the worse of the two.

\begin{figure}[t]
\centering
\includegraphics[width=\columnwidth]{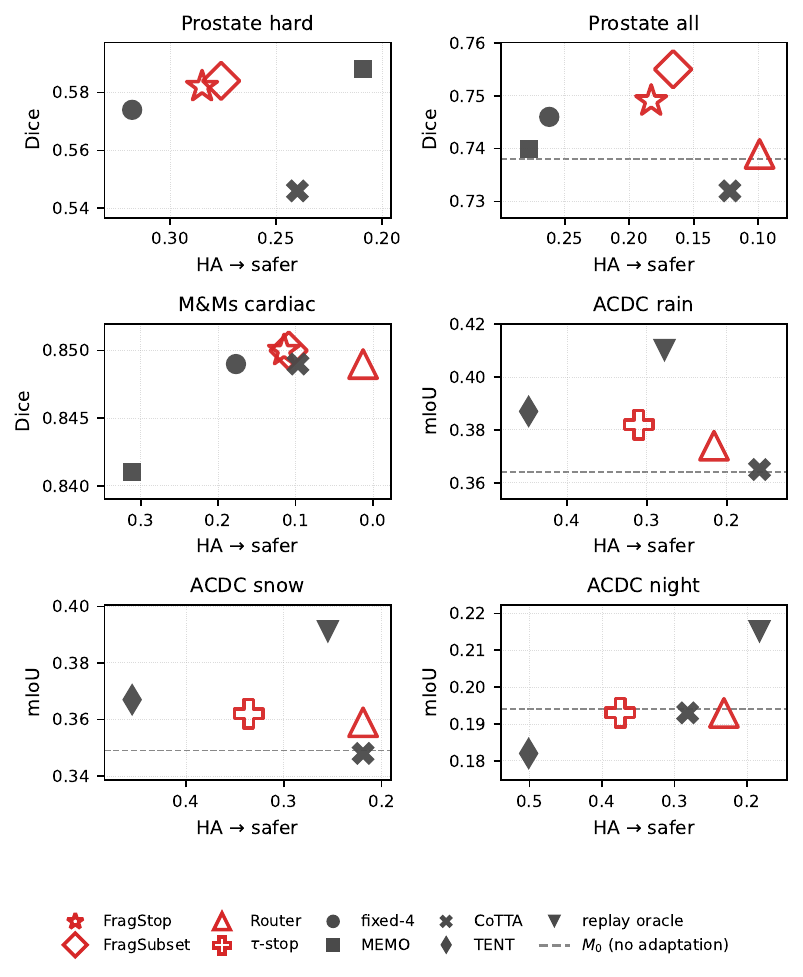}
\caption{Reliability--accuracy trade-off (Tier~C, descriptive) across six deployments. $x$: HA, axis inverted (right $=$ safer); $y$: Dice or mIoU (never shared); colour marks ours (red) against baselines (grey) and shape identifies the method, as in the shared legend; marker size encodes nothing (ours are drawn larger for legibility), so deployed steps are read from the tables rather than from the figure; dashed $=M_0$ where a no-adaptation row is available (four of the six panels). The router is absent from prostate OOD-hard (an empty low bucket under the deployed cut-points); the replay oracle (per-case best-mIoU step, Sec.~\ref{sec:supp-pairwise}) appears only on the three ACDC panels where those trajectories are stored. Reading of the panels: Sec.~\ref{sec:supp-eval}.}
\label{fig:pareto}
\end{figure}

Within the deployed pipeline, labels are used offline in exactly four places:
(1)~to select the reference budget on the holdout;
(2)~to fit the stop threshold $\tau$ on the calibration split (a quantile of the
fragmentation risk at the \emph{oracle} early-stop step, which needs the metric);
(3)~to fit $\hat\lambda$ for the conformal gate; and
(4)~to score HA and Dice\,/\,mIoU after the fact; the design choices made with an
evaluation split visible are a separate exposure, disclosed in Sec.~\ref{app:algo}. The fragmentation statistics themselves, $(\mu_n,\sigma_n)$, $(\mu_\delta,\sigma_\delta)$ and the routing cut-points, are computed from predicted masks on those same splits and use no labels. Nothing
of any kind is fit inside the adaptation loop or the per-case decision. Paired bootstrap with
$5000$ resamples for all confidence intervals; the tier labels A / A$'$ / B / C
used throughout are the ones defined in the main-text footnote (deployable
budget / default fixed-4 / retrospective eval-best $k$ / descriptive); the Tier-C
positioning is Fig.~\ref{fig:pareto}. Read panel by panel: our controllers move to the safer
side of \emph{their own budget reference} in every panel, and on ACDC the ordering
TENT$\rightarrow\tau$-stop$\rightarrow$router moves rightwards on all conditions (fog is not
shown; Table~\ref{tab:supp-acdc-fogsnow}); on prostate OOD-hard MEMO and CoTTA sit safer
than our controllers, at the accuracy cost documented in the main text; on prostate OOD-all
the router is the safest point shown but sits on the $M_0$ Dice line, the boundary
Sec.~\ref{sec:prostate} discusses and Table~\ref{tab:helphurt} decomposes.
Holm correction is applied in exactly one place: the ordering claim
\emph{among our own controllers} (modulation beyond stopping, FragSubset vs.\ FragStop
on prostate OOD-hard). Comparisons against the retrospective budget are
reported as single paired-bootstrap intervals without correction, because each
is a pre-specified comparison against one named reference rather than a member
of a searched family. The ACDC intervals and
$P(\Delta{<}0)$ values are computed per comparison from the paired
bootstrap and are reported \emph{uncorrected}; with four conditions and two
reference methods they should be read as describing each condition rather
than as a family-wise test. We flag this because the ACDC HA reductions are
large and consistent enough that no correction would change their sign,
whereas the small mIoU differences there are exactly the quantities a
correction would attenuate.

\subsection{HA--HD95 Construct-Validity Check}
\label{app:hd95}
\begin{figure}[h]
\centering
\includegraphics[width=\linewidth]{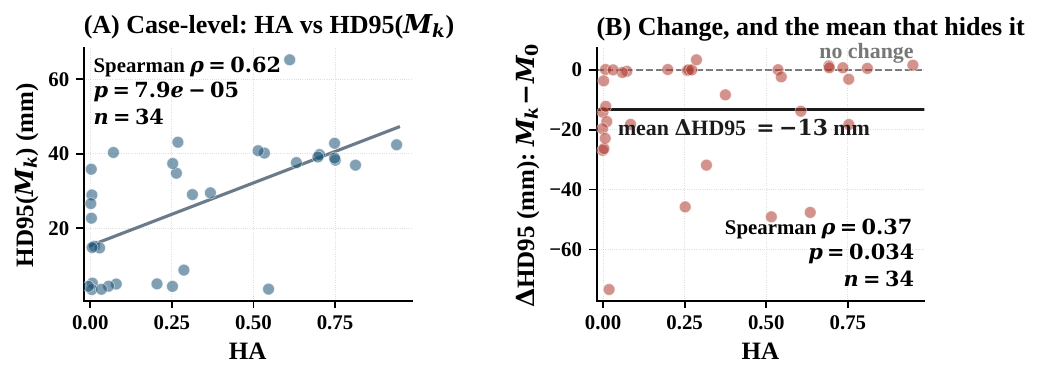}
\caption{\textbf{HA vs.\ a clinically used boundary metric} (prostate OOD-hard, EATA fixed-4,
$n{=}34$; lower is better everywhere). (A)~vs.\ HD95($M_k$), $\rho{=}0.62$; the straight line is an
ordinary least-squares fit shown as a guide. (B)~vs.\ $\Delta$HD95
($\rho{=}0.37$): black$=$cohort mean (${-}13$\,mm), dashed$=$no change; the mean improves while
HA accumulates.}
\label{fig:ha-hd95}
\end{figure}

Boundary metric: HD95 (medpy~0.5.2) on the same masks used for HA
(prostate OOD-hard, EATA fixed-4, $n{=}34$). The fixed-4 mask dump covers
$50$ cases (all of sites B and E plus eight source-site G cases), and we
restrict the check to the $34$ that are in the OOD-hard evaluation split, so
the cohort matches the split reported in the prostate tables.
Spearman $\rho(\mathrm{HA},\mathrm{HD95}_{M_k}){=}0.62$ ($p{=}7.9{\times}10^{-5}$);
$\rho(\mathrm{HA},\Delta\mathrm{HD95}){=}0.37$ ($p{=}0.034$);
cohort-mean $\Delta\mathrm{HD95}{=}{-}13$\,mm.

The same check on cardiac uses the Sec.~\ref{app:rerun} rerun masks (all $230$ vendor-B
evaluation cases, EATA fixed-$4$, per-case spacing from the image headers, foreground-binarized
as HA is). HA tracks the \emph{change} in boundary error at $\rho{=}0.34$
($p{=}1.0{\times}10^{-7}$) and the adapted boundary error itself at $\rho{=}0.27$
($p{=}4.6{\times}10^{-5}$), against $0.37$ and $0.62$ on prostate. The weaker level correlation
is what the sparsity of cardiac edits predicts: HA is zero on $144$ of the $230$ cases and HD95
does not move on $173$, so most of the cohort is tied at the origin; among the $30$ cases where
both quantities move, $\rho(\mathrm{HA},\Delta\mathrm{HD95})$ is $0.78$. The level correlation is weaker because the edit set is smaller:
at fixed-$4$ the cardiac disagreement covers $\delta{=}3.3{\times}10^{-4}$ against
$5.6{\times}10^{-3}$ on prostate (Sec.~\ref{app:accounting}), so HA sits at zero on most cardiac
cases and cannot order their absolute boundary error. The dissociation also takes a different
form. On prostate the cohort mean improves by $13$\,mm while HA accumulates; on cardiac it
barely moves (${-}0.60$\,mm, median $0$) while HA still has per-case structure. That is the
shape of Sec.~\ref{sec:budget}'s reading of $\Delta$Dice on this benchmark, where the cohort
mean is indistinguishable from zero while $58.7\%$ of cases are individually worse: on cardiac
both a region metric and a boundary metric average to nothing over a cohort that contains harm.

\subsection{Region-Level Geometry of Harmful vs.\ Beneficial Edits}
\label{sec:supp-edit-geometry}
The main text claims fragmentation as a \emph{case-level} signal and explicitly does
not claim that an individual harmful edit is geometrically identifiable. This section
is the evidence for that disclaimer. On the same EATA fixed-4 dumped masks we label
every disagreement component as harmful or beneficial by the HA definition of the
main text (identical connectivity and $\tau_{\mathrm{area}}{=}16$; neutral
components, where $M_k$ and $M_0$ are correct on equally many voxels, are $0.1\%$ of
regions on prostate and $0.4\%$ on M\&Ms and are dropped) and test the natural
region-level reading of the signal: that harmful edits are smaller, less compact,
more often tiny islands, and more detached from the bulk of the disagreement.

\begin{table}[h]
\centering
\small
\setlength{\tabcolsep}{4pt}
\begin{tabular}{lcccc}
\toprule
Feature & mean H & mean B & $p$ & Cliff's $d$ \\
\midrule
\multicolumn{5}{l}{\emph{Prostate OOD-hard} ($n{=}34$ cases; $218$ H, $524$ B)} \\
area (vox)          & $614.7$  & $405.5$  & $0.054$              & $-0.09$ \\
compactness $A/S^2$ & $0.0279$ & $0.0246$ & $0.053$              & $+0.09$ \\
dist.\ to largest   & $50.7$   & $89.5$   & $8{\times}10^{-23}$  & $-0.46$ \\
frac.\ of $|D|$     & $0.049$  & $0.031$  & $4{\times}10^{-7}$   & $+0.24$ \\
small @$5\%$        & $0.789$  & $0.903$  & $3{\times}10^{-5}$   & $-0.11$ \\
\midrule
\multicolumn{5}{l}{\emph{M\&Ms vendor-B} ($n{=}230$ cases; $144$ H, $139$ B)} \\
area (vox)          & $159.4$  & $165.5$  & $0.94$  & $-0.00$ \\
compactness $A/S^2$ & $0.0251$ & $0.0248$ & $0.94$  & $+0.00$ \\
dist.\ to largest   & $14.7$   & $19.7$   & $0.031$ & $-0.14$ \\
frac.\ of $|D|$     & $0.288$  & $0.246$  & $0.15$  & $+0.10$ \\
small @$5\%$        & $0.153$  & $0.209$  & $0.22$  & $-0.06$ \\
\bottomrule
\end{tabular}
\caption{Region-level contrast between harmful (H) and beneficial (B) edits, EATA
fixed-4. Mann--Whitney $U$ (two-sided); Cliff's $d$ is $\mathrm{H}-\mathrm{B}$, so
a negative $d$ means harmful regions score \emph{lower}. We write it $d$ rather than
$\delta$ to keep it distinct from the disagreement ratio $\delta$ used throughout. ``small @$5\%$'' is the
fraction of regions below $5\%$ of $|D|$. }
\label{tab:supp-edit-geometry}
\end{table}

That reading is not supported (Table~\ref{tab:supp-edit-geometry}). On prostate three
contrasts reach significance, and the two large ones run the \emph{opposite} way:
harmful regions sit much closer to the
largest disagreement component ($d{=}{-}0.46$) and take a larger share of $|D|$
($d{=}{+}0.24$), while it is the \emph{beneficial} edits that are more often tiny
islands (small@$5\%$ $0.90$ vs.\ $0.79$; the gap is stable at $2\%$, $5\%$ and $10\%$
thresholds). Area and compactness do not separate the two classes at $\alpha{=}0.05$.
On M\&Ms the contrast is essentially null, with only distance-to-largest weakly echoing prostate ($p{=}0.031$), so the one prostate effect that is large does not even
replicate in strength across datasets. Mechanistically, harmful edits on prostate look
like attached over-growth of the main disagreement mass rather than a swarm of debris.

None of this weakens the case-level claim: a rising
$n_{\mathrm{reg}}$ says the update produced many mutually unrelated changes, which
predicts case-level HA (main-text signal--HA table) without requiring that any single
harmful region be smaller or more scattered than any single beneficial one. It does
rule out a region-level gate built on the same geometry, because an edit-wise filter keyed to ``small and detached'' would preferentially discard the beneficial edits. That is why our controllers accept or reject at the case level. Per-region features and the plotting code will be released on publication.

\subsection{Case-Level Help/Hurt Decomposition}
\label{sec:supp-help-hurt}
For each case we compute $\Delta = \mathrm{metric}(M_k) - \mathrm{metric}(M_0)$ (Dice on
medical, mIoU on ACDC): a case is \emph{rolled back} if the router locked the deployed
output to $M_0$, and otherwise \emph{helped} ($\Delta{>}0$) or \emph{hurt} ($\Delta{<}0$).
Classification is by decision, not by a numerical tolerance. This matters on
prostate, where a rollback re-runs 3D sliding-window inference instead of
reusing the cached $M_0$ and leaves differences of order $5{\times}10^{-5}$ per
voxel; an $|\Delta|$ threshold would mis-count those cases, whereas the routing
itself is exact ($\texttt{rolled\_back}=1$, step $0$, HA $0$ for all $42$ hard
cases). Under the semantic definition the prostate rolled-back share
($45.2\%$) matches its hard bucket exactly, and the ACDC and M\&Ms figures are
unchanged. The uncontrolled reference differs by domain, matching each main
table: fixed-4 on M\&Ms and prostate, TENT$_8$ on ACDC. Baselines have no
rollback action, so their rolled-back share is $0$ by construction.

\begin{table}[h]
\centering
\setlength{\tabcolsep}{3pt}
\small
\begin{tabular}{llcccc}
\toprule
Split & Method & helped ($\mu_{\Delta^+}$) & hurt ($\mu_{\Delta^-}$) & rolled back & overall $\mu_\Delta$ \\
\midrule
ACDC fog & TENT$_8$ & 52.0 ($+0.054$) & 48.0 ($-0.034$) & --- & $+0.012$ \\
ACDC fog & Router & 44.0 ($+0.023$) & 38.0 ($-0.009$) & 18.0 & $+0.007$ \\
ACDC rain & TENT$_8$ & 69.0 ($+0.047$) & 31.0 ($-0.031$) & --- & $+0.023$ \\
ACDC rain & Router & 59.0 ($+0.020$) & 20.0 ($-0.012$) & 21.0 & $+0.010$ \\
ACDC snow & TENT$_8$ & 61.0 ($+0.050$) & 39.0 ($-0.033$) & --- & $+0.018$ \\
ACDC snow & Router & 46.0 ($+0.027$) & 25.0 ($-0.009$) & 29.0 & $+0.010$ \\
\bottomrule
\end{tabular}
\caption{Case-level decomposition of $\Delta$ against $M_0$ on the three remaining
ACDC conditions (percentages of all cases in the condition); M\&Ms, prostate and
ACDC night are in Table~\ref{tab:helphurt}. The router lowers the hurt fraction on
each of these three as well, and on all three it retains more helped cases than hurt
ones. On prostate a further $4.3\%$ of cases adapt but leave the prediction
numerically unchanged and fall into none of the three classes; an exactly zero $\Delta$ for an adapted case does not occur on M\&Ms.}
\label{tab:supp-helphurt}
\end{table}

\subsection{Qualitative ACDC Example}
\label{sec:supp-acdc-qual}

\begin{figure}[H]
\centering
\includegraphics[width=\textwidth]{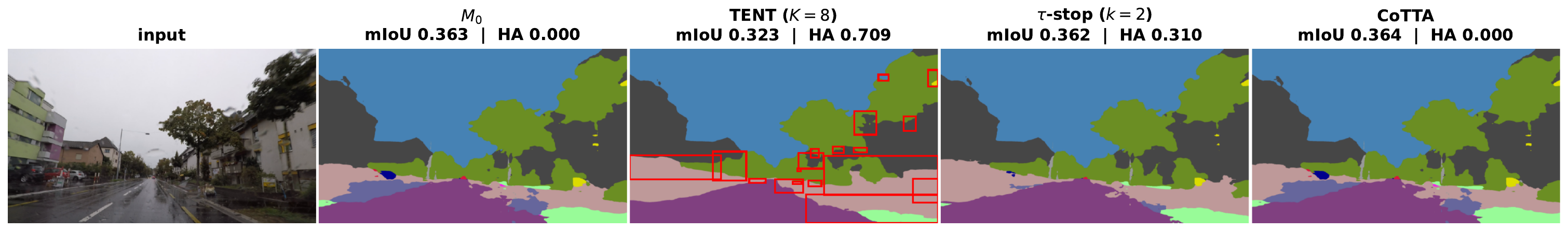}
\caption{Harm invisible to mIoU (ACDC rain, \texttt{GP020402\_frame\_000748}).
L$\rightarrow$R: input, $M_0$, TENT ($K{=}8$; every harmful region above the
$200$\,px threshold is boxed in red), the $\tau$-stop of Sec.~\ref{sec:controllers} (ours),
CoTTA. TENT \emph{lowers} mIoU ($0.363{\to}0.323$) while HA
explodes to $0.709$; $\tau$-stop stops near the mIoU peak ($k{=}2$), keeping
mIoU $0.362$ at HA $0.310$ (this case's HA coincides with the rain cohort mean of
Table~\ref{tab:supp-acdc-fogsnow} to three decimals, $0.3101$ against $0.3103$; its mIoU
does not); episodic CoTTA barely adapts (output
$\approx M_0$, mIoU $0.364$).  Unlike the medical qualitative rows, whose masks are
replayed from the archived dumps, this panel recomputes the TENT, $\tau$-stop and CoTTA
predictions for this image, so reproducing it needs a forward pass rather than the stored
trajectories.}
\label{fig:acdc-qual-supp}
\end{figure}

Figure~\ref{fig:acdc-qual-supp} is a single-case illustration of the ACDC rain
failure mode of Figure~\ref{fig:teaser}(A) (mIoU flat or down
while HA rises under fixed-$K$ TENT), here at the aggressive budget rather than the mild
one shown there. Same protocol as the main-text ACDC rain
comparison (lr $2{\times}10^{-3}$, $K{=}8$; full-train $\tau$ calibration, Sec.~\ref{app:calib}).

The archived script for this figure was edited after the published panel was produced; the
edit affects only the four panel-title strings, not the case, the calibration constants, the
minimum-region size or the metric conventions, all of which we checked line by line against
the values quoted above. Re-running it reproduces the panel with slightly different title
formatting. We release the script as it stands rather than reconstructing the earlier strings
from the rendered figure.

\subsection{Cross-Modality Disagreement Scale (Fundus vs.\ Cardiac)}
\label{sec:supp-fundus-delta}
The main-text Limitations cite a concrete scale gap between fundus and cardiac
disagreement ratios. Protocol and numbers:

\begin{itemize}
\item \emph{Statistic.} Disagreement ratio $\delta = |D|/|\Omega|$, where
      $D=\{x:M_0(x)\neq M_k(x)\}$ and $|\Omega|$ is the number of spatial
      locations (same definition as the main-text fragmentation signal).
\item \emph{Fundus.} Episodic TENT on RIM-ONE-DL~\citep{rimonedl} ($n{=}174$ hospital-test OOD),
      with a frozen REFUGE-trained source; mean disagree area
      $\delta{\approx}0.0389$ ($3.9\%$) at the adapted state used for the
      cross-modality scope probes.
\item \emph{Cardiac (comparator).} Episodic EATA / default TTA on M\&Ms
      Vendor-B; cohort-mean step-1 $\delta$ is on the order of $10^{-3}$
      (${\approx}0.001$; the measured fixed-4 step-1 mean is $5.45{\times}10^{-4}$).
\item \emph{Ratio.} Three matched comparisons give values between $37\times$ and
      $71\times$; we quote the conservative figure of roughly $40\times$. Absolute $\delta$
      cutoffs therefore do not transfer across this modality shift without
      recalibration; relative features ($\Delta n_k$, within-domain z-scores /
      tertiles) remain the intended deployment protocol.
\end{itemize}

\paragraph{What the ported threshold does.} A threshold carried over without recalibration ends below its own fixed-budget baseline. The run behind this is episodic TENT on RIM-ONE ($n{=}174$) with the medical gate applied at its
in-domain setting. Un-gated TENT reaches mean Dice $0.646$ against the frozen source's
$0.635$. Porting the gate rolls back $97.8\%$ of cases (whole-image) or $85.1\%$
(region-level), which lands at $0.635$ and $0.636$, at the un-adapted source and below the un-gated run it was meant to protect. The failure mode is exactly what the scale
gap predicts: a cut-point calibrated where $\delta{\sim}10^{-3}$ rejects almost everything
where $\delta{\sim}4{\times}10^{-2}$.

This is a \emph{scope boundary}, not a fundus leaderboard claim: we do not
report FG-TTA controllers as primary results on fundus.

\subsection{Qualitative Case Selection (Medical Figure)}
\label{sec:supp-qual-select}

The main-text medical qualitative figure (``Harm invisible to Dice'')
illustrates fixed-budget over-adaptation that is nearly invisible to mean
overlap. The main text shows the M\&Ms row only; the prostate row is
Fig.~\ref{fig:med-qual-prostate} below. Ideal selection would jointly satisfy four desiderata on the
\emph{displayed} slice: (i)~high HA under fixed-4, (ii)~flat
$|\Delta\mathrm{Dice}|$, (iii)~both helpful and harmful disagreement
components visible on that slice, and (iv)~$M_0{\approx}$GT visually.
On M\&Ms Vendor-B fixed-4 dumps ($n{=}230$), \emph{no} slice meets all four
simultaneously. We therefore prioritize (i)--(ii), since those are the
properties the figure claims to illustrate, and take (iii) when it does not
conflict with flat $\Delta$Dice.

\paragraph{Picks.}
\begin{itemize}
\item \emph{M\&Ms (main text):} \texttt{A8E1F4\_frame09}. Panel metrics are
      \emph{slice-level}. Fixed-4 raises slice Dice $0.55{\to}0.56$ while
      HA${=}0.90$; the display slice contains both helpful and harmful edits.
      On this slice GT has no RV voxels, so mean FG-Dice is blind to the large
      false-RV expansion (FP on an empty class does not move Dice); HA still
      flags it. Weak $M_0$ is intentional domain-shift motivation. Router
      returns HA${=}0$ at $M_0$.
\item \emph{Prostate (Fig.~\ref{fig:med-qual-prostate}):} \texttt{prostate\_B\_0017}. High-$M_0$
      complement (slice $D{\approx}0.93$); fixed-4 HA${=}0.62$ at near-flat
      Dice; FragSubset HA${=}0$. The volume mixes helpful and harmful components in
      3D (7\,H / 8\,B); the peak-harm display slice is harm-dominated.
\end{itemize}

A rejected alternative (\texttt{B5T6V0\_frame00}) has high $M_0$ Dice on the
peak-harm slice but drops Dice by ${\sim}0.13$ under fixed-4, which would
\emph{contradict} an ``invisible to Dice'' caption; we do not use it.

The cross-backbone figure of Sec.~\ref{app:swin} (Fig.~\ref{fig:swin-geom}) uses a different
rule: the same case under both backbones, both $M_0$ within the matched band, and the median
transformer-side disagreement volume among those. Its slice is the one carrying the most
nnU-Net supra-threshold disagreement area. That slice rule was fixed after an earlier one
(most transformer-side disagreement) tied on the chosen case and, at either tied slice,
showed no nnU-Net component.

\begin{figure}[t]
\centering
\includegraphics[width=0.85\columnwidth]{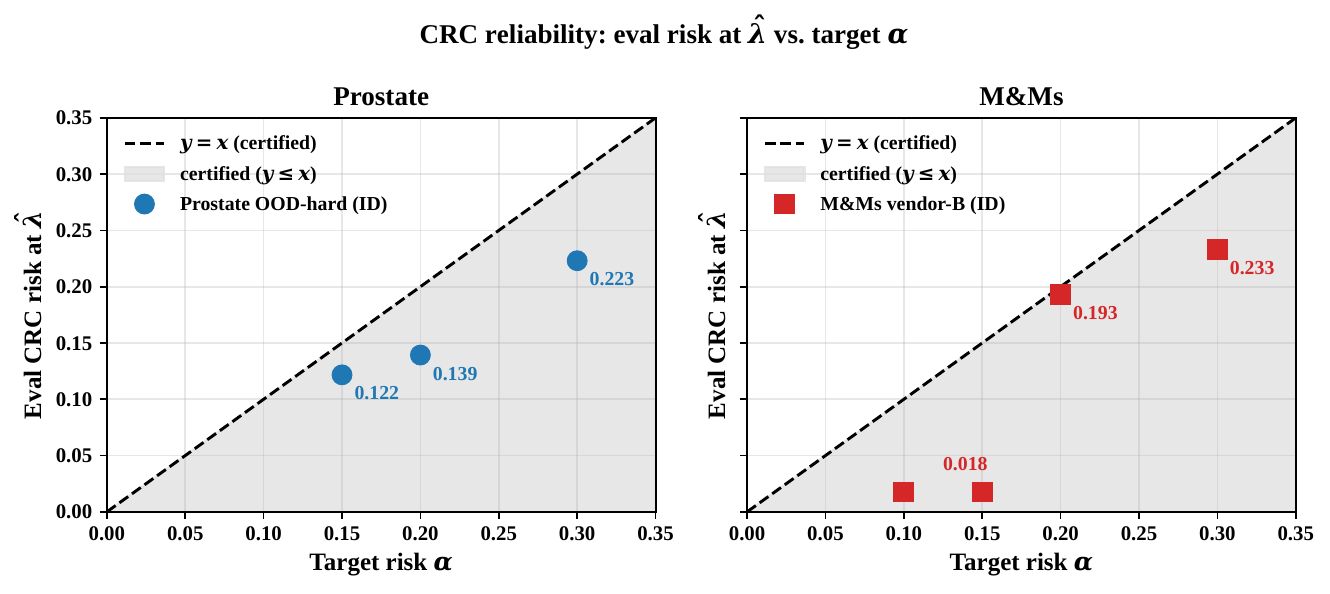}
\caption{CRC reliability diagram. Target $\alpha$ vs.\ \emph{eval} risk at $\hat\lambda$; the dashed diagonal is $y{=}\alpha$ and the shaded half below it is where the certificate holds ($y\leq\alpha$). Both benchmarks sit in the safe region at every $\alpha$ they are feasible at; prostate has no point at $\alpha{=}0.10$, where $n_{\mathrm{cal}}{=}8$ puts the conformal floor above the budget (Sec.~\ref{sec:supp-crc-op}). Computed on the verification rerun described there.}
\label{fig:crc}
\end{figure}

\section{Results Deferred from the Main Text}
\label{sec:supp-deferred}

This section holds material compressed out of the main paper for space. Every
item is referenced from the main text at the point where it was cut.

\subsection{Prostate OOD-hard Panel}
\label{sec:supp-oodhard}

The main text reports prostate MRI on the full OOD-all set ($n{=}93$) and summarizes this subset in one paragraph. Its cut-points are fitted once on the OOD holdout and applied to the $93$-case stream; filtering that stream down to its $34$ hardest cases leaves $0/6/28$ cases in the three buckets, which is too degenerate for a bucket-level number to mean anything. Any router row a reader reconstructs for this subset is descriptive (Tier~C) and should not be compared against the intervals reported for the full split. Table~\ref{tab:supp-oodhard} gives the
panel. The ordering reverses here: MEMO attains the best HA overall and the best
Dice among the non-fixed methods, at the full four-step budget with extra backprop, against FragSubset's $2.29$ steps
without gradients. The COQR router is not reported on this subset (Sec.~\ref{sec:supp-calib}).

\begin{table}[h]
\centering
\begin{tabular}{lccc}
\toprule
Method & Dice & HA & Steps \\
\midrule
$M_0$ (no adapt, ref.) & 0.556 & 0 & 0 \\
\midrule
fixed-1 & 0.589 & 0.290 & 1.00 \\
fixed-2 & 0.586 & 0.309 & 2.00 \\
fixed-3 \emph{(retrospective)} & 0.581 & 0.257 & 3.00 \\
fixed-4 \emph{(default)} & 0.574 & 0.318 & 4.00 \\
\midrule
FragStop (ours) & 0.582 & 0.285 & 2.56 \\
FragSubset (ours) & 0.584 & 0.276 & \textbf{2.29} \\
\midrule
MEMO & \textbf{0.588} & \textbf{0.209} & 4.00 \\
CoTTA & 0.546 & 0.240 & 4.00 \\
\bottomrule
\end{tabular}
\caption{Prostate MRI OOD-hard ($n{=}34$), the hardest subset of OOD-all; online.
The fixed-$k$ rows are the trajectory checkpoints of Table~\ref{tab:budget-sweep};
FragStop and FragSubset come from separate executions of the same protocol on the same cases, which reproduces those checkpoints at step~1 to $1{\times}10^{-4}$ in Dice and $9{\times}10^{-6}$ in $\delta$, with identical $M_0$ predictions; region counts differ on two cases in the FragStop row (Sec.~\ref{app:rerun}). \textbf{Bold} marks the best value per column among
adaptive methods (the fixed-$k$ ladder is reported for context).}
\label{tab:supp-oodhard}
\end{table}

Paired bootstrap on this subset is in Sec.~\ref{sec:supp-pairwise}.

\subsection{ACDC Rain: Stop Rule Across Learning Rates}
\label{sec:supp-lrsweep}

The z-scored fragmentation stop cuts HA against the fixed-budget end along the entire
budget axis; Table~\ref{tab:supp-acdc-lr} gives the sweep. The effect is monotone in the
learning rate: a larger budget produces more over-adaptation, and the stop rule
removes more of it. Each row is independently calibrated, with $(\mu_n,\sigma_n)$ and $\tau$ refit on the train split at that learning rate, so the trend is not one threshold
reused across budgets. This is why we read the HA cut as tuning-robust rather than as an artifact of one learning-rate choice. The accuracy side moves little: $\tau$-stop trails fixed-$8$ mIoU by $0.002$ to $0.010$ at every rate while cutting HA by $0.067$ to $0.138$.

\begin{table}[h]
\centering
\begin{tabular}{lcccc}
\toprule
lr & $\tau$-stop (mIoU\,/\,HA) & fixed-$8$ (mIoU\,/\,HA) & HA cut & $\tau$-stop steps \\
\midrule
$1{\times}10^{-4}$ & $0.368$\,/\,$0.152$ & $0.370$\,/\,$0.219$ & $-0.067$ & $5.17$ \\
$5{\times}10^{-4}$ & $0.375$\,/\,$0.255$ & $0.385$\,/\,$0.336$ & $-0.081$ & $3.80$ \\
$1{\times}10^{-3}$ & $0.378$\,/\,$0.272$ & $0.388$\,/\,$0.393$ & $-0.122$ & $2.71$ \\
$2{\times}10^{-3}$ & $0.382$\,/\,$0.310$ & $0.387$\,/\,$0.448$ & $-0.138$ & $2.49$ \\
\bottomrule
\end{tabular}
\caption{ACDC rain: $\tau$-stop vs.\ fixed-K across learning rates at a fixed $K{=}8$ (val, $n{=}100$); accuracy and deployed depth are reported beside HA, since HA alone is not interpretable (Sec.~\ref{sec:prelim}). Fixed-$8$ deploys $8$ steps by definition. The $2{\times}10^{-3}$ row is the main-text operating point.}
\label{tab:supp-acdc-lr}
\end{table}

The case bootstrap at the main operating point, covering both accuracy and reliability, is in Sec.~\ref{sec:supp-pairwise}.

\subsection{Ablation Configurations and Their Names}
\label{app:ablations}
\label{sec:supp-ablation-names}
The budget-axis ablations are instances of one template: a modulation rule applied per step,
optionally combined with the fragmentation stop. The names used in the main text are:

\begin{itemize}
\item \texttt{FragStop}: the stop alone, with no modulation. It halts at the first rise
      $\Delta n_k > 0$.
\item \texttt{combo+param+Stop} (reported as \textbf{FragSubset}). The configuration key records
      the intended design: entropy weight down and Fisher weight up with $r_k$ (\texttt{combo}), restriction of the updated parameter subset (\texttt{param}), and the stop. The configuration actually run, and reported in both main tables, is
      \texttt{mode=param\_subset}: the entropy weight is held at $1.0$ and the Fisher weight
      at $\lambda_{F,0}{=}1.0$, and the only risk-dependent quantity is the parameter subset
      (decoder normalization when $r_k > t_{\mathrm{mid}}{=}0$, all normalization otherwise).
      The stored configuration also carries $\beta{=}0.7$ and $\lambda_{F,\max}{=}4.0$;
      \textbf{neither is read in this mode}, and we list them only so the stored settings can
      be matched to this description. Both medical benchmarks use these settings
      (\texttt{fragopt\_oodall}: Dice $0.7549$, HA $0.1659$, $2.677$ steps, i.e.\ the FragSubset row
      of Table~\ref{tab:prostate}).
\item \texttt{ent-weight+Stop}: partial modulation, with the entropy weight only, without the
      Fisher term or the parameter subset, plus the stop. The two continuous modulations
      referred to by the keys above scale the entropy weight down and the Fisher weight up
      with the risk, $w_{\text{ent}}(r_k){=}\mathrm{clip}(e^{-\beta r_k}, w_{\min}, 1)$ and
      $\lambda_{\text{Fisher}}(r_k){=}\mathrm{clip}(\lambda_{F,0}e^{+\beta r_k},
      \lambda_{F,0}, \lambda_{F,\max})$; these are the forms $\beta$ and
      $\lambda_{F,\max}$ parametrize.
\item \texttt{loss-switch+Stop}: the negative control, a discrete switch between the
      entropy and Fisher objectives instead of continuous modulation, plus the stop.
\end{itemize}

On prostate OOD-hard ($n{=}34$), against the fixed-4 row of
Table~\ref{tab:supp-oodhard} (HA $0.318$): \texttt{combo+param+Stop} $-0.043$,
\texttt{FragStop} $-0.033$, \texttt{ent-weight+Stop} $-0.005$, \texttt{loss-switch+Stop}
$+0.024$ (Dice $0.588$). All four are differences against that single baseline, so each
closes against the table. That baseline comes from a different execution (caption
there); at $0.043$ the largest is twenty times the $\pm 0.002$ cross-execution scale of
Sec.~\ref{app:rerun}. The ordering is what the main text reads as the
parameter-subset member carrying the reduction while modulating the entropy weight alone
does not.

One further qualification is benchmark-specific. The parameter subset is the only
risk-dependent quantity in this mode, and whether it varies at all depends on where
$r_k$ falls relative to $t_{\mathrm{mid}}{=}0$. On prostate it varies: $r_k>0$ on
$23.7\%$ of steps in the OOD-all run and on $11$ of $78$ (case, step) rows ($14.1\%$)
in the OOD-hard run behind Table~\ref{tab:supp-oodhard}. On M\&Ms it does not: $r_k$
ranges from $-2.000$ to $-0.392$ (median $-1.869$) and the condition fires on none of
$788$ rows, so both branches produce an identical adaptation configuration. The M\&Ms
FragSubset row of Table~\ref{tab:mnm} is therefore not FragStop plus modulation; the
two runs share their adaptation configuration and differ only in how the stop is evaluated: FragStop tests a single-step rise $n_k > n_{k-1}$, while this path scans
the trajectory for the last step before the first rise. The residual HA difference
($0.109$ vs.\ $0.115$, i.e.\ $-0.0054$) therefore has nothing to do with the modulation,
and it comes from four cases out of $230$: the two rules select different depths on two
of them ($-0.0037$ of the total), and two further boundary regions flip between the runs,
which are separate executions agreeing at step~1 to $1.6{\times}10^{-4}$ in Dice
(Sec.~\ref{app:rerun}). The remaining $226$ cases contribute $+0.0001$.

\subsection{A Fifth Episodic Objective: SAR}
\label{sec:supp-sar}

To test compilability beyond the TENT / EATA / CoTTA / MEMO trajectories, we wrap
the controllers around SAR on prostate OOD-all under the identical
protocol. We disable SAR's EMA recovery: under our binary entropy scale its reset
condition fires at every step, which collapses $M_1$ back to $M_0$ and makes the trajectory trivial. We did not instead rescale that threshold by $\ln C$, as Sec.~\ref{app:newobj} does for DeYO's; the asymmetry is a limitation of this probe rather than a considered choice, and a rescaled reset may well keep the recovery mechanism alive.

SAR is a regime where stopping cannot create value, and we report it because that
boundary is informative rather than because it is favorable:
\begin{itemize}
\item SAR falls below the un-adapted source from the first step
      (Dice $0.734$ at $k{=}1$ against $M_0$ $0.738$), so there is no early
      segment worth keeping.
\item It accumulates the highest HA of any objective we evaluate on prostate
      OOD-all ($0.374$ at $k{=}4$; the driving benchmark reaches higher still,
      Table~\ref{tab:supp-acdc-fogsnow}).
\item FragStop tracks the ladder rather than beating it: $0.724$\,/\,$0.339$ at
      $2.06$ steps, worse than SAR fixed-1 ($0.734$\,/\,$0.300$) on both axes;
      $\Delta$HA vs.\ fixed-4 is $-0.035$ $[-0.078,+0.009]$, crossing zero.
\item The oracle stop on the same trajectories reaches $0.753$\,/\,$0.266$, so
      per-case structure does exist here that fragmentation does not recover.
\end{itemize}
This bounds the stop rule's applicability: it requires a trajectory with an early
segment worth keeping. The router is unaffected by this bound, since rollback
does not depend on an improving prefix.

\subsection{Certified Operating Point: A Monotone Surrogate}
\label{sec:supp-crc-op}

\paragraph{The surrogate that is certified.} Conformal risk control
(CRC)~\citep{vovk2005algorithmic,crc2024} requires a risk \emph{monotone} in the gate
threshold, and deployment HA is not: the dynamic-denominator property that makes it
non-monotone in the budget (Sec.~\ref{sec:budget}) acts here in the gate variable, since
$|D(\lambda)|$ shrinks as the gate rejects edits, so rejecting more can leave HA flat or
raise it. Fixing the denominator to $|D_{\max}|$ and scoring each region by its mean entropy drop gives a strictly monotone risk to which CRC applies unchanged. The obstruction is specific to monotone CRC: distribution-free procedures for non-monotone bounded losses, such as Learn-Then-Test, could in principle certify deployment HA itself on a discretised threshold grid, and we did not run one. The
certificate therefore covers the fixed-denominator surrogate, \emph{not} the deployed
dynamic HA.

The main text states the correctness result (once the fixed-denominator surrogate replaces deployment HA, the region gate certifies at the requested budget) and defers the operating points here ($\hat\lambda$ is calibrated as in Sec.~\ref{app:calib}).

\paragraph{What this section is computed on.} The certificate needs per-voxel
probabilities, which the archived trajectories do not store, so the operating points below come from a second verification rerun of the same EATA trajectories as the main tables, separate from the one in Sec.~\ref{app:rerun} and checked against the archived runs by the same criteria: on cardiac, step-$1$ gradient norms match to $2.7{\times}10^{-7}$ and cohort Dice and HA to $2.4{\times}10^{-6}$ and $1.5{\times}10^{-4}$. Two conventions differ from the rest of the paper and are worth
stating. First, only cases whose step-$1$ edit contains a component above the
minimum-size threshold can be gated at all, since the gate has nothing to accept or
reject otherwise: $168$ of the $230$ cardiac evaluation cases and $10$ of the $20$
calibration cases qualify, against $34$ of $34$ and $8$ of $8$ on prostate. Second,
the certified regions are connected components of the raw multi-class disagreement,
whereas HA on the medical benchmarks scores components of the foreground-binarized
disagreement, so a class-to-class edit is a region here and is not one there; on
cardiac the two counts are $168$ and $167$ on the same masks. The $167$ here is computed on the verification rerun (Sec.~\ref{app:rerun}); the archived count at $k{=}1$ in Table~\ref{tab:app-accounting} coincides, though the two runs differ by one case at $k{=}2$ and $k{=}3$ (Sec.~\ref{app:tau}).

\begin{itemize}
\item \emph{Prostate OOD-hard.} Certifies at all three feasible budgets, with eval
      risk $0.122$, $0.139$ and $0.223$ against calibration bounds of $0.142$, $0.184$
      and $0.270$; $\alpha{=}0.10$ is infeasible at $n_{\mathrm{cal}}{=}8$ (below).
\item \emph{M\&Ms vendor-B.} Certifies at all four, with eval risk $0.018$ at
      $\alpha{=}0.10$ and $0.15$, $0.193$ at $0.20$ and $0.233$ at $0.30$.
\item \emph{Reliability diagram.} Fig.~\ref{fig:crc} plots target $\alpha$
      against eval risk at $\hat\lambda$. Both benchmarks certify at every $\alpha$
      they are feasible at, and feasibility is what separates them: the two sit on
      opposite sides of the same $1/(n_{\mathrm{cal}}{+}1)$ floor (below).
\end{itemize}

Two boundaries are inherent to CRC at small holdouts, and neither is specific to
fragmentation:
\begin{itemize}
\item \emph{Feasibility is bounded by $n_{\mathrm{cal}}$.} The prostate certificate is
      calibrated on the $8$ OOD-hard cases held out of the $34$-case evaluation split (the
      remainder of the $42$-case pool, Sec.~\ref{sec:supp-eval}) and evaluated on the $34$.
      At $\alpha{=}0.10$ with $n_{\mathrm{cal}}{=}8$ the conformal bound is at best
      $1/(n_{\mathrm{cal}}{+}1){=}1/9{=}0.111 > \alpha$ even if every region is rejected,
      so no gate can certify. The M\&Ms certificate is calibrated on the $10$ scorable
      cases of the $20$-case vendor-B calibration split, where the same floor is
      $1/11{=}0.091$ and $\alpha{=}0.10$ clears it by $0.009$. The two benchmarks
      therefore fall on opposite sides of one arithmetic bound, which is what makes this
      a property of the calibration set size rather than of either dataset.
      It relaxes on the larger ACDC splits.
\item \emph{The certified surrogate is deliberately not deployment HA.} At the
      same $\hat\lambda$, gated deployment HA runs two to four times the certified
      surrogate risk (prostate $\alpha{=}0.20$: $0.367$ against $0.139$, $2.6\times$;
      M\&Ms $\alpha{=}0.10$: $0.070$ against $0.018$, $4.0\times$).
      Across the seven feasible operating points (prostate $\alpha{=}0.10$ is infeasible at
      $n_{\mathrm{cal}}{=}8$) the ratio ranges from $1.7\times$ to $4.0\times$, the widest
      gap being M\&Ms at $\alpha{=}0.10$ and $0.15$. We report the
      deployment number openly so the certificate is read for what it
      guarantees, a monotone region-risk bound on the fixed-denominator surrogate, rather than mistaken for a direct bound on deployment HA. The
      gap is the intended output of the dual-metric design.
\end{itemize}

\subsection{Prostate Qualitative Row}
\label{sec:supp-med-qual-prostate}

\begin{figure}[h]
\centering
\includegraphics[width=\columnwidth]{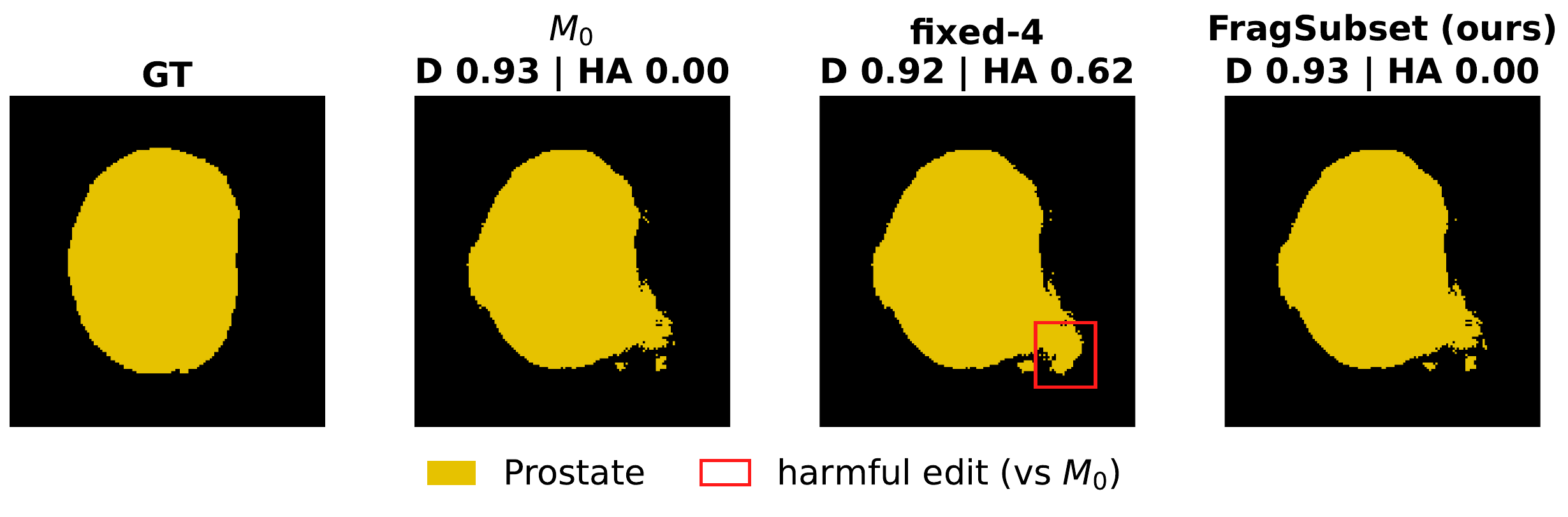}
\caption{Harm invisible to Dice, prostate row (FragSubset, \texttt{prostate\_B\_0017}); the
companion of the main-text M\&Ms row. L$\rightarrow$R: \textbf{GT} (evaluation reference only); $M_0$; fixed-4, where the red box locates the harmful
accepted edit it introduces relative to $M_0$ (the flagged component is $130$ pixels on this slice, under
$1\%$ of the panel, so we mark its position rather than trace a boundary too small to read);
and FragSubset. Panel D/HA are
\emph{slice-level}, not case means: slice $M_0$ D${=}0.93$; fixed-4 D${=}0.92$ at
HA${=}0.62$; FragSubset keeps HA${=}0$. Selection protocol in
Sec.~\ref{sec:supp-qual-select}.}
\label{fig:med-qual-prostate}
\end{figure}

The main text carries the M\&Ms row of the ``harm invisible to Dice'' figure;
Fig.~\ref{fig:med-qual-prostate} is its prostate companion, selected by the protocol of
Sec.~\ref{sec:supp-qual-select}.

\subsection{ACDC: All Four Conditions}
\label{sec:supp-acdc-fogsnow}

The main text discusses night (the hard condition, where $M_0$ is near-collapsed) and
gives the four-condition HA cascade inline; Table~\ref{tab:supp-acdc-fogsnow} is the full
panel, and rain is the condition where adaptation helps most. Fog and snow behave
like rain: the cascade TENT$_8\rightarrow\tau$-stop$\rightarrow$Router lowers HA at a small
mIoU cost, and CoTTA holds HA down only by sitting at $M_0$, which it does on all four conditions. On rain the HA reduction grows with the budget ($\Delta$HA $-0.138$ on
rain at lr $2{\times}10^{-3}$, against $-0.067$ at lr $10^{-4}$).

\begin{table}[h]
\centering
\small
\begin{tabular}{lccccc}
\toprule
Split & $M_0$ & TENT$_8$ & CoTTA & $\tau$-stop & Router \\
\midrule
fog   & $0.474$ & $0.486$\,/\,$0.445$ & $0.473$\,/\,$0.171$ & $0.483$\,/\,$0.211$ & $0.481$\,/\,$0.147$ \\
rain  & $0.364$ & $0.387$\,/\,$0.448$ & $0.365$\,/\,$0.159$ & $0.382$\,/\,$0.310$ & $0.374$\,/\,$0.216$ \\
snow  & $0.349$ & $0.367$\,/\,$0.455$ & $0.348$\,/\,$0.219$ & $0.362$\,/\,$0.336$ & $0.359$\,/\,$0.219$ \\
night & $0.194$ & $0.182$\,/\,$0.501$ & $0.193$\,/\,$0.282$ & $0.193$\,/\,$0.375$ & $0.193$\,/\,$0.232$ \\
\bottomrule
\end{tabular}
\caption{ACDC, all four conditions (mIoU\,/\,HA; lr $2{\times}10^{-3}$, $K{=}8$;
$n{=}100$ for fog/rain/snow, $n{=}106$ for night). In the roles easy / main / cross / hard,
fog and snow are the easy and cross conditions; rain and night are the two the main text
discusses. TENT$_8$ is TENT run to the fixed budget with no control; $M_0$ takes no step, so
its HA is $0$ by construction. Router steps per case: $0.96$ (fog), $0.97$ (rain),
$1.01$ (snow), $1.67$ (night).}
\label{tab:supp-acdc-fogsnow}
\end{table}

On fog the source model is strongest and the signal correspondingly quiet. Under the mild
budget (lr $10^{-4}$, $K{=}4$) the stop rule runs the full budget on $49\%$ of cases, and
$\tau$-stop mIoU ($0.4751$) sits within $0.002$ of both fixed-K ($0.4755$) and the
validation oracle ($0.4766$): fragmentation does not over-fire when there is nothing to
stop. The hard tertile is again where $M_0$ is weakest (fog $0.456$ against $0.477$ in the
low tertile).

\subsection{Remaining Pairwise Intervals}
\label{sec:supp-pairwise}

This section is the single authoritative list of \emph{deployment-level controller-vs-budget} paired intervals (excluding the SAR probe of Sec.~\ref{sec:supp-sar}, the frozen-design test of Sec.~\ref{app:algo}, and the backbone and objective extensions of Secs.~\ref{app:swin} and~\ref{app:newobj}, each reported in place): every such comparison quoted anywhere in the paper appears here once, so a number has exactly one source. Matched-control intervals (Sec.~\ref{app:matched}) and calibration-sensitivity bands (Sec.~\ref{app:calibsens}) are different comparisons and live in their own sections. All are $B{=}5000$, seed $1000$, paired on cases. Point estimates
are computed on unrounded case-level values, so each may differ from the corresponding
three-decimal difference of the main tables by up to $0.001$; a larger gap would indicate
that the per-case source and the table are not the same file.

\emph{M\&Ms vendor-B ($n{=}230$).}
\begin{itemize}
\item Router vs.\ deployable fixed-2: $\Delta$HA $-0.151$ $[-0.191,-0.115]$ (Tier-A),
      at $\Delta$Dice $-0.0015$ $[-0.0032,+0.0001]$.
\item Router vs.\ retrospective fixed-3: $\Delta$HA $-0.116$ $[-0.152,-0.081]$,
      Wilcoxon $p{=}9.1{\times}10^{-10}$, at $\Delta$Dice $-0.0008$
      $[-0.0022,+0.0006]$; about $5.8\times$ the corresponding FragSubset-vs-fixed-3
      margin on this split.
\item Router vs.\ default fixed-4: $\Delta$HA $-0.163$ $[-0.202,-0.126]$ (Tier-A$'$),
      at $\Delta$Dice $-0.0000$ $[-0.0014,+0.0012]$.
\item Router vs.\ CoTTA: $\Delta$HA $-0.084$ $[-0.113,-0.055]$ at $\Delta$Dice $+0.0003$
      $[-0.0004,+0.0009]$.
\item Router vs.\ FragSubset: $\Delta$HA $-0.096$ $[-0.130,-0.062]$, at a third of the deployed
      updates, or about half the executed ones.
\item FragSubset vs.\ deployable fixed-2: $\Delta$HA $-0.055$ $[-0.084,-0.028]$ (Tier-A).
\item FragSubset vs.\ retrospective fixed-3: $\Delta$HA $-0.020$ $[-0.040,-0.000]$
      (the upper bound is below zero but rounds to it, which is why $P{=}0.978$ rather
      than $1.0$; barely separated, Tier-B).
\end{itemize}

\emph{Prostate OOD-all ($n{=}93$).}
\begin{itemize}
\item Router vs.\ default fixed-4: $\Delta$HA $-0.162$ $[-0.227,-0.098]$ (Tier-A$'$),
      $61/93$ wins, at $\Delta$Dice $-0.0077$ $[-0.0236,+0.0072]$.
\item Router vs.\ deployable fixed-1: $\Delta$HA $-0.142$ $[-0.194,-0.094]$,
      $P(\Delta\mathrm{HA}{<}0){=}1.0$, at $\Delta$Dice $-0.0130$ $[-0.0252,-0.0022]$
      (the accuracy interval excludes zero, unlike every Dice comparison on M\&Ms).
\item Router vs.\ retrospective fixed-3: $\Delta$HA $-0.072$ $[-0.139,-0.007]$,
      $P(\Delta\mathrm{HA}{<}0){=}0.99$, $42/93$ wins, at $\Delta$Dice $-0.0122$
      $[-0.0279,+0.0020]$.
\item FragSubset vs.\ default fixed-4: $\Delta$HA $-0.096$ $[-0.143,-0.049]$ (Tier-A$'$).
\item FragSubset vs.\ deployable fixed-1: $\Delta$HA $-0.076$ $[-0.113,-0.041]$ (Tier-A;
      $P{=}1.0$, Wilcoxon $p{=}5.7{\times}10^{-5}$, 48/93 wins).
\item FragSubset vs.\ retrospective fixed-3: $\Delta$HA $-0.006$ $[-0.046,+0.035]$
      (CI crosses zero; not claimed), at $\Delta$Dice $+0.004$ $[-0.003,+0.012]$.
\item FragStop vs.\ retrospective fixed-3: $\Delta$HA $+0.011$ $[-0.023,+0.047]$
      (directionally worse; not claimed).
\end{itemize}

\emph{Prostate OOD-hard ($n{=}34$).} Panel in Sec.~\ref{sec:supp-oodhard}.
\begin{itemize}
\item FragSubset vs.\ default fixed-4: $\Delta$HA $-0.043$ $[-0.086,-0.001]$
      ($P(\Delta\mathrm{HA}{<}0){=}0.98$; marginal at the upper bound).
\item FragSubset vs.\ retrospective fixed-3: $\Delta$HA $+0.019$ $[-0.025,+0.068]$
      (point estimate reverses; not claimed).
\item FragSubset vs.\ FragStop (modulation beyond stopping): Wilcoxon $p{=}0.016$ but
      Holm-adjusted $p{=}0.065$ with a bootstrap CI crossing zero. We label this
      \emph{not established}; it is the only ordering claim we make between two of our own
      controllers, and the only place Holm correction is applied
      (Sec.~\ref{sec:supp-eval}).
\end{itemize}

\emph{ACDC rain ($n{=}100$, lr $2{\times}10^{-3}$, $K{=}8$).}
\begin{itemize}
\item Router vs.\ the HA-best fixed-1: $\Delta$HA $-0.048$ $[-0.088,-0.012]$ (Tier-A);
      vs.\ fixed-8: $\Delta$HA $-0.233$ $[-0.281,-0.187]$.
\item $\tau$-stop vs.\ fixed-1: $\Delta$HA $+0.047$ $[+0.025,+0.072]$ (trades HA for
      steps/mIoU against the HA-optimal fixed budget); vs.\ fixed-8: $\Delta$HA $-0.138$
      $[-0.172,-0.105]$ ($P{=}1.0$; the main operating point of the learning-rate sweep,
      Sec.~\ref{sec:supp-lrsweep}).
\item $\tau$-stop vs.\ fixed-8, accuracy side: $\Delta$mIoU $-0.0053$
      $[-0.0124,+0.0017]$ (CI $\ni 0$), so the HA cut is mIoU-neutral.
\item ACDC snow, same setting: $\tau$-stop vs.\ fixed-8 $\Delta$HA $-0.119$
      $[-0.155,-0.085]$ ($P{=}1.0$) at $\Delta$mIoU $-0.0053$ $[-0.0127,+0.0018]$.
\item Router vs.\ $\tau$-stop on rain, accuracy side: $\Delta$mIoU $-0.008$
      $[-0.014,-0.003]$, the one condition where routing on top of stopping costs
      significant accuracy.
\item ACDC night ($n{=}106$): Router vs.\ TENT$_8$ $\Delta$mIoU $+0.011$
      $[+0.004,+0.018]$ at $\Delta$HA $-0.269$; both are damage control against a
      TENT run that falls below $M_0$, not an adaptation gain.
\end{itemize}

\emph{ACDC replay oracle.} A replay oracle that selects, per case, the best-mIoU point along
the stored aggressive-budget trajectory reaches mIoU $0.410$ at HA $0.278$ on rain,
$0.391$ at $0.255$ on snow and $0.215$ at $0.183$ on night, at $3.5$, $3.5$ and $2.4$
deployed steps respectively. It is an upper bound on \emph{accuracy}, not on reliability:
on rain the router sits below it on HA ($0.216$ against $0.278$) while the
$\tau$-stop does not ($0.310$). We read oracle numbers as
bounds rather than as stable per-case labels, because on plateau-like trajectories the
utility surface over $\{M_k\}$ is nearly flat, so the arg-max step need not be unique; we
never train against them.

\section{Matched-Control Decomposition Across Regimes}
\label{app:matched}

The main-text tables compare controllers against \emph{budgets}. This section asks a
different question, the one a reader who suspects the metric can be gamed will ask: at the
router's own rollback rate, how much of the HA reduction is attributable to the
\emph{fragmentation score} rather than to the act of declining to adapt? Every alternative
selector below deploys exactly the same actions at exactly the same bucket sizes; only the
score that assigns cases to buckets changes. All replays reproduce the deployed controller's
own row to the precision at which the main tables print it, which is the calibration that
licenses the counterfactuals: prostate $0.7387/0.0994/0.548 \to 0.739/0.099/0.55$ with buckets
$27/24/42$, and M\&Ms $0.8489/0.0133/1.104 \to 0.849/0.013/1.10$ with buckets $130/46/54$.

\paragraph{M\&Ms: fragmentation separates from every alternative.}
\begin{table}[h]
\centering\small\setlength{\tabcolsep}{5pt}
\begin{tabular}{lccc}
\toprule
Selector (same actions, buckets $130/46/54$) & Dice & HA & Steps \\
\midrule
\textbf{Fragmentation} ($\delta$, deployed) & $0.8489$ & $\mathbf{0.0133}$ & $1.104$ \\
Augmentation agreement ($8$ forward passes) & $0.8490$ & $0.0451$ & $1.104$ \\
Entropy & $0.8496$ & $0.0537$ & $1.104$ \\
MSP & $0.8494$ & $0.0567$ & $1.104$ \\
Random (mean of $1{,}000$ draws) & $0.8494$ & $0.0605$ & $1.104$ \\
\midrule
Oracle on true HA at step $1$ & $0.8503$ & $0.0023$ & $1.104$ \\
Oracle on true $\Delta$Dice & $0.8515$ & $0.0217$ & $1.104$ \\
$M_0$ (no adaptation) & $0.8489$ & $0$ & $0$ \\
\bottomrule
\end{tabular}
\caption{Matched-bucket controls on M\&Ms vendor-B ($n{=}230$). Every selector row rolls back the same number of cases and deploys the same per-bucket depths; only the ranking score differs.}
\label{tab:app-matched-mnm}
\end{table}

No random assignment out of $1{,}000$ reaches the router's HA. Paired bootstrap over cases
($B{=}5000$) against each informed alternative: $\Delta$HA $-0.040$ $[-0.065,-0.018]$ vs.\
entropy, $-0.043$ $[-0.070,-0.019]$ vs.\ MSP, and $-0.032$ $[-0.057,-0.009]$ vs.\ augmentation
agreement, at $\Delta$Dice of $-0.0007$, $-0.0005$ and $-0.0001$ respectively. The margin over
the strongest alternative is $16\times$ the run-to-run reproducibility of a cohort HA mean
(Sec.~\ref{app:rerun}), and the consistency baseline pays eight extra forward passes per step
for a signal that fragmentation obtains from the two masks it already has.

\paragraph{Prostate: informative, but not distinguishable from confidence.}
At the router's rollback rate ($42/93$), fragmentation-scored routing gives HA $0.0994$ against
$0.1325$ $[0.1053,0.1607]$ for random assignment, $0.1044$ for entropy and $0.1030$ for MSP.
Fragmentation beats random ($P{=}0.010$ over $1{,}000$ draws), so the score carries real
information here too, but paired bootstrap against the confidence proxies gives $\Delta$HA
$-0.0050$ $[-0.0572,+0.0475]$ and $-0.0036$ $[-0.0569,+0.0497]$, intervals that straddle zero, while $\Delta$Dice is $-0.0110$ $[-0.0206,-0.0026]$, which does not. The two hard
buckets overlap on only $24$ of $42$ cases, so this is not one partition perturbed. The reading
consistent with Sec.~\ref{sec:prostate} is that the HA cut on this split comes mostly from
rolling back difficult cases, and several risk-correlated signals rank difficulty.

\paragraph{ACDC: a controlled substitute, and a null result.}
The archived per-case records for ACDC store only the action each case actually received, so a
rolled-back case has no recorded ``had it run'' outcome and the full counterfactual is not
available offline. We therefore report a narrower experiment: rollback maps to $M_0$ and every
retained case to the $\tau$-stop outcome, which \emph{is} recorded for all cases. \textbf{This
does not reproduce the deployed ACDC router} (whose retain action is the parameter-free
$\Delta n_k$ stop rule within a per-bucket cap, not the global $\tau$-stop; the
per-condition tertiles assign buckets, they are not the stop rule), and its numbers are not comparable with the
main-text ACDC table; it isolates one question only: given a fixed rollback quota and a fixed
retain action, does $\delta$ pick the right cases? Selecting the top-$n_{\mathrm{hard}}$ cases
by $\delta$ reproduces the deployed hard bucket exactly in all four conditions
($18/21/29/43$), so the selector is faithful.

\begin{table}[h]
\centering\small\setlength{\tabcolsep}{5pt}
\begin{tabular}{lcccccc}
\toprule
Condition & $n$ & $n_{\mathrm{hard}}$ & $\delta$ HA & Random HA [95\%] & Oracle HA & $P(\text{rand}\le\delta)$ \\
\midrule
fog   & $100$ & $18$ & $\mathbf{0.150}$ & $0.173$ $[0.156,0.188]$ & $0.106$ & $\mathbf{0.005}$ \\
rain  & $100$ & $21$ & $0.242$ & $0.245$ $[0.221,0.266]$ & $0.147$ & $0.366$ \\
snow  & $100$ & $29$ & $0.256$ & $0.238$ $[0.214,0.262]$ & $0.136$ & $0.934$ \\
night & $106$ & $43$ & $0.206$ & $0.222$ $[0.198,0.246]$ & $0.113$ & $0.087$ \\
\midrule
pooled & --- & --- & $0.213$ & $0.220$ & $0.126$ & --- \\
\bottomrule
\end{tabular}
\caption{Rollback-selection control on ACDC at the aggressive budget (retain action fixed to
$\tau$-stop). $\delta$-based selection is not better than random overall, fog excepted.}
\label{tab:app-matched-acdc}
\end{table}

The null is not a surprise but a prediction: at this budget $\delta$'s association with HA is weak ($0.139$) and $n_{\mathrm{reg}}$'s has inverted (Sec.~\ref{app:ci}), and a score that barely correlates with harm cannot rank cases by it. The routing increment the main text reports at this budget therefore comes chiefly
from the rollback quota rather than from case selection. A full counterfactual would require
re-running the ACDC trajectories under each alternative assignment; we list this as a
contingency rather than a claim.

\paragraph{Prostate158: the quota, not the ranking.} The frozen-design test of
Sec.~\ref{app:algo} is the one deployment in this paper whose benchmark took no part in the
design's selection, so it is also the cleanest place to ask what the score contributes there. We
replay it under the protocol used above: bucket sizes $48/39/39$ and depths $3/2/0$ held
fixed, only the ranking score changed. The replay reproduces the deployed arm: cut-points refit from the $32$-case calibration split agree with the archived values to the last digit,
assignment by cut-point and by rank agree on all $126$ evaluation cases, and the four
fixed-$k$ rows reproduce Table~\ref{tab:app-frozen-p158} to printed precision.

The deployed coordinate is not better than random here. Fragmentation ($\delta$ at step~$1$)
reaches HA $0.161$ against $0.164$ $[0.144,0.184]$ for random assignment at the same quota,
and $39.5\%$ of $1{,}000$ random draws match or beat it; entropy ($0.161$),
$n_{\mathrm{reg}}$ ($0.157$) and gradient norm ($0.170$) all fall inside that band, and none
separates on Dice. What this isolates is the ranking alone: no shrink outcome is recorded for
cases the deployed design did not send to the mid bucket, so every arm is compared at depths
$3/2/0$, and the shrink action contributes a further $-0.022$ HA independently of how cases
are ordered. The frozen design's advantage over that benchmark's own retrospective-best
budget is therefore unaffected; what does not survive is the attribution of that advantage to
case selection.

As on ACDC, the null is a prediction rather than a surprise, and here it is a quantitative
one. Measured on the same trajectories, with the pipeline that reproduces the published pooled prostate entries ($n{=}124$) of Table~\ref{tab:signal-ha} to the last digit, $\delta$'s level correlation with HA is $0.156$ here, against $0.557$ on that pooled prostate set and $0.516$ on
cardiac, while its step-differenced correlation is $0.449$, in line with those benchmarks
($0.515$ and $0.347$). The router ranks cases by the level, and the level is what has
collapsed; $n_{\mathrm{reg}}$ is weaker still ($0.050$), so its narrow edge above is noise
and not a coordinate preference. The case-level ranker of Table~\ref{tab:app-auroc}
agrees: scored the same way, the step-$1$ signal ranks top-tertile HA at $0.475$
($\delta$) and $0.506$ ($n_{\mathrm{reg}}$) here, against $0.686$ and $0.685$ ($\delta$, $n_{\mathrm{reg}}$) on the pooled prostate set and $0.692$ and $0.716$ on cardiac. What the budget axis uses, the step-to-step rise, is intact,
which is consistent with the frozen design still clearing the fixed budget.

\paragraph{Pre-specification reconciliation.} Criteria for the M\&Ms and prostate replays were written to \texttt{operation.md} before those numbers were computed.

\begin{table}[h]
\centering\small\setlength{\tabcolsep}{4pt}
\begin{tabular}{p{0.42\linewidth}p{0.24\linewidth}p{0.26\linewidth}}
\toprule
Pre-specified expectation & M\&Ms & Prostate \\
\midrule
Random at matched rate is worse & held ($0/1{,}000$) & held ($P{=}0.010$) \\
Entropy/MSP intermediate or not better & held, and significantly & held in point estimates; difference not significant \\
Oracle is an upper bound & \textbf{failed} (see below) & held \\
\bottomrule
\end{tabular}
\caption{Pre-specification reconciliation for the M\&Ms and prostate matched-control replays.}
\label{tab:app-prereg}
\end{table}

The third expectation specified an oracle that buckets by true $\Delta$Dice. That oracle
reaches HA $0.0217$ on M\&Ms, \emph{worse} than the deployed fragmentation router's $0.0133$, so it is not an upper bound on HA. The specification, not the execution, was wrong: an
oracle bounds only the quantity it optimizes, and a $\Delta$Dice oracle optimizes Dice-level
harm. The HA oracle ($0.0023$) is the correct bound. We report the failure rather than
substituting the working oracle silently, and note that it is itself an instance of this
paper's thesis: selecting cases with perfect knowledge of the Dice outcome still accepts nearly
ten times the harmful area that selecting on HA would.

\paragraph{Three independent lines, one regime split.} The decomposition above is a
deployment-level replay. Two further analyses with no methodology in common reach the same
partition.

\begin{table}[h]
\centering\footnotesize\setlength{\tabcolsep}{3pt}
\begin{tabular}{p{0.20\linewidth}p{0.16\linewidth}p{0.17\linewidth}p{0.16\linewidth}p{0.16\linewidth}}
\toprule
Analysis & M\&Ms & Prostate & Prostate158 & ACDC (aggr.) \\
\midrule
Deployment replay (this section) & separates from every alternative & not distinguishable from confidence & not better than random & not better than random \\
Case-level ranking AUROC (Sec.~\ref{app:ci}) & $0.716$ vs.\ grad-norm $0.616$ & $0.685 \approx 0.681 \approx 0.684$ & $0.475$ and $0.506$, at chance & --- \\
Calibration perturbation (Sec.~\ref{app:calibsens}) & beats fixed-3 in $100\%$ of resamples at $n_{\mathrm{cal}}{=}5$ & ${\sim}1/5$--$1/3$ of resamples lose to fixed-3 & not run & router's value at or above the band center \\
\bottomrule
\end{tabular}
\caption{The same three-regime split recovered by deployment replay, case-level ranking and
calibration-perturbation propagation. Prostate158 falls in the same regime as ACDC on the deployment replay, the only analysis run on both; the calibration-perturbation propagation was not run on it.}
\label{tab:app-three-pillars}
\end{table}

\section{Clustered Intervals, Per-Step Structure, and Case-Level Ranking}
\label{app:ci}

The pooled correlations in the main text treat the four $(\text{case},\text{step})$ rows of one
case as independent observations. They are not: the steps of a case share an image, a source
prediction and one adaptation trajectory. Every interval below therefore resamples
\emph{cases}, carrying all of a case's steps together ($B{=}5000$).

\begin{table}[h]
\centering\small\setlength{\tabcolsep}{4pt}
\begin{tabular}{llcc}
\toprule
Benchmark & Signal & $\rho$ vs.\ HA [95\%] & $\rho(\Delta,\Delta\mathrm{HA})$ [95\%] \\
\midrule
Prostate, $n{=}124$ & $n_{\mathrm{reg}}$ & $0.594$ $[0.481,0.686]$ & $0.464$ $[0.352,0.563]$ \\
 & $\delta$ & $0.557$ $[0.428,0.659]$ & $0.515$ $[0.416,0.606]$ \\
 & entropy & $0.119$ $[-0.021,0.248]$ & $-0.290$ $[-0.399,-0.181]$ \\
 & MSP & $-0.161$ $[-0.296,-0.010]$ & $0.278$ $[0.170,0.385]$ \\
 & grad-norm & $0.270$ $[0.175,0.354]$ & $-0.035$ $[-0.108,0.034]$ \\
 & GraTa-cos & $-0.093$ $[-0.208,0.027]$ & $-0.009$ $[-0.106,0.086]$ \\
\midrule
Cardiac, $n{=}230$ & $n_{\mathrm{reg}}$ & $0.597$ $[0.532,0.657]$ & $0.455$ $[0.377,0.527]$ \\
 & $\delta$ & $0.516$ $[0.439,0.585]$ & $0.347$ $[0.266,0.421]$ \\
 & grad-norm & $0.146$ $[0.082,0.206]$ & $0.035$ $[-0.031,0.101]$ \\
 & entropy & $0.032$ $[-0.069,0.130]$ & $-0.006$ $[-0.082,0.073]$ \\
 & MSP & $-0.012$ $[-0.108,0.087]$ & $0.017$ $[-0.060,0.092]$ \\
 & $1-$agreement (aug.) & $0.264$ $[0.161,0.362]$ & $0.051$ $[-0.037,0.138]$ \\
 & $1-$agreement (dropout) & $0.047$ $[-0.064,0.157]$ & $0.016$ $[-0.073,0.105]$ \\
\midrule
Driving mild, $n{=}406$ & $n_{\mathrm{reg}}$ & $0.503$ $[0.445,0.557]$ & $0.290$ $[0.236,0.341]$ \\
 & $\delta$ & $0.478$ $[0.419,0.534]$ & $0.238$ $[0.168,0.306]$ \\
Driving aggr., $n{=}306$ & $n_{\mathrm{reg}}$ & $-0.147$ $[-0.238,-0.053]$ & $0.046$ $[0.001,0.091]$ \\
 & $\delta$ & $0.139$ $[0.051,0.230]$ & $0.187$ $[0.129,0.243]$ \\
\bottomrule
\end{tabular}
\caption{Case-clustered bootstrap intervals for every signal--HA correlation reported in the main text. Confidence proxies contain zero on cardiac, and on prostate only entropy's level interval does; so does the injected-dropout agreement; the sign inversion at the aggressive driving budget does not.}
\label{tab:app-clustered-ci}
\end{table}

\paragraph{Per-step structure, and what pooling costs.} Pooling can misrepresent a signal whose
scale drifts across steps, and one of ours is affected.

\begin{table}[h]
\centering\small\setlength{\tabcolsep}{6pt}
\begin{tabular}{llccccc}
\toprule
Benchmark & Signal & $k{=}1$ & $k{=}2$ & $k{=}3$ & $k{=}4$ & pooled \\
\midrule
Prostate & $n_{\mathrm{reg}}$ & $0.522$ & $0.679$ & $0.653$ & $0.428$ & $0.594$ \\
 & grad-norm & $0.460$ & $0.586$ & $0.551$ & $0.144$ & $0.270$ \\
Cardiac & $n_{\mathrm{reg}}$ & $0.527$ & $0.560$ & $0.675$ & $0.603$ & $0.597$ \\
 & grad-norm & $0.332$ & $0.164$ & $0.384$ & $0.190$ & $0.146$ \\
Driving mild & $n_{\mathrm{reg}}$ & $0.499$ & $0.428$ & $0.388$ & $0.405$ & $0.503$ \\
\bottomrule
\end{tabular}
\caption{$\rho(\text{signal},\mathrm{HA})$ computed within each step. Fragmentation is positive
at every step of every benchmark, so its pooled value is not an artifact of pooling. Grad-norm
on prostate is comparable to fragmentation through $k{=}3$ and collapses at the horizon; pooling
compresses that into a single low number, which understates it. We report the pooled protocol in
the main table for comparability and this decomposition alongside it.}
\label{tab:app-perstep}
\end{table}

At the aggressive driving budget the inversion is present from the first step
($-0.101$, $-0.218$, $-0.218$, $-0.171$, $-0.142$, $-0.153$, $-0.147$, $-0.141$ for
$k{=}1{\ldots}8$): the signal is not degrading along the trajectory, it is wrong at that
learning rate from the outset. Restricting those same trajectories to $k{\leq}4$, matching the
depth of the mild-budget runs, does not restore it ($-0.054$ rain, $-0.222$ snow, $-0.142$
night), which is what attributes the collapse to update size rather than trajectory length.
Per condition at the mild budget, $\rho(n_{\mathrm{reg}},\mathrm{HA})$ is $0.462$ (fog), $0.420$
(rain), $0.321$ (snow) and $0.372$ (night), with $\Delta$ columns $0.275$, $0.256$, $0.269$ and
$0.246$; the corresponding $\delta$ level values are $0.393$, $0.361$, $0.334$ and $0.359$. Per-step signal statistics at the aggressive budget were not stored for fog, which is why the pooled aggressive row covers three conditions.

\paragraph{Case-level ranking: which harm is predictable.} A correlation over
$(\text{case},\text{step})$ pairs does not say whether a step-$1$ reading ranks \emph{cases} by
the harm that deployment will incur. We therefore score each case by its step-$1$ signal and ask
how well that ranks two different notions of harm at the deployed budget.

\begin{table}[h]
\centering\small\setlength{\tabcolsep}{6pt}
\begin{tabular}{llcc}
\toprule
Benchmark & Signal & AUROC, $\Delta$Dice${<}0$ & AUROC, HA in top tertile \\
\midrule
Prostate & $n_{\mathrm{reg}}$ & $0.464$ $[0.36,0.56]$ & $0.685$ $[0.59,0.77]$ \\
 & $\delta$ & $0.446$ & $0.686$ \\
 & entropy & $0.555$ & $0.681$ \\
 & grad-norm & $0.441$ & $0.684$ \\
 & MSP & $0.435$ & $0.297$ (inverted) \\
\midrule
Cardiac & $n_{\mathrm{reg}}$ & $0.434$ & $\mathbf{0.716}$ $[0.65,0.78]$ \\
 & $\delta$ & $0.425$ & $0.692$ \\
 & grad-norm & $0.374$ $[0.30,0.45]$ & $0.616$ $[0.54,0.69]$ \\
\bottomrule
\end{tabular}
\caption{Step-$1$ signal as a case-level ranker, scored against two notions of deployment harm.
Orientation is fixed to ``higher signal $\Rightarrow$ positive''; MSP therefore ranks in the
inverted direction, which is itself informative. No signal ranks Dice degradation in the stated direction, cardiac grad-norm ranking it inverted; every signal ranks HA-level harm.}
\label{tab:app-auroc}
\end{table}

First, the two harms are different quantities and only one of them is
predictable, which is the case-level counterpart of the measurement-level claim in
Sec.~\ref{sec:prelim}. The sharpest single instance is grad-norm on cardiac: AUROC $0.374$ with
an interval excluding $0.5$, i.e.\ a \emph{high} gradient norm predicts a Dice
\emph{improvement}, on cases whose local edits are meanwhile harmful. Second, the ranking view
recovers the regime split of Sec.~\ref{app:matched} without sharing any machinery with it. On prostate the four informative signals are indistinguishable at the $0.68$ level, on cardiac
fragmentation separates from grad-norm.

\section{Which Half the Router Declines, and What It Costs}
\label{app:defenses}

\paragraph{Each signal deployed as a stop rule.} Table~\ref{tab:signal-ha} compares the
signals as \emph{predictors}; deploying each as a stop rule on prostate OOD-hard
($n{=}34$) compares them as \emph{controllers}. Fragmentation stops at HA $0.285$ at Dice
$0.582$ with no backward pass; grad-norm reaches $0.295$ at Dice $0.593$, so the
prediction-space signal matches it on harm while giving up a little overlap, which is the
trade it makes against a gradient-space one. The entropy and MSP variants reach $0.293$ and
$0.291$. All four therefore land within $0.010$ HA of each other. These are unpaired point
estimates on $34$ cases and we do not read the ordering among them as significant; what the
comparison establishes is that a read-only signal is not worse as a stop rule, not that it
is better.

\paragraph{Which half it declines.} Rolling back to $M_0$ gives
HA${=}0$ trivially but throws away accuracy ($0.738$ vs.\ FragSubset's $0.755$ on OOD-all), so the
goal is \emph{Dice-matched} HA reduction. The router rolls back $130$ of $230$ M\&Ms cases
and $42$ of $93$ prostate cases, and we read that as the intended behavior rather than as a
defect: \emph{whether this case should be adapted at all} is the question the paper poses,
and a controller that answers ``no'' for half a cohort is answering it. What separates this
from the degenerate solution is which half: on M\&Ms the retained cases are net-helped and
Dice stays at $M_0$ while HA falls to $0.013$, whereas on prostate they are not
(main text, prostate section). Indiscriminate abstention would sit at $M_0$ on both axes and order nothing. It is worth stating the comparison against $M_0$ plainly. On cardiac nobody wins it: fixed-4 adaptation is worth $+0.00006$ Dice in aggregate and the router $+0.000016$, both indistinguishable from doing nothing. On prostate fixed-1 does win it ($+0.014$), and the router, at $+0.0008$, declines nearly all of that gain. That is the claim the main text makes at this point (Sec.~\ref{sec:headline}). Where a stream does have something to gain the router keeps it: on ACDC it adds $+0.007$ to $+0.010$ mIoU over $M_0$ on fog, rain and snow, and gives up $0.0015$ on night ($-0.002$ at the precision of Table~\ref{tab:helphurt}), the condition where $M_0$ is near-collapsed. Nor is the reduction available to any equally conservative rule: at the router's
own rollback rate on M\&Ms, random assignment yields HA $0.060$ $[0.043,0.080]$ (all $1{,}000$ draws above the router's $0.013$), confidence-scored routing $0.054$--$0.057$, and
augmentation-agreement routing $0.045$ at eight forward passes per step; fragmentation reaches
$0.013$ with none, at $\Delta$Dice within $0.001$ of every alternative (Sec.~\ref{app:matched}). The
deployed edit set is small (coverage $\delta{=}3{\times}10^{-5}$; $17$ of $230$ cases retain
supra-threshold regions, with within-region HA $0.180$ and BA $0.248$; Sec.~\ref{app:accounting}): the
router's $0.013$ reflects deploying few edits and choosing them well, not many edits that
happen to be harmless.

\paragraph{The signal is not the metric.} The controllers do not read or optimize HA at decision time: HA requires
ground truth (Eq.~\ref{eq:ha}), whereas they act on label-free fragmentation, and the two
correlate at $\rho$ between $0.478$ and $0.597$ across benchmarks and coordinates at
comparable budgets (Table~\ref{tab:signal-ha}), not $1.0$. The negative control of Sec.~\ref{app:ablations}, which \emph{raises} HA, confirms that
the reduction follows from reading the right signal.

\paragraph{Cost.} Fragmentation costs ${\sim}37$\,ms/step with no backprop, against
${\sim}150$\,ms and $+2.5$\,GB for gradient-norm. Our controllers cut HA against the fixed budget without extra backprop, and the router does so at $0.55$--$1.10$ deployed updates per case ($1.00$--$1.67$ SGD steps once the discarded probe is counted, against $4.00$). CoTTA also cuts HA without extra backprop, at $M_0$'s Dice on M\&Ms and below it on prostate. The certified operating point for the monotone surrogate of Sec.~\ref{sec:controllers} is in Sec.~\ref{sec:supp-crc-op}.

\section{Edit-Area Accounting: Harmful, Beneficial and Neutral}
\label{app:accounting}

HA reports the harmful share of the edited area. The same partition has two other cells, and
reporting all three changes what the number means. For each disagreement region we compare
region-level error under $M_k$ and under $M_0$: the region is harmful if error rises,
\emph{beneficial} (BA) if it falls, and neutral if it is unchanged. All three are fractions of the same denominator $|D|$, so $\mathrm{HA}+\mathrm{BA}\le 1$ and the remainder is neutral area. ``Neutral'' covers two different things and the table does not separate them: disagreement voxels that lie in a region large enough to be scored but whose region-level error is unchanged, and disagreement voxels that lie in no scored region at all because every component containing them falls below the $16$-voxel minimum. The second is not a small residual: in $376$ of the $920$ cardiac $(\text{case},\text{step})$ rows the disagreement set is non-empty but no component survives the threshold, so the whole edit is unscored. A reader should therefore read neutral area as ``not judged harmful'' rather than as ``judged harmless''. This is a decomposition of
\emph{area}; Table~\ref{tab:helphurt} decomposes the same cases by \emph{case-level} Dice
change. The two are not interchangeable, and Table~\ref{tab:app-auroc} is the reason: the harm
visible in one is not the harm predictable from the signal in the other.

Because $|D|$ moves with the gate (the dynamic-denominator property of Sec.~\ref{sec:prelim}), BA${-}$HA differences are not comparable between controllers that
deploy different amounts of edit. The comparable quantity is the ratio BA/HA, and we report
coverage $\delta$ so the size of the denominator is visible. We also report each row twice:
over all cases, and over the cases that actually contain a region above the minimum-size
threshold.

\begin{table}[h]
\centering\small\setlength{\tabcolsep}{4pt}
\begin{tabular}{llccccccc}
\toprule
 & & \multicolumn{4}{c}{all cases} & \multicolumn{2}{c}{cases with regions} & \\
\cmidrule(lr){3-6}\cmidrule(lr){7-8}
Benchmark & Method / bucket & $n$ & HA & BA & BA/HA & $n$ & HA & coverage $\delta$ \\
\midrule
Cardiac & Router (all) & $230$ & $0.013$ & $0.018$ & $1.37$ & $17$ & $0.180$ & $3{\times}10^{-5}$ \\
 & \quad rollback bucket & $130$ & $0$ & $0$ & --- & $0$ & --- & $0$ \\
 & \quad shrink bucket & $46$ & $0.037$ & $0.091$ & $2.44$ & $14$ & $0.123$ & $7{\times}10^{-5}$ \\
 & \quad low bucket (depth 3) & $54$ & $0.025$ & $0.000$ & $0.00$ & $3$ & $0.446$ & $5{\times}10^{-5}$ \\
 & \quad retained (both) & $100$ & $0.031$ & $0.042$ & $1.37$ & $17$ & $0.180$ & $6{\times}10^{-5}$ \\
 & fixed-1 & $230$ & $0.186$ & $0.234$ & $1.26$ & $167$ & $0.256$ & $5.4{\times}10^{-4}$ \\
 & fixed-2 & $230$ & $0.164$ & $0.225$ & $1.37$ & $144$ & $0.263$ & $4.6{\times}10^{-4}$ \\
 & fixed-3 & $230$ & $0.129$ & $0.159$ & $1.23$ & $101$ & $0.294$ & $3.2{\times}10^{-4}$ \\
 & fixed-4 & $230$ & $0.177$ & $0.149$ & $0.84$ & $132$ & $0.308$ & $3.3{\times}10^{-4}$ \\
\midrule
Prostate & Router (all) & $93$ & $0.099$ & $0.098$ & $0.99$ & $35$ & $0.264$ & $8.3{\times}10^{-4}$ \\
 & \quad rollback bucket & $42$ & $0$ & $0$ & --- & $0$ & --- & $0$ \\
 & \quad mid bucket & $24$ & $0.252$ & $0.340$ & $1.35$ & $24$ & $0.252$ & $2.6{\times}10^{-3}$ \\
 & \quad low bucket & $27$ & $0.118$ & $0.036$ & $0.31$ & $11$ & $0.290$ & $5.4{\times}10^{-4}$ \\
 & \quad retained (both) & $51$ & $0.181$ & $0.179$ & $0.99$ & $35$ & $0.264$ & $1.5{\times}10^{-3}$ \\
 & fixed-1 & $93$ & $0.242$ & $0.327$ & $1.35$ & $77$ & $0.292$ & $5.4{\times}10^{-3}$ \\
 & fixed-2 & $93$ & $0.211$ & $0.283$ & $1.34$ & $69$ & $0.284$ & $5.6{\times}10^{-3}$ \\
 & fixed-3 & $93$ & $0.171$ & $0.269$ & $1.57$ & $69$ & $0.231$ & $5.4{\times}10^{-3}$ \\
 & fixed-4 & $93$ & $0.262$ & $0.284$ & $1.09$ & $81$ & $0.300$ & $5.6{\times}10^{-3}$ \\
\bottomrule
\end{tabular}
\caption{Three-way accounting of edited area. The \emph{cases with regions} block spreads the
same totals over the cases that retain a supra-threshold region, i.e.\ it rescales the
all-case columns by $n/n_{\text{regions}}$; the within-region BA this implies for the cardiac
retained row is $0.248$, the value quoted in Sec.~\ref{app:defenses}.
Neutral area is the complement of HA${+}$BA and is the largest cell in the all-case columns. BA/HA is comparable across rows; BA${-}$HA is not, because coverage
differs by an order of magnitude between the router and the fixed budgets.}
\label{tab:app-accounting}
\end{table}

Three things the table makes visible that HA alone does not. \emph{Adaptation repairs more than
it breaks in the early steps and reverses at the horizon}: on cardiac the fixed budgets run
BA/HA $1.26$, $1.37$, $1.23$ and then $0.84$ at $k{=}4$, which is the accounting form of the
non-monotonicity of Sec.~\ref{sec:budget}. Mean overlap is flat across that reversal.
\emph{The router's low HA is mostly abstention}: only $17$ of $230$ cardiac cases retain a
supra-threshold region at all, and coverage falls by an order of magnitude, so $0.013$ reports
deploying few edits with a favorable composition rather than many harmless ones.
\emph{The low-fragmentation bucket is the net-harmful one} on both medical benchmarks
(BA/HA $0.00$ on cardiac and $0.31$ on prostate) while the mid bucket is net-beneficial
($2.44$ and $1.35$; on cardiac the low bucket is also the one allowed to run longest,
at depth $3$, whereas prostate deploys both at depth $1$): the score orders where to do less, not where more would help, which is the
quantitative form of the limitation stated in Sec.~\ref{sec:prostate}.

\paragraph{Accuracy, at full precision.} Table~\ref{tab:mnm} prints the router and $M_0$ at the same
three-decimal Dice on cardiac, which invites the question of whether that is a rounding
coincidence. It is not: the paired difference over all $230$ cases is $+0.000016$ with a
case-bootstrap interval of $[-0.000070,+0.000100]$ and $P(\Delta{<}0){=}0.355$. The
decomposition by bucket shows why the two agree so closely: the $130$ rolled-back cases
contribute exactly zero by construction, and the two retained buckets very nearly cancel
(a mean $+0.000536$ over the $46$ shrunk cases against a mean $-0.000390$ over the $54$
low-bucket cases; weighted by bucket size these sum to the $+0.000016$ above).
The router therefore returns the source model's accuracy while deploying roughly a tenth of the edited area, which is a different statement from returning the source model.

\section{Calibration Sensitivity Propagated to Results}
\label{app:calibsens}

Perturbing the calibration split changes the cut-points, the cut-points change the bucket
assignment, and the bucket assignment changes what is deployed. This section propagates the
perturbation all the way to deployment metrics rather than stopping at the routing labels. For
each calibration size we draw $500$ subsamples without replacement, refit the cut-points on the
subsample, and replay deployment on the full evaluation split.

The quantile convention is not shared across benchmarks and must be copied per domain rather
than assumed: the M\&Ms router uses \texttt{np.percentile(x,[33.33,66.67])} and the driving
router \texttt{np.percentile(x,[33,66])}, both with linear interpolation. On the $n{=}400$
ACDC fog training scores the two differ in the fourth decimal ($0.01249$ vs.\ $0.01257$),
enough to move cases across a bucket boundary. The medical convention also coincides with
\texttt{np.quantile(x,[1/3,2/3])} to five decimals on this split, so the two are not
distinguishable from their output here; the convention has to be copied from the source
rather than inferred from a value. Refitting the deployed cut-points from the full calibration sets
reproduces them exactly on prostate and driving, and on cardiac to within $10^{-8}$, a difference that leaves the bucket assignment identical for all $230$ cases.

\begin{table}[h]
\centering\small\setlength{\tabcolsep}{5pt}
\begin{tabular}{llccc}
\toprule
Benchmark & $n_{\mathrm{cal}}$ & Dice [95\%] & HA [95\%] & beats fixed-3 \\
\midrule
Prostate & $5$ & $0.7398$ $[0.7370,0.7474]$ & $0.121$ $[0.026,0.226]$ & $75.6\%$ \\
 & $10$ & $0.7393$ $[0.7370,0.7421]$ & $0.114$ $[0.034,0.196]$ & $66.4\%$ \\
 & $15$ & $0.7392$ $[0.7369,0.7415]$ & $0.121$ $[0.069,0.183]$ & $79.4\%$ \\
 & $23$ (deployed) & $0.7387$ & $0.0994$ & --- \\
\midrule
Cardiac & $5$ & $0.8490$ $[0.8488,0.8492]$ & $0.022$ $[0.004,0.048]$ & $\mathbf{100\%}$ \\
 & $10$ & $0.8490$ $[0.8488,0.8492]$ & $0.018$ $[0.008,0.045]$ & $\mathbf{100\%}$ \\
 & $15$ & $0.8489$ $[0.8489,0.8490]$ & $0.017$ $[0.010,0.033]$ & $\mathbf{100\%}$ \\
 & $20$ (deployed) & $0.8489$ & $0.0133$ & --- \\
\bottomrule
\end{tabular}
\caption{Deployment metrics under calibration-set subsampling ($500$ draws per size). The last
column is the fraction of resamples whose deployed HA is below that of the retrospective
fixed-3 budget. Point estimates are means over the $500$ draws; brackets are the
$2.5$/$97.5$ percentiles of the draw distribution. Dice is almost invariant to the calibration set on cardiac and moves little on prostate; HA is not.}
\label{tab:app-calibsens}
\end{table}

The headline result survives the perturbation and the boundary result does not, which is the
same split the other two analyses find. On cardiac every resample at every size tested still
beats the retrospective fixed-3, the worst single draw at $n_{\mathrm{cal}}{=}5$ reaching HA
$0.055$ against fixed-3's $0.129$, while Dice moves by less than $5{\times}10^{-4}$: the router
never trades accuracy for the reduction regardless of how the tertiles are fitted. On prostate a fifth to a third of resamples lose to fixed-3, and the mean deployed HA is not monotone in the
calibration size ($0.121$, $0.114$, $0.121$ at $n_{\mathrm{cal}}{=}5$, $10$, $15$), which is
what a variance-dominated regime looks like.

On driving we can vary only the rollback quota, for the reason given in
Sec.~\ref{app:matched}; under that restriction the bands narrow monotonically with
$n_{\mathrm{cal}}$ (fog $[0.062,0.211]$ at $5$ to $[0.130,0.175]$ at $100$; night
$[0.056,0.328]$ to $[0.173,0.228]$) but the router's values sit at or above the band center
(snow $0.256$ against a mean of $0.250$ at $n_{\mathrm{cal}}{=}100$; these are the
fixed-quota values of Sec.~\ref{app:matched}, not the deployed router's
Table~\ref{tab:supp-acdc-fogsnow} figures). Better calibration does
not help when the coordinate itself does not separate at that budget.

\section{Sensitivity of HA to the Minimum-Region Threshold}
\label{app:tau}

HA counts only disagreement components of at least $\tau_{\text{area}}$ voxels. The
threshold was never swept while the archived run held only the step-4 endpoint mask; the
per-step dump of Sec.~\ref{app:rerun} makes the sweep a pure recomputation over that
rerun's masks. We report $\tau_{\text{area}} \in \{8,16,32,64\}$ around the deployed value of $16$.

\begin{table}[h]
\centering\small\setlength{\tabcolsep}{5pt}
\begin{tabular}{lccccccc}
\toprule
 & \multicolumn{4}{c}{cohort HA by step} & \multicolumn{3}{c}{deployed} \\
\cmidrule(lr){2-5}\cmidrule(lr){6-8}
$\tau_{\text{area}}$ & $k{=}1$ & $k{=}2$ & $k{=}3$ & $k{=}4$ & router & fixed-3 & $\Delta$HA \\
\midrule
$8$  & $0.197$ & $0.176$ & $0.144$ & $0.196$ & $0.024$ & $0.144$ & $-0.120$ \\
$16$ & $0.186$ & $0.166$ & $0.129$ & $0.179$ & $0.017$ & $0.129$ & $-0.113$ \\
$32$ & $0.172$ & $0.151$ & $0.111$ & $0.159$ & $0.009$ & $0.111$ & $-0.103$ \\
$64$ & $0.154$ & $0.138$ & $0.100$ & $0.130$ & $0.005$ & $0.100$ & $-0.095$ \\
\bottomrule
\end{tabular}
\caption{HA on cardiac ($n{=}230$) recomputed at four minimum-region thresholds. Replayed
from per-step checkpoints, so the M\&Ms mid bucket is scored at its step-2 state before
shrink; the router row here ($0.0167$ at $\tau_{\text{area}}{=}16$) is therefore not the
deployed value of Table~\ref{tab:mnm} ($0.0133$). All four rows share this convention, so
the comparison across $\tau_{\text{area}}$ is internally consistent. The fixed-$k$ rows are
computed on the per-step masks of the Sec.~\ref{app:rerun} rerun, which is why the
$\tau_{\text{area}}{=}16$ row differs from Table~\ref{tab:budget-sweep} by $0.002$ at
$k{=}2$ and $k{=}4$, the run-to-run scale that section reports.}
\label{tab:app-tau}
\end{table}

HA never increases with $\tau_{\text{area}}$ in any of the $920$ (case, step) rows, but that is an arithmetic property of the
definition rather than an observation about edit geometry: the denominator $|D|$ counts all
disagreement voxels and does not depend on the threshold, while the numerator sums only
components above it, so raising $\tau_{\text{area}}$ can only remove area from the numerator.
Nothing follows from it about whether small components are more often harmful; the
region-level contrast of Sec.~\ref{sec:supp-edit-geometry} addresses that question directly
and on prostate finds the opposite of the ``harmful edits are small debris'' reading. What
does not follow arithmetically, and is the point of the sweep, is that the router's rank is unchanged across an eightfold change in the threshold, with its margin over the retrospective-best budget staying between
$-0.095$ and $-0.120$. The count of cases whose edits are scored at all does move: at $k{=}4$, $148$ of
$230$ at $\tau_{\text{area}}{=}8$ against $86$ at $64$ (at $\tau_{\text{area}}{=}16$, on these same
rerun masks, the count is $167$/$145$/$102$/$132$ for $k{=}1{\ldots}4$, against
$167$/$144$/$101$/$132$ in Table~\ref{tab:app-accounting}: one boundary case flips at
$k{=}2$ and one at $k{=}3$, the same cross-execution effect as the HA offset above), so the threshold is in effect a
choice of how large an edit has to be before it counts.

\section{Verification Rerun and Reproducibility of HA}
\label{app:rerun}

The archived cardiac run stored only the deployed endpoint mask, so anything requiring the
intermediate masks (re-thresholding the minimum region size, a different connectivity, a boundary metric, a qualitative panel at $k{<}4$) needed a rerun. We reran EATA fixed-4 on all
$230$ vendor-B cases, dumping every step, and used the opportunity to log the two
probability-derived signals the archived cardiac trajectory lacks (entropy, MSP) plus a
consistency signal. The archived run remains authoritative for every main-table deployment result. Analyses
that need intermediate masks or per-voxel probabilities are computed on this rerun and
are identified where reported.

\paragraph{Layered alignment.} A rerun that does not reproduce the archived run is a different
experiment, so we check it, but at the layer where a deviation can actually originate.
Step $1$ is one adaptation step from an identical frozen state, so its gradient norm is fixed
by the learning rate, the entropy margin, the Fisher weight, the parameter subset and the
preprocessing: we require it to match to $10^{-6}$, and it does ($2.8{\times}10^{-7}$ maximum
over $230$ cases, median $7.5{\times}10^{-9}$, i.e.\ at float32 rounding). Downstream, sliding-window
inference and export are not bitwise deterministic across executions, so per-case Dice drifts by up to $1.5{\times}10^{-3}$; we therefore check the quantities the paper actually reports: per-step cohort-mean Dice (maximum deviation $3.6{\times}10^{-5}$ against a $5{\times}10^{-4}$ tolerance)
and cohort HA. Gating on per-case Dice, which no table prints, would leave only one way to pass:
fitting the tolerance to the observed spread.

\paragraph{HA is reproducible to about $\pm 0.002$, and why.} The cohort mean HA at step $4$
differs by $+0.0020$ between the two runs, which collapses to $-0.0001$ once five of $230$ cases
are excluded. Those five are not noise in the usual sense: HA is a fraction of a region-filtered
disagreement set, so when $|D|$ is small a single blob crossing the $16$-voxel minimum flips a
whole case. The largest single deviation is a case whose step-$2$ HA is $0.279$ in one run and
$0$ in the other, on a disagreement set of $86$ voxels, with region counts $1$ and $0$. Of the
eight cases with $|\Delta\mathrm{HA}|>0.05$, four cross the size threshold and four keep the
same region count while a borderline region changes side. We report $\pm 0.002$ as the run-to-run reproducibility of a cohort HA mean and note that every $\Delta$HA claimed in
this paper is at least ten times that scale, with exceptions in Secs.~\ref{app:ablations}, \ref{app:matched}, \ref{app:algo} and~\ref{app:newobj}, where the differences quoted are two to about four times it. The difference between FragSubset and FragStop on cardiac ($0.109$ vs.\ $0.115$) is one of them and should not be read as an ordering.

\paragraph{Consistency signals, and one specification that could not be run.} A dropout-agreement
monitor in the style of the closest related work is not computable on this backbone: the source
model is an nnU-Net \texttt{PlainConvUNet} constructed with \texttt{dropout\_op: None}, so
stochastic forward passes return identical outputs and the agreement column would be the
constant $1$. We therefore report two variants. The primary one perturbs with light
augmentations ($8$ passes, seeds $1000$--$1007$, independent flips of the two spatial axes at
$p{=}0.5$ with the geometry inverted exactly on the logits, brightness in $[0.9,1.1]$ and
Gaussian noise with $\sigma\in[0,0.05]$ on z-scored intensities); an identity-augmentation
self-test asserts agreement exactly $1$ before any value is logged. The secondary one injects
\texttt{nn.Dropout2d} ($p{=}0.1$) before each segmentation head via forward pre-hooks, which
leaves the state dict untouched. The model was never trained under dropout, so that perturbation
is out of distribution for it, and the result matches the caveat we recorded in advance: its
correlation with HA is $0.047$ with an interval containing zero (Table~\ref{tab:app-clustered-ci}),
against $0.264$ for augmentation agreement. We report it because the caveat was written before
the measurement, not after.

\paragraph{Provenance of the cardiac columns in the cross-benchmark table.} Entropy, MSP and the
consistency signals for cardiac are computed on this verification rerun rather than on the
archived run. The signals the two runs share agree to $\pm 0.002$
($n_{\mathrm{reg}}$ $0.596$ vs.\ $0.597$, $\delta$ $0.516$ vs.\ $0.516$, grad-norm $0.148$ vs.\
$0.146$), which is what licenses reading the new columns alongside the archived ones.

\paragraph{Provenance of the GraTa-cos row.} GraTa's alignment signal needs two
extra backward passes per step, so it was measured in a separate execution of the same
$n{=}124$ trajectories; the stochastic state therefore differs from the run behind the other
five rows. That run yields, for the other five signals, $0.594/0.447$ ($n_{\mathrm{reg}}$),
$0.554/0.510$ ($\delta$), $0.119/-0.298$ (entropy), $-0.162/0.283$ (MSP) and $0.261/-0.035$
(grad-norm), a maximum absolute difference of $0.017$ from the values printed in
Table~\ref{tab:signal-ha},
with the ranking of the signals and both conclusions the table supports unchanged. We report
the values as measured in each run rather than re-running to a single seed.

\section{Controller, Compute, and Design Provenance}
\label{app:algo}
\label{app:controller}

\begin{table}[h]
\centering\small
\begin{tabular}{p{0.94\linewidth}}
\toprule
\textbf{Algorithm 1}\quad COQR routing for one test case (no labels, and no backward pass for the routing decision itself) \\
\midrule
\textbf{Input:} frozen source weights $M_0$; case $x$; cut-points $(c_{\mathrm{lo}}, c_{\mathrm{hi}})$; per-bucket policy $\Pi_d$. All fitted once offline on a labeled holdout disjoint from the test stream. \\
\quad 1: reset to $M_0$; predict $\hat y_0$ \\
\quad 2: take one adaptation step; predict $\hat y_1$ \\
\quad 3: $u \leftarrow g_d(\phi(\hat y_0, \hat y_1))$ \hfill \emph{two masks only; no backward pass} \\
\quad 4: \textbf{if} $u \ge c_{\mathrm{hi}}$: restore $M_0$; \textbf{return} $\hat y_0$ \hfill \emph{rollback} \\
\quad 5: $b \leftarrow \textsc{mid}$ if $u > c_{\mathrm{lo}}$ else \textsc{low} \\
\quad 6: \textbf{if} $\Pi_d$ is fixed-depth \emph{(medical)}: run to depth $\Pi_d[b]$; on M\&Ms's \textsc{mid} bucket, interpolate half-way back to $M_0$; \textbf{stop} \\
\quad 7: \textbf{else} \emph{(driving)}: continue while $\Delta n_k \le 0$, evaluated only for $k \ge 2$, up to the cap $\Pi_d[b]$ \hfill \emph{step 1 is always kept} \\
\textbf{Output:} the deployed mask, and the number of updates actually deployed. \\
\bottomrule
\end{tabular}
\caption{The decision consumes only the two masks $(\hat y_0, \hat y_1)$ and is taken \emph{after} one adaptation step, so a rolled-back case still pays that step; what rollback saves is the remainder of the budget, while the \emph{Steps} column counts deployed updates, so a rolled-back case contributes $0$. Line 6 and line 7 are two different controllers, not two settings of one: the medical routers deploy a constant depth per bucket, and the stop condition present in that code path is vacuous under its indexing (see below).}
\label{tab:app-algorithm}
\end{table}

\begin{table}[h]
\centering\small\setlength{\tabcolsep}{6pt}
\begin{tabular}{lccc}
\toprule
Component & Per step & Backward passes & Peak extra memory \\
\midrule
Fragmentation $\phi(M_0,M_k)$ & ${\sim}37$\,ms & $0$ & negligible \\
Gradient norm & ${\sim}150$\,ms & $1$ & $2.5$\,GB \\
GraTa alignment & --- & $2$ & --- \\
Augmentation agreement ($8$ passes) & $8\times$ forward & $0$ & negligible \\
\bottomrule
\end{tabular}
\caption{Decision-time cost of each signal, over and above the adaptation step itself. The controller adds no backward pass; the gradient-based comparators do. GraTa's wall-clock and peak memory were not measured, and only its backward-pass count is relevant here. As a decision signal it is read-only (Sec.~\ref{sec:supp-baselines}); where Sec.~\ref{app:newobj} runs it as an adaptation objective, it is the objective rather than the controller.}
\label{tab:app-compute}
\end{table}

\paragraph{How the controller's design was arrived at, and what that costs.}
Four choices in the deployed controllers were made while an evaluation split
was visible, and three constants were not considered at all. We separate them
by a criterion a reader can check against the released code: whether the choice
has a command-line option and an archived alternative that was run and scored,
or is a hard-coded module constant with neither.

\emph{Selected on M\&Ms with evaluation-split visibility.} Three choices were
settled in an exploratory sweep in which twelve router variants were evaluated
on the vendor-B test split, the same $n{=}230$ split the main table reports.
That sweep moved the scoring coordinate from the two-term risk of Eq.~\ref{eq:risk} to
the disagreement ratio $\delta$. At the cut-points fitted for it, the two-term risk did
not partition this split at all: all $230$ cases fell in a single bucket, so those
variants reduce to fixed-depth policies, and the first of them reproduces the fixed-1 row
of Table~\ref{tab:budget-sweep} to within run-to-run noise (HA $0.186$ at depth $1$; the
two differ by $10^{-4}$, well inside the $\pm 0.002$ of Sec.~\ref{app:rerun}). The
budget-axis controllers read $r_k$ against a fixed sign threshold rather than tertiles and
are unaffected. With $\delta$ the same architecture gave a three-way split at HA $0.066$,
and later variants reached as low as the $0.040$ of the final architecture below. The
sweep also settled whether the middle bucket shrinks, halts or is delayed; and it settled
the depth of the low bucket, through an option named for a stopping policy
(\texttt{default} / \texttt{delay4} / \texttt{nostop}) whose three settings, under the
fixed-depth behavior documented above, amount to deploying the low bucket at depth $1$,
$3$ or $4$. The archived variants isolating that option reach HA $0.044$
(\texttt{delay4}, the setting carried into the deployed design) and $0.060$
(\texttt{nostop}). Both isolate the option under the online sliding-window quantile mode
(\texttt{quantile\_warmup}${=}15$, cut-points computed on the test stream) rather than the
seed-only calibration used everywhere in this paper (Sec.~\ref{sec:supp-calib}); the same
\texttt{delay4} setting under seed-only calibration is the $0.040$ of the final architecture. We name
the option as it appears in the code rather than as it reads, because a reader
who saw only the flag would take it for a data-driven stopping rule and not
for a choice of integer depth. The signal-validity argument of the main text
does not cover these choices and should not be read as covering them: that
argument establishes where a coordinate ceases to be usable, and on cardiac
neither coordinate is near its boundary, and $n_{\mathrm{reg}}$ is in fact the
stronger of the two on that benchmark (Table~\ref{tab:signal-ha}). The
reason $\delta$ is deployed there is the sweep, not the boundary.

\emph{Compared on ACDC with evaluation-split visibility.} One further choice
was made while looking at the driving evaluation split, and it is smaller in
both kind and degree: a single two-way comparison, not a search. The tertile
upper cut-point was compared at $P66$ and $P75$ on val, and $P66$ is deployed;
it is better on all four conditions (mean HA $0.203$ against $0.225$) at mIoU
differences of $0.0012$ or less. $P66$ is the defining value of a tertile, and the medical routers use the same convention at $P66.67$, so the
comparison did not move the deployed configuration off the default. An author who never looked at val would have deployed the same cut-point. And what $P66$
buys is a larger hard bucket ($18/21/29/43$ against $9/17/25/32$), i.e.\ more
rollback rather than better separation, which is what Sec.~\ref{app:matched}
independently finds at this budget and another instance of the signal
ordering where to do less. Neither qualification makes the comparison
label-blind, and we record it as an exposure.

\paragraph{A vacuous branch in the medical router.} The medical router's code path
retains a stop condition inside its per-step loop, but the condition cannot fire
selectively: the region-count history is appended at the end of the step body and
read at the start, so at step $k$ it holds $k-1$ entries and the test
$\textsc{choose}(\cdot) < k$ is satisfied for every case regardless of the geometry.
The consequence is the fixed depth per bucket documented in
Table~\ref{tab:supp-routing-coord} and measured in Table~\ref{tab:supp-depth-var}:
the fragmentation state enters the bucket assignment and not the depth. Every
deployed number in this paper is a measurement of that behavior and none of them
changes. The standalone FragStop controller
reported in the ablations evaluates the same condition on a complete archived
trajectory and is unaffected; and the matched-control experiment
(Sec.~\ref{app:matched}) varies only the score that assigns cases to buckets, so
what it measures is the whole of the signal's contribution rather than part of it.
Correcting the indexing yields a different controller, whose per-bucket depths,
tertiles and middle action would all have to be re-selected; we have not run it and
do not report a corrected version.

\emph{Unconsidered constants.} The shrink weight $\alpha{=}0.5$ is the default
value of the configuration object and was never varied. The ACDC step caps are
likewise hard-coded module constants with no command-line option and no
archived alternative: \texttt{MID\_MAX\_STEPS}${=}4$ carries over from the
medical configuration and \texttt{LOW\_MAX\_STEPS}${=}8$ is set equal to that
domain's own budget. These are unconsidered constants rather than tuned ones.

The thresholds were subsequently refit on a disjoint vendor-B calibration
split of twenty cases and are clean (the improvement from $0.040$ to $0.013$
between the final architecture and the reported result is entirely that
recalibration: the two runs agree on every recorded configuration field and
differ only in the cut-points), but recalibrating thresholds does not undo a
design selected with the test split visible.

We report this rather than presenting the architecture as given. Three things bound how much it
can be worth. The transfer results fix the template and refit thresholds, with the coordinate and per-bucket actions set per domain, but they are not uniformly clean and we now separate them. Prostate OOD-all
and Prostate158 took no part in any of the selections above: no coordinate,
action, depth or cut-point on either was chosen while their evaluation splits
were visible, and Prostate158 in addition freezes the entire M\&Ms design
rather than refitting a coordinate. Those two are unexposed. ACDC is not: the
cut-point comparison above was made on its evaluation split, so the driving
numbers carry an exposure of their own, smaller than the M\&Ms sweep but not
absent. What the driving section actually rests on is unaffected. The monotone TENT$\rightarrow\tau$-stop$\rightarrow$router cascade
(Sec.~\ref{sec:acdc}) and the signal-validity boundary at the aggressive
budget (Sec.~\ref{sec:signal-transfer}) are descriptive observations about
archived trajectories, and neither depends on how the cut-point was picked. The bound this paragraph offers, however, now covers two benchmarks rather than three. The matched-control experiment of Sec.~\ref{app:matched} compares scoring coordinates at a
fixed architecture, so the margin it measures is not attributable to the architecture search.
And the direction of the exposure is visible in Sec.~\ref{app:calibsens}: the cardiac advantage
survives calibration sets of five cases, which is not what a design fitted to $230$ evaluation
cases would be expected to do. A fourth bound is available offline, and we ran it after the fact. The sweep compared the
two-term risk against $\delta$; it never considered $n_{\mathrm{reg}}$, which is therefore a
coordinate the selection could not have favored. Substituting it into the deployed architecture, with the same actions, bucket sizes and calibration split and only the ranking score changed, gives HA $0.0366$ at Dice $0.8488$, against $\delta$'s $0.0133$ at $0.8489$. The
un-swept coordinate is worse than the swept one, as one would expect, but it still separates
from every matched control in Sec.~\ref{app:matched} (augmentation agreement $0.0451$, entropy
$0.0537$, MSP $0.0567$, random $0.0605$): the margin over the alternatives is not an artifact
of having searched, because a coordinate that was never searched retains most of it.

That experiment also supplies a reason for the cardiac choice that does not require the sweep.
Fitted natively rather than at matched quota, $n_{\mathrm{reg}}$ has no resolution on this
benchmark: its values on the twenty calibration cases are $0$ ($\times 10$), $1$ ($\times 6$)
and $2$ ($\times 4$), so its tertiles collapse to $0$ and $1$ and the middle bucket is empty.
The resulting router rolls back $167$ of $230$ cases and reaches HA $0.0039$ with
BA $0.0000$. It wins on HA by abstaining almost everywhere, which is the degenerate solution
Sec.~\ref{sec:defenses} rules out. $\delta$ takes twenty distinct values on the same twenty
cases and yields three non-empty buckets. \textbf{That contrast is visible in the calibration
set alone, with no evaluation data}, so an \emph{a priori} rule would have selected $\delta$ on cardiac without the sweep: do not route on a coordinate whose tertiles degenerate on the calibration split. We did not apply such a rule at the time, and we are not claiming
retrospectively that we did; we report it because a reader is entitled to know whether the
choice is defensible on evidence that was available without looking at the test split, and
here it is.

\paragraph{A frozen-design test on a benchmark outside the sweep.} The bounds
above all ask whether the search could have produced the reported margin. A
different question is whether the design that came out of it transfers at all,
and that one admits a direct test. We froze the M\&Ms router entirely, including the scoring coordinate $\delta$, the tertile structure and the per-bucket depths $0$ / $2$ with shrink / $3$, and ran it once on Prostate158
($n{=}126$), a benchmark that took no part in the sweep and that shares no
cases with any split reported elsewhere in this paper. Only the cut-points
were refit, on a $32$-case holdout of that benchmark, by the same seed-only
procedure. Nothing was tuned, and the run was scored once.

\begin{table}[h]
\centering\small\setlength{\tabcolsep}{6pt}
\begin{tabular}{lcccc}
\toprule
Method & Dice & HA & BA & Steps \\
\midrule
$M_0$ (no adapt, ref.) & 0.7985 & 0 & 0 & 0 \\
fixed-1 & 0.8023 & 0.311 & 0.439 & 1.00 \\
fixed-2 & 0.8031 & 0.249 & 0.434 & 2.00 \\
fixed-3 \emph{(deployable = retrospective)} & 0.8023 & 0.228 & 0.376 & 3.00 \\
fixed-4 \emph{(default)} & 0.8004 & 0.298 & 0.382 & 4.00 \\
\midrule
Frozen M\&Ms design & 0.8000 & $\mathbf{0.139}$ & 0.224 & 1.76 \\
\bottomrule
\end{tabular}
\caption{The frozen-design test on Prostate158 ($n{=}126$ evaluation cases). The design is
frozen from M\&Ms; only the cut-points are refit, on a $32$-case holdout of this benchmark.
The router rolls back $39$ of $126$ cases. The deployable and retrospective budgets coincide
here at $k{=}3$.}
\label{tab:app-frozen-p158}
\end{table}

Three readings, one of which is unfavorable. First, the reduction clears the
strongest reference available: $\Delta$HA against the retrospective-best
fixed-3 is $-0.089$ $[-0.124,-0.055]$ with $P{=}1.000$, at $\Delta$Dice
$-0.0023$ $[-0.0050,+0.0001]$, an interval that includes zero, and the router
sits above $M_0$ ($+0.0015$ $[+0.0007,+0.0023]$). The Dice interval only
barely includes zero ($P(\Delta{<}0){=}0.972$), so we read it as an interval
containing zero rather than as an absence of cost. Second, the three reference
tiers collapse to two here: the $32$-case holdout selects $k{=}3$, which is
also the evaluation optimum, so Tier-A and Tier-B coincide. This is the
situation ACDC already poses (Sec.~\ref{sec:acdc}), a budget that is both optimal and selectable, and the answer is the same, with a larger margin
($-0.089$ against ACDC's $-0.048$). Third, the composition of the deployed
edits does not improve: BA/HA is $1.61$ for the router against $1.28$--$1.74$
across the fixed-$k$ ladder, so the HA reduction comes from deploying less
rather than from deploying better. That is a further independent replication
of the limitation stated in Sec.~\ref{sec:prostate} and quantified in
Sec.~\ref{app:accounting}.

As elsewhere we report retained-only figures so the abstention component is
visible: over the $87$ retained cases HA is $0.201$ at coverage
$\delta{=}1.02{\times}10^{-3}$, against $0.139$ and $7.02{\times}10^{-4}$ over
all $126$. Unlike cardiac, where only $17$ of $230$ cases retain a
supra-threshold region, here the set of cases with scorable regions coincides
exactly with the retained set: every case this router does not roll back
deploys an edit large enough to be scored.

\paragraph{What the frozen test does and does not bound.} It establishes that
the design transfers: applied unchanged to a benchmark it could not have been
fitted to, it clears that benchmark's own retrospective optimum. It does not
establish that the M\&Ms number is unbiased, and the three-benchmark
comparison says as much. The relative HA reduction is $90\%$ on M\&Ms, $42\%$
on prostate OOD-all and $39\%$ on Prostate158. The two benchmarks outside the sweep agree with each other and not with the one inside it. Three accounts are consistent with that pattern and we cannot separate them here. The rollback
rates differ in the same direction ($57\%$, $45\%$, $31\%$), so part of the
gap is that M\&Ms declines more; but the relation is not proportional. Twelve points more rollback than prostate buys forty-eight points more reduction, so rollback rate alone does not account for it. The edit sets also differ by an
order of magnitude in coverage ($3{\times}10^{-5}$ on cardiac against
$7{\times}10^{-4}$ here), and HA is a fraction of that set, so a sparse edit
set is easier to drive near zero. Selection effect is the third account. On Prostate158 a matched-bucket
control now bears on it directly: at the same quota and the same depths the
ranking is not better than random (Sec.~\ref{app:matched}), so the reduction
there is carried by the quota rather than by case selection. Whether the same
is true of the $90\%$ on M\&Ms is not settled here, and we do not claim to have
excluded a selection effect on the benchmark the design was chosen on. A reader who treats $39\%$--$42\%$ as the unbiased estimate of what this design and its template buy, and $90\%$ as an upper bound
specific to the benchmark it was selected on, is reading the evidence the way
we do.

A design search on a held-out split, which we did not run, would settle the question; we list
it as the first thing to redo rather than as a limitation that argues itself away.

The verification rerun of
Sec.~\ref{app:rerun} took ${\sim}27$\,s per case for four steps with both consistency
signals; materializing the shrink action for all $230$ cardiac cases took $2.3$\,s per case.

\section{A Transformer Backbone: What Ports and What Does Not}
\label{app:swin}

\begin{figure}[t]
\centering
\includegraphics[width=0.72\linewidth]{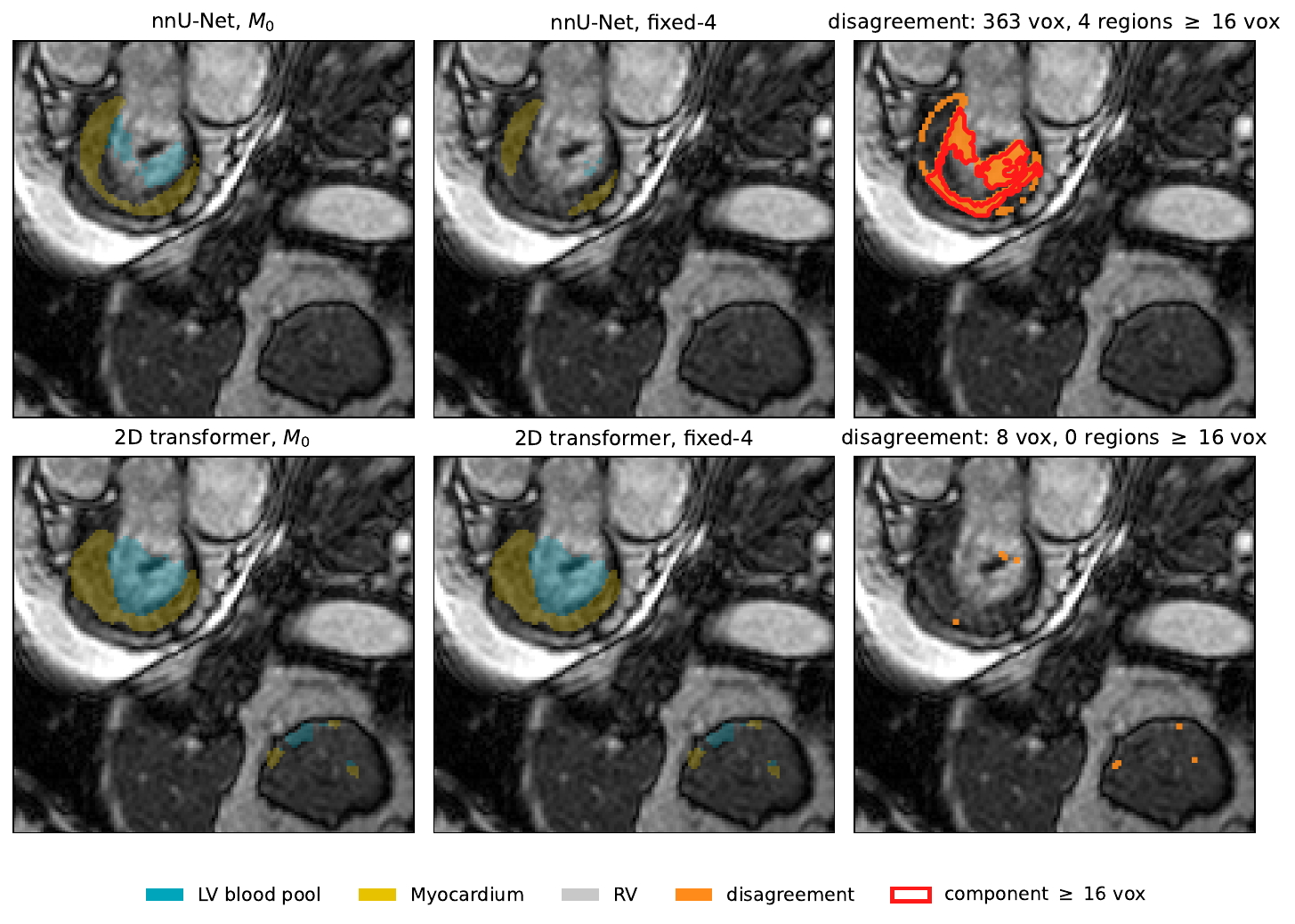}
\caption{Edit geometry across backbones, one case (\texttt{G1N6S7\_frame12}, slice $8$;
$M_0$ Dice $0.75$ nnU-Net / $0.63$ 2D transformer). Columns: $M_0$, fixed-$4$, and their
disagreement (orange), with $3$D components of at least $16$ voxels outlined in red. Over the volume the nnU-Net edit forms
$5$ such components and the transformer edit forms none (largest $2$ voxels); on the slice
shown, $4$ and none. On this case fixed-$4$ lowers nnU-Net Dice by $0.016$ at HA $0.77$,
which is what the scorable components capture. The two edits differ in volume
($363$ against $8$ voxels here) and in structure; Sec.~\ref{app:swin} rests on the second.
Orange is dilated by one pixel for visibility, so it overstates the true area; the counts
in the titles are undilated. The grey legend entry has no instance here: neither model
predicts RV on this slice, and the ground truth has none. Case and slice selected by the rule of Sec.~\ref{sec:supp-qual-select}: the median
transformer-side disagreement volume among the $7$ cases with both $M_0$ in $[0.60,0.80]$,
and the slice carrying the most nnU-Net supra-threshold disagreement area. The case is
conservative for the claim (nnU-Net largest component $116$, below the band mean of $345$).}
\label{fig:swin-geom}
\end{figure}

\paragraph{A source model below its own usability line.} A 3D Swin-UNETR reached $0.6861$
source-domain Dice against nnU-Net 2D's $0.8916$, triggering a pre-registered fallback to a
2D transformer. Three 2D runs placed their final checkpoints at $0.8394$, $0.8130$ and
$0.8220$. The best checkpoint cleared the $0.8416$ fallback line every time and the final checkpoint never did. Under the pre-registered convention of reporting the final checkpoint, this source model therefore sits $0.020$ below it. On vendor-B its $M_0$ Dice is $0.630$
against $0.849$, and it segments non-cardiac structures that nnU-Net does not (at lower
right in Fig.~\ref{fig:swin-geom}, where the ground truth labels nothing). Its step-1 edit volume on vendor-B is $\delta{=}1.45{\times}10^{-4}$, $0.27$ of nnU-Net's $5.45{\times}10^{-4}$, at the learning rate the pre-registration fixed in advance. What follows is conditional on a source model weaker than the main text's.

\paragraph{One coordinate transfers, the other does not.} At the pre-registered
$\tau_{\mathrm{area}}{=}16$, $n_{\mathrm{reg}}$ predicts HA at $\rho{=}0.669$
$[0.446,0.836]$, clearing the $0.4$ bar; $\delta$ reaches $0.151$ $[0.081,0.206]$. The
criterion requires both, so it is not met, and it is $\delta$ that fails. The interval is
wide because HA is non-zero in only $9$ of $920$ case-steps ($8$ of $230$ cases), leaving the case-clustered bootstrap little to resample. We read only that its lower bound clears
$0.4$, and claim no ordering against the nnU-Net line, whose narrower interval
$[0.532,0.657]$ lies wholly inside this one. One reservation covers this whole section. Cohort HA never exceeds $0.0053$ here at any $k$, so almost none of the edited area is scored as harmful. Every verdict in this section is therefore read off a set of harmful edits that barely exists.

\paragraph{The edits have no coherent core.} Matching on $M_0$ within $[0.60,0.80]$
(each backbone's own cases in the band; the two distributions overlap on only $7$), the
largest disagreement component averages $345$ voxels on nnU-Net ($43$ cases) against $3.5$
on the 2D transformer ($115$ cases), a factor of $97.7$. The share of changed voxels in singletons is $0.33$ against $0.83$. Lowering the threshold recovers nothing: $\rho$ falls monotonically
to $0.574$, $0.560$ and $0.337$ at $\tau_{\mathrm{area}}{=}8,4,2$. Here $16$ sits close to
the boundary between edit and noise. This is independent support for the main text's choice, from a backbone whose edit geometry is entirely different.

\paragraph{Consequences.} $\delta$ is a share of changed voxels and under scatter mostly
measures noise, whereas $n_{\mathrm{reg}}$ counts only supra-threshold components: the same
$\tau_{\mathrm{area}}$ filters for one and does nothing for the other. The deployed M\&Ms
router scores on $\delta$ and the prostate router on $n_{\mathrm{reg}}$, so on this backbone
it is the cardiac coordinate that fails to port. The router's $+0.00044$
$[+0.00031,+0.00057]$ Dice over fixed-$4$ is no gain: router${-}M_0$ is $-0.00004$
$[-0.00011,+0.00003]$ while fixed-$4{-}M_0$ is $-0.00048$ $[-0.00061,-0.00036]$. It avoids a
loss rather than earning a gain: the router sits at $M_0$ here as on cardiac
(Sec.~\ref{sec:headline}), and the difference is that fixed-$4$ falls below it.

\section{Two Other Adaptation Objectives: What the Conclusions Survive}
\label{app:newobj}

\paragraph{What this section is for.} The main text claims that \emph{whether} a step should
be deployed is independent of \emph{how} the trajectory is optimized. This section tests that
on two published objectives: GraTa \citep{grata2025}, a gradient-alignment method for
medical segmentation, and DeYO \citep{deyo2024}, a classification method whose sample
selection we translate to slices. Both generate fixed-$k$ trajectories on the benchmarks,
splits and protocol of the main tables, and the router of Sec.~\ref{app:matched} is replayed
offline on each. Criteria V0--V4 were fixed before any run and are reported whether or not
they hold (Table~\ref{tab:newobj}); one arm and one reduction added after the data existed
are marked as such.

\paragraph{What was ported, and what deviates.} GraTa follows the official release step for
step: an entropy gradient at $\theta$ and a unit-step perturbation
$\theta'{=}\theta{-}\nabla_{\mathrm{ent}}$ with no learning rate; a target averaged over six
weak views (identity, two flips, three rotations) after inverting each view's geometry on the
probabilities; a single strongly augmented view; and an Adam step along the gradient
$\nabla_{\mathrm{con}}$ of the consistency cross-entropy $L_{\mathrm{con}}(\theta')$ with
$\eta{=}10^{-4}\cdot\tfrac14(\cos{+}1)^2$, where $\cos$ is the cosine between the two
gradients. InstanceNorm affine parameters, the shared optimizer and budget, and the episodic
reset follow Sec.~\ref{sec:supp-eata} rather than each method's official choice (the official
DeYO, for instance, uses SGD), so that the only intended difference between objectives is the
objective. Five deviations are specific to this section. (i) Head: the official head is
sigmoid and two-class; ours is softmax over $C$ classes ($C{=}4$ cardiac, $C{=}2$ prostate),
and the mean of six softmax maps is \emph{not} renormalized. (ii) Steps: the official method
takes one step per case; we run $K{=}4$ to obtain a ladder, so fixed-$1$ \emph{is} native
GraTa. (iii) Gamma: the five augmentation probabilities, $0.75/0.75/0.75/0.5/0.5$, are the
official ones, but our gamma transform keeps $\mathtt{retain\_stats}$ true where the official
release leaves the library default of false. (iv) Blur: the official blur is applied per
channel with probability $0.5$ inside a sample selected with probability $0.5$; ours omits the
per-channel draw, so on single-channel MRI a strong view is blurred with probability $0.5$
rather than $0.25$. This was an oversight in the port, found only when checking against the
official source, and its effect is measured below. (v) Intensity range: the strong
augmentation acts on $z$-scored intensities rather than the official $[0,1]$ fundus range, so
multiplicative brightness is a gain change and a noise variance of $0.05$ is $\sigma{\le}0.22$
in $z$ units.

\paragraph{The official-lr arm does not move.} On cardiac vendor-B, GraTa at fixed-$1$ scores
$0.8489$ Dice against $M_0$'s $0.8489$ and TENT's $0.8507$; on prostate, $0.7364$ against
$0.7379$ and $0.7518$. The pre-registered V0 requires GraTa to reach at least TENT and not to
fall more than $0.01$ below $M_0$. The second half holds on both benchmarks ($+0.0000$ and
$-0.0015$); the first fails on both. The published GraTa gains $5.9$ Dice points over its
source model across five fundus domains and beats TENT and SAR there, so the direction does
not reproduce: what appears is not a loss but no effect. Step-$1$ $\delta$ is
$2.26{\times}10^{-5}$ on cardiac and $3.27{\times}10^{-4}$ on prostate, $0.041$ and $0.062$
of what the main text's objective produces on the same data, and the effective learning rate,
$1.3{\times}10^{-5}$ and $2.1{\times}10^{-5}$, is a thirty-eighth to a twenty-fourth of the
shared base.

\paragraph{Why the learning rate collapses.} The multiplier $\tfrac14(\cos{+}1)^2$ is small
when the two gradients disagree, and here they disagree systematically: $\cos$ averages
$-0.300$ on cardiac ($n{=}920$ case-steps, s.d.\ $0.206$) and $-0.108$ on prostate ($n{=}372$,
s.d.\ $0.199$). Of five non-degeneracy checks, four pass; the fifth, a multiplier spanning at
least $[0.2,0.9]$, fails, since it never exceeds $0.53$. Part of the mechanism is the source
model's confidence. At step $1$ on the evaluation cases ($n{=}230$ cardiac, $93$ prostate), the
six-view target at $\theta'$ has mean per-pixel entropy $0.0039$ and $0.0150$, $4.9$ and $5.2$
times that of the identity view ($0.0008$, $0.0029$). Averaging near-one-hot maps that disagree
only at boundaries gives a softer target, so descending the consistency cross-entropy pushes
\emph{against} entropy descent. That explains about half of the sign: hardening the target to
one-hot moves step-$1$ $\cos$ from $-0.295$ to $-0.170$ on cardiac and from $-0.100$ to
$-0.051$ on prostate, still negative, while the entropy gradients of the plain and strongly
augmented inputs align at $+0.53$ and $+0.43$ on average. The remainder is not identified: target
hardening and the official blur of (iv) were tested and neither closes it, and (v) was measured
but not ablated. Deviation (v) contributes: the backbone was trained under this augmentation
family, and a strong view leaves $0.963$ and $0.908$ of labels unchanged on the union of
foreground, supplying little of the disagreement GraTa expects to exploit. Deviation (iv) runs
the other way, and only slightly: with the official blur $\cos$ becomes marginally more
negative ($-0.300$, $-0.109$), so the oversight made GraTa's steps slightly larger, not
smaller. This is a
statement about GraTa's applicability to a very confident backbone, not about its correctness.

\paragraph{Given a comparable learning rate, the objective does produce edits, and on prostate
it does harm.} To separate ``the trajectory did not move'' from ``the objective does nothing
on this backbone'', a supplementary arm, \emph{GraTa (shared lr)}, uses base learning rate
$5{\times}10^{-4}$, the value DeYO, TENT and EATA already use, and keeps the cosine modulation
as part of the objective. It was added \textbf{after} the official-lr arm had run and is not
pre-registered, since the pre-registration says to keep GraTa's own rule; it is therefore reported beside the official-lr arm and never in place of it. Cardiac HA rises from $0.008/0.019/0.026/0.041$
to $0.061/0.102/0.119/0.127$, a factor of $3.1$ at $k{=}4$, and $\delta$ rises to $0.197$ and
$0.311$ of the reference, still well short of it, since $\cos$ is unchanged by construction.
Dice does not follow. On cardiac it stays at $M_0$ ($0.8491$ at $k{=}1$, $0.8493$ at $k{=}4$);
on prostate it falls monotonically, $0.7296 \to 0.7225 \to 0.7165 \to 0.7099$, from an $M_0$
of $0.7379$. V0's first half still fails in both arms.

\paragraph{A slice-level adaptation of DeYO, and one reduction changed after the fact.} The
official DeYO has no segmentation experiment, so this is a translation and is named as one
throughout, as Sec.~\ref{sec:supp-eata} does for EATA. The sample is a slice, and the per-pixel drop
$p(x)_{\hat y}-p(x')_{\hat y}$ under a block shuffle $x'$ is taken against the original
pseudo-label. Thresholds are the official defaults converted to $C$ classes:
$\tau_{\mathrm{Ent}}{=}0.5\ln C$, $\mathrm{Ent}_0{=}0.4\ln C$, $\tau_{\mathrm{PLPD}}{=}0.2$.
Reducing the drop over \emph{all} pixels, as pre-registered, selected \emph{no slice at all} on
either benchmark at any of the four steps ($2434$ cardiac and $1050$ prostate evaluation slices
per step), leaving both DeYO trajectories identical to $M_0$. The cause is class imbalance, not
the shuffle: $96.9\%$ and $96.7\%$ of a slice's pixels are pseudo-labelled background, whose
near-zero drop dilutes the slice mean to at most $0.124$ and $0.143$, while on pseudo-label
foreground the median drop is $0.989$ and $0.981$. We therefore reduce over foreground pixels.
\textbf{This is a post-hoc change to a pre-registered quantity}, authorized and recorded as such. Selection
stays at slice level, not the pixel-level variant the pre-registration excludes, and slices
with no pseudo-label foreground get a drop of zero. After the change, $74.9\%$ and $99.0\%$ of
slices are selected; exactly one cardiac case-step out of $920$ still selects nothing (mean
drop $0.070$), and none on prostate.

\paragraph{Budget non-monotonicity does not reproduce.} V2 asks that cohort HA be non-monotone
in $k$ with $\mathrm{HA}(4)$ above the ladder minimum. It holds in one of six cells. In the
other five HA rises monotonically with the minimum at $k{=}1$, including both GraTa arms, and
since the shared-lr arm tripled HA and stayed monotone, this is not an artifact of small steps.
The passing cell is the DeYO translation on prostate, $0.2405/0.2954/0.3090/0.3038$, where a
$0.0052$ fall from $k{=}3$ to $k{=}4$ meets both conditions. The non-monotone budget of the
main text depends on the objective and the horizon it uses; under these two objectives the
cheapest budget is also the safest.

\begin{table}[t]
\centering
\small
\caption{The four pre-registered criteria on two objectives, two benchmarks, and the
supplementary shared-learning-rate arm; criteria as defined in the text. $\rho(\delta)$
and $\rho(n_{\mathrm{reg}})$ are the pooled case-clustered Spearman correlations against
HA on which V1 is decided. V4 is the share of the router's HA reduction reproduced by a
matched-quota control and is reported, not tested. GraTa (shared lr) and the foreground
reduction of the DeYO drop were both introduced after the data existed.}
\label{tab:newobj}
\begin{tabular}{llcccccc}
\toprule
Objective & Benchmark & $\rho(\delta)$ & $\rho(n_{\mathrm{reg}})$ & V1 & V2 & V3 & V4 random \\
\midrule
GraTa (official lr)   & cardiac  & $0.362$ & $0.737$ & $\times$ & $\times$ & $\checkmark$ & $0.742$ \\
GraTa (official lr)   & prostate & $0.734$ & $0.883$ & $\times$ & $\times$ & $\checkmark$ & $0.855$ \\
GraTa (shared lr)     & cardiac  & $0.564$ & $0.719$ & $\times$ & $\times$ & $\checkmark$ & $0.720$ \\
GraTa (shared lr)     & prostate & $0.802$ & $0.720$ & $\times$ & $\times$ & $\checkmark$ & $0.752$ \\
DeYO (slice-level)    & cardiac  & $0.475$ & $0.463$ & $\checkmark$ & $\times$ & $\checkmark$ & $0.694$ \\
DeYO (slice-level)    & prostate & $0.372$ & $0.367$ & $\times$ & $\checkmark$ & $\times$ & $0.821$ \\
\bottomrule
\end{tabular}
\end{table}

\paragraph{The signal reproduces in one cell of six.} V1 asks that the pooled case-clustered
$\rho$ of both $\delta$ and $n_{\mathrm{reg}}$ against HA fall in $[0.4,0.6]$, a two-sided
band, unlike the one-sided $0.4$ bar of Sec.~\ref{app:swin}; each was fixed before its own
runs and neither is revised here. Only the DeYO translation on cardiac satisfies both ($0.475$
$[0.375,0.561]$ and $0.463$ $[0.369,0.549]$). The other five leave the band in both directions
(Table~\ref{tab:newobj}). Above it lie six of the eight GraTa values: $n_{\mathrm{reg}}$ in all
four cells and $\delta$ on prostate in both arms. Below it lie $\delta$ in the official-lr arm
on cardiac ($0.362$) and both values for the DeYO translation on prostate. So three of the five failures are failures of a correlation stronger than pre-registered, one (the DeYO translation on prostate) is one where it has weakened, and one is mixed: in the official-lr arm on cardiac $\delta$ falls below the band while $n_{\mathrm{reg}}$ sits above it. Raising the step size moves cardiac $\rho(\delta,\mathrm{HA})$ from
$0.362$ into the band at $0.564$, which suggests the low official-lr value reflects a
trajectory that barely moves rather than a signal that does not exist. Under the
pre-registered criterion, the claim that the fragmentation signal reproduces across adaptation
objectives is not established.

\paragraph{The router wraps these trajectories; its ranking is not what earns the
reduction.} With cut-points refit on each arm's own calibration split and the actions of
Sec.~\ref{app:matched} unchanged (cardiac $3/2/0$ with the mid bucket at depth $2$ and no
shrink, prostate $1/1/\text{rollback}$), the router lowers HA in all six cells. V3's Dice
condition, that the interval on the cost of routing end at most $+0.005$, holds everywhere but
the DeYO translation on prostate ($+0.0178$ $[-0.0007,+0.0389]$, the cell where fixed-$k$
gained the most Dice and the router declines it), so the related-work claim that a
fragmentation router wraps a GraTa-optimized trajectory as it wraps ours is supported in the
narrow sense that wrapping is possible and nearly free. On the shared-lr arm on prostate the
router is $0.0284$ $[0.0140,0.0443]$ Dice \emph{better} than fixed-$4$, because fixed-$4$
there is harmful. A matched-quota random control (bucket sizes and actions kept, ranking
replaced by $1000$ permutation draws) removes most of the weight from that support: it
reproduces $69.4\%$ to $85.5\%$ of the router's HA reduction across the six cells, and
ranking by $n_{\mathrm{reg}}$ reproduces $79.2\%$ to $100.0\%$. On prostate the two
non-rollback buckets share an action, so the router reduces to declining a fixed quota of
cases, which a random quota of that size does comparably well. The defensible reading is that
these trajectories \emph{can} be wrapped without losing Dice, not that the benefit of wrapping
them comes from ranking cases by fragmentation.

\paragraph{Two disclosures.} First, an earlier round of these runs was discarded. The strong
augmentation seeded only NumPy's global generator while two of the five transforms draw from
Python's \texttt{random}, and the per-step seed omitted the case identifier, so every case got
the same augmentation draw and, in the DeYO translation, the same block permutation. No conclusion changed
when this was fixed, but that round was not reproducible from its own recorded seeds and is
kept only as a record. In it, a three-case pilot reported V0 as satisfied on cardiac ($0.8624$
against $0.8529$) where all $230$ cases give $0.8489$ against $0.8507$. A pilot can refute a
threshold but cannot establish one. Second, Sec.~\ref{app:swin} faced a comparable question
and answered it differently. Its pre-registration committed to leaving the learning rate alone
even at a large $\delta$ mismatch, and its measured ratio ($0.27$) never approached the one
here ($0.041$), so it has no shared-lr arm. The sources also differ: there $\delta$ follows
from the backbone's edit geometry, so changing the learning rate would change what is being
measured; here it follows from GraTa's own learning-rate rule, so the shared-lr arm ablates
that rule's base factor and nothing else. $\cos$ is unmoved by the base rate ($-0.306$ against
$-0.300$ on cardiac, $-0.115$ against $-0.108$ on prostate), so the arm should not be read as a
rate tuned when convenient.

\end{document}